\documentclass[final,5p,times,twocolumn]{elsarticle}
\usepackage[utf8]{inputenc}
\usepackage{hyperref}
\usepackage{bm}
\usepackage{nicefrac}
\usepackage{amsmath}
\usepackage{caption}
\usepackage[capitalise]{cleveref}
\usepackage[ruled,vlined]{algorithm2e}
\usepackage{booktabs}

\usepackage{tikz}
\def\checkmark{\tikz\fill[scale=0.4](0,.35) -- (.25,0) -- (1,.7) -- (.25,.15) -- cycle;} 

\hypersetup{colorlinks=true,linkcolor=blue,anchorcolor=red,citecolor=blue,filecolor=red,urlcolor=red,pdfauthor=author}

\biboptions{semicolon,round,sort,authoryear}

\newcommand{\abbreviations}[1]{%
  \nonumnote{\textit{Abbreviations:\enspace}#1}}

\begin{document}

\begin{frontmatter}

\title{Sketch-based Medical Image Retrieval}

\author[a,b]{Kazuma Kobayashi\corref{correspondingauthor}}
\cortext[correspondingauthor]{Corresponding to: Kazuma Kobayashi, M.D., Ph.D. {\it Postal address:} Division of Medical AI Research and Development, National Cancer Center Research Institute, 5-1-1 Tsukiji Chuo-ku, Tokyo, 104-0045, Japan; {\it Email address:} {\texttt kazumkob@ncc.go.jp}}
\ead{kazumkob@ncc.go.jp}

\author[c,d]{Lin Gu}
\ead{lin.gu@riken.jp}

\author[e,b]{Ryuichiro Hataya}
\ead{ryuichiro.hataya@riken.jp}

\author[f]{Takaaki Mizuno}
\ead{takamizu@ncc.go.jp}

\author[g]{\\Mototaka Miyake}
\ead{mmiyake@ncc.go.jp}

\author[g]{Hirokazu Watanabe}
\ead{hirwatan@ncc.go.jp}

\author[h]{Masamichi Takahashi}
\ead{masataka@ncc.go.jp}

\author[i]{Yasuyuki Takamizawa}
\ead{ytakamiz@ncc.go.jp}

\author[j]{\\Yukihiro Yoshida}
\ead{yukiyosh@ncc.go.jp}

\author[k,l,m]{Satoshi Nakamura}
\ead{satonaka@ncc.go.jp}

\author[n,a]{Nobuji Kouno}
\ead{nkouno@ncc.go.jp}

\author[b,a]{Amina Bolatkan}
\ead{abolatka@ncc.go.jp}

\author[d,c]{Yusuke Kurose}
\ead{kurose@mi.t.u-tokyo.ac.jp}

\author[d,c]{\\Tatsuya Harada}
\ead{harada@mi.t.u-tokyo.ac.jp}

\author[a,b]{Ryuji Hamamoto}
\ead{rhamamot@ncc.go.jp}

\address[a]{Division of Medical AI Research and Development, National Cancer Center Research Institute, \\5-1-1 Tsukiji, Chuo-ku, Tokyo 104-0045, Japan}
\address[b]{Cancer Translational Research Team, RIKEN Center for Advanced Intelligence Project, \\1-4-1 Nihonbashi, Chuo-ku, Tokyo 103-0027, Japan}
\address[c]{Machine Intelligence for Medical Engineering Team, RIKEN Center for Advanced Intelligence Project, \\1-4-1 Nihonbashi, Chuo-ku, Tokyo 103-0027, Japan}
\address[d]{Research Center for Advanced Science and Technology, The University of Tokyo, \\4-6-1 Komaba, Meguro-ku, Tokyo 153-8904, Japan}
\address[e]{Medical Data Deep Learning Team, Advanced Data Science Project, RIKEN Information R\&D and Strategy Headquarters, \\1-4-1 Nihonbashi, Chuo-ku, Tokyo 103-0027, Japan}
\address[f]{Department of Experimental Therapeutics, National Cancer Center Hospital, \\5-1-1 Tsukiji, Chuo-ku, Tokyo 104-0045, Japan}
\address[g]{Department of Diagnostic Radiology, National Cancer Center Hospital, \\5-1-1 Tsukiji, Chuo-ku, Tokyo 104-0045, Japan}
\address[h]{Department of Neurosurgery and Neuro-Oncology, National Cancer Center Hospital, \\5-1-1 Tsukiji, Chuo-ku, Tokyo 104-0045, Japan}
\address[i]{Department of Colorectal Surgery, National Cancer Center Hospital, \\5-1-1 Tsukiji, Chuo-ku, Tokyo 104-0045, Japan}
\address[j]{Department of Thoracic Surgery, National Cancer Center Hospital, \\5-1-1 Tsukiji, Chuo-ku, Tokyo 104-0045, Japan}
\address[k]{Radiation Safety and Quality Assurance Division, National Cancer Center Hospital, \\5-1-1 Tsukiji, Chuo-ku, Tokyo 104-0045, Japan}
\address[l]{Division of Research and Development for Boron Neutron Capture Therapy, National Cancer Center, Exploratory Oncology Research \& Clinical Trial Center, \\5-1-1 Tsukiji, Chuo-ku, Tokyo 104-0045, Japan}
\address[m]{Medical Physics Laboratory, Division of Health Science, Graduate School of Medicine, Osaka University \\Yamadaoka 1-7, Suita-shi, Osaka 565-0871, Japan}
\address[n]{Department of Surgery, Kyoto University Graduate School of Medicine \\54 Shogoin Kawahara-cho, Sakyo-ku, Kyoto 606-8507, Japan}

\date{December 2022}

\begin{abstract}
The amount of medical images stored in hospitals is increasing faster than ever; however, utilizing the accumulated medical images has been limited. This is because existing content-based medical image retrieval (CBMIR) systems usually require example images to construct query vectors; nevertheless, example images cannot always be prepared. Besides, there can be images with rare characteristics that make it difficult to find similar example images, which we call isolated samples. Here, we introduce a novel sketch-based medical image retrieval (SBMIR) system that enables users to find images of interest without example images. The key idea lies in feature decomposition of medical images, whereby the entire feature of a medical image can be decomposed into and reconstructed from normal and abnormal features. By extending this idea, our SBMIR system provides an easy-to-use two-step graphical user interface: users first select a template image to specify a normal feature and then draw a semantic sketch of the disease on the template image to represent an abnormal feature. Subsequently, it integrates the two kinds of input to construct a query vector and retrieves reference images with the closest reference vectors. Using two datasets, ten healthcare professionals with various clinical backgrounds participated in the user test for evaluation. As a result, our SBMIR system enabled users to overcome previous challenges, including image retrieval based on fine-grained image characteristics, image retrieval without example images, and image retrieval for isolated samples. Our SBMIR system achieves flexible medical image retrieval on demand, thereby expanding the utility of medical image databases.
\end{abstract}

\begin{keyword}
Sketch-based image retrieval\sep content-based image retrieval\sep feature decomposition\sep query by sketch\sep query by example
\end{keyword}

\abbreviations{2D, two-dimensional; 3D, three-dimensional; AC, anatomy code; CBMIR, content-based medical image retrieval; CT, computed tomography; ED, peritumoral edema; ET, gadolinium-enhancing tumor; FLAIR, fluid-attenuated inversion recovery sequence; GAN, generative adversarial network; Gd, gadolinium; GUI, graphical user interface; KL, Kullback--Leibler; MRI, magnetic resonance imaging; nDCG: normalized discounted cumulative gain; NET, necrotic and non-enhancing tumor core; NN, nearest neighbor; PT, primary tumor; SBMIR, sketch-based medical image retrieval; SVM, support vector machine; T1, T1-weighted sequence; T1CE, T1-weighted contrast-enhanced sequence; VAE, variational autoencoder}

\end{frontmatter}

\section{Introduction}\label{sec:introduction}

As the amount of medical images stored in hospital databases is increasing much faster than healthcare professionals can manage, the need for effective medical image retrieval has become greater owing to its potential in patient care, research, and development \citep{Quellec2011, Allan2012, Tschandl2020, Chen2022}. Indeed, healthcare professionals refer not only to evidence in the literature but also to the accumulated past cases because they often provide valuable insights into what differentiates individual patients from common characteristics in the patient population. Because the radiographic phenotype of a disease is closely related to its diagnosis, treatment response, and prognosis \citep{Aerts2014}, content-based medical image retrieval (CBMIR), which can calculate the similarity of medical images based on image contents, has been the mainstay for the development of medical image retrieval. 

\begin{figure*}[t!]
  \centering
  \includegraphics[]{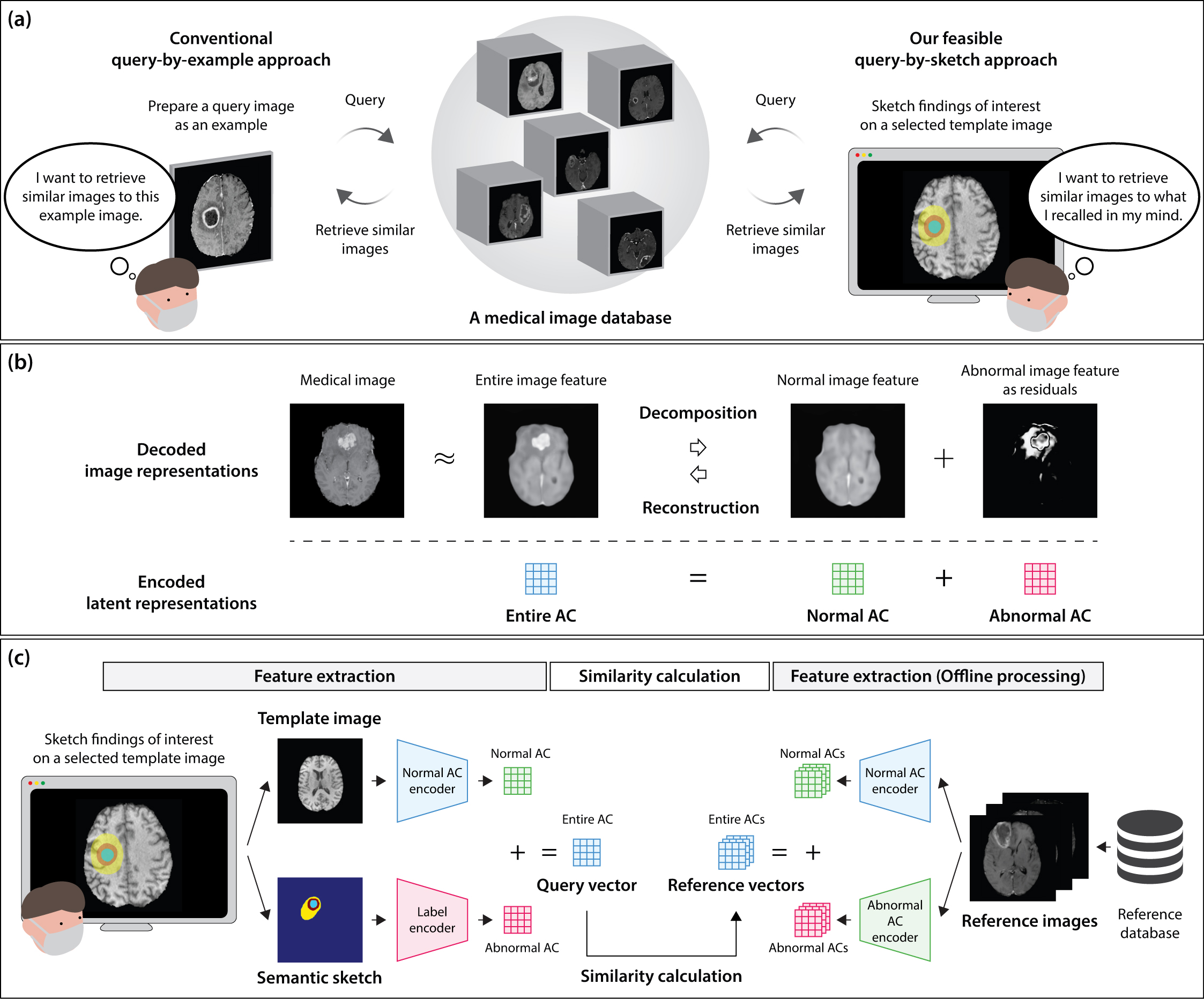}
  \caption{\textbf{Overview of the algorithm for the sketch-based medical image retrieval (SBMIR) system.} \textbf{a} The conventional \emph{query-by-example} approach in content-based medical image retrieval extracts a query vector from a query image as an example; nevertheless, it is not always easy to prepare an example image when users need to refer to images with specific findings. Here, we introduce a feasible \emph{query-by-sketch} approach in which users can specify a query vector by selecting a template image and by sketching the findings of interest onto the template image. \textbf{b} The technical basis is feature decomposition of medical images, where the entire feature (the entire anatomy code [AC]) of a medical image can be decomposed into and reconstructed from a normal feature (a normal AC) and an abnormal feature (an abnormal AC). \textbf{c} In our SBMIR system, a query vector is composed of a normal AC extracted from a selected template image and an abnormal AC approximated from a sketched label, while reference vectors are automatically computed from reference images. The reference images with the reference vector closest to the query vector are obtained as search results. AC, anatomy code.}
  \label{fig:query_by_sketch}
\end{figure*}

To date, CBMIR has been treated as a computer science problem to devise how to measure the similarity between medical images by capturing the unstructured nature of clinical findings. Based on a \emph{query-by-example} approach \citep{Pinho2019}, most CBMIR systems exploit an example image as a query image, from which a query vector representing requested information is extracted (see the left part of \textbf{\cref{fig:query_by_sketch}a}). As a search result, reference images with the closest reference vectors to the query vector are retrieved from a database. CBMIR emerged as a method to calculate the similarity based on visual features such as color, texture, shape, and spatial relationships among regions of interest \citep{Long2003, Li2018}. More recently, it has incorporated deep learning \citep{Lecun2015} to efficiently extract semantic features from example images in order to create query vectors \citep{Hosny2018}. This achieves better results by minimizing the semantic gap between high-level semantic concepts and low-level visual features in medical images \citep{Zheng2018}.

However, the information-seeking objectives of healthcare professionals can be diverse beyond the extent to which solely calculating the similarity between the query and reference images can handle. For example, to validate the clinical decision for a present case, healthcare professionals may need not only past cases with exactly the same clinical findings in the same anatomical location but also cases with similar findings in a different location or with different findings in the same location. Besides, suppose that one needs to search for images with particular clinical findings from medical image repositories on the Internet, which are becoming popular recently \citep{Prior2017}. In such cases, it is usually difficult for users to find the right image to initiate the search because they can not prepare an example image in advance. Furthermore, clinical medicine often puts high reference value on rare cases, including rare diseases and rare clinical findings \citep{Turro2020}. Nevertheless, it can be challenging for healthcare professionals to retrieve the relevant images from a database because similar example images are hardly available due to their rarity.

Based on these points of view, we focus on two potential limitations of conventional CBMIR systems using the query-by-example approach: \emph{usability} and \emph{searchability} problems. Usability is referred to as a qualitative attribute representing how easy it is for users to satisfy their information-seeking objectives whenever necessary. Since query-by-example approaches are dependent on example images, it is challenging to search in situations where no example images can be available in advance. Also, adding or subtracting arbitrary features to a query vector extracted from an example image is difficult, hindering the refinement of user queries for better results based on the search results. On the other hand, searchability indicates the scope of potentially retrievable images to all images stored in a database. Suppose a latent space where feature vectors corresponding to image characteristics are distributed. While a typical image can be surrounded by other typical images in the vicinity, a rare image may be located far away from the others due to its unique characteristics, which we call an \emph{isolated sample} in a database. The isolated sample might hardly be present in the nearest neighborhood of usually available example images, implying that the rarer the case, the more difficult its retrieval can be. By using ResNet-101 features \citep{He2016} (see \textbf{\ref{app:resnet_feature}}), we demonstrate the substantial number of isolated samples in medical image datasets in \textbf{\ref{app:isolated_samples}}. 

To mitigate these problems, one alternative could be a \emph{query-by-sketch} approach \citep{Sangkloy2016, Li2018Sketch, Dutta2019, Zhang2020, Vinker2022, Bhunia2022}, which has achieved notable success in the field of computer vision. A user's query sketch can convey the shape, orientation, and fine-grained details of objects without preparing example images. Nevertheless, sketch-based queries have not been demonstrated in medical image retrieval, perhaps because sketching all anatomical information seems too laborious for a real-world application. Therefore, a practical CBMIR system that does not require users to provide example images or to comprehensively sketch anatomical features is needed.

\subsection{SBMIR: sketch-based medical image retrieval system}\label{sec:sbmir}

Here, we introduce a feasible query-by-sketch approach that minimizes the effort required for sketching (see the right part of \textbf{\cref{fig:query_by_sketch}a}), which we used to establish the SBMIR system. The underlying assumption is as follows. Individual disease phenotypes are diverse and indefinite, whereas the surrounding normal anatomy shares many common features within a given population. Hence, if we can specify the semantic features of a disease with sketches and those of the surrounding normal anatomy by selecting a template image that approximates the normal anatomical features, we can construct a query vector conveying the target image content with its desired anatomical location. Based on this assumption, our SBMIR system uses two different modalities--\emph{a template image} and \emph{a semantic sketch of the disease}--to construct a query vector. 
The technological basis for constructing a query vector from these different modalities is the \emph{feature decomposition} of medical images \citep{Kobayashi2021}, whereby the entire feature vector of a medical image, which is referred to as an entire anatomy code (AC), can be decomposed into and reconstructed from two semantically different feature vectors, a normal AC and an abnormal AC (see \textbf{\cref{fig:query_by_sketch}b}). The relationship among the three ACs can be formulated as follows:
\begin{equation}\label{eq:decomposition}
\mathrm{Entire\,AC} = \mathrm{Normal\,AC} + \mathrm{Abnormal\,AC}.
\end{equation}
As demonstrated in \textbf{\cref{fig:query_by_sketch}b}, the normal AC represents the counterfactual normal anatomy that should have existed in the absence of an abnormality, and the abnormal AC represents any abnormality as a residual from the normal baseline. By extending this concept, we devised a deep-learning-based algorithm to extract a normal AC from a template image and an abnormal AC from a semantic sketch of the disease to construct a query vector by adding them together. 

\begin{figure*}[t!]
  \centering
  \includegraphics[]{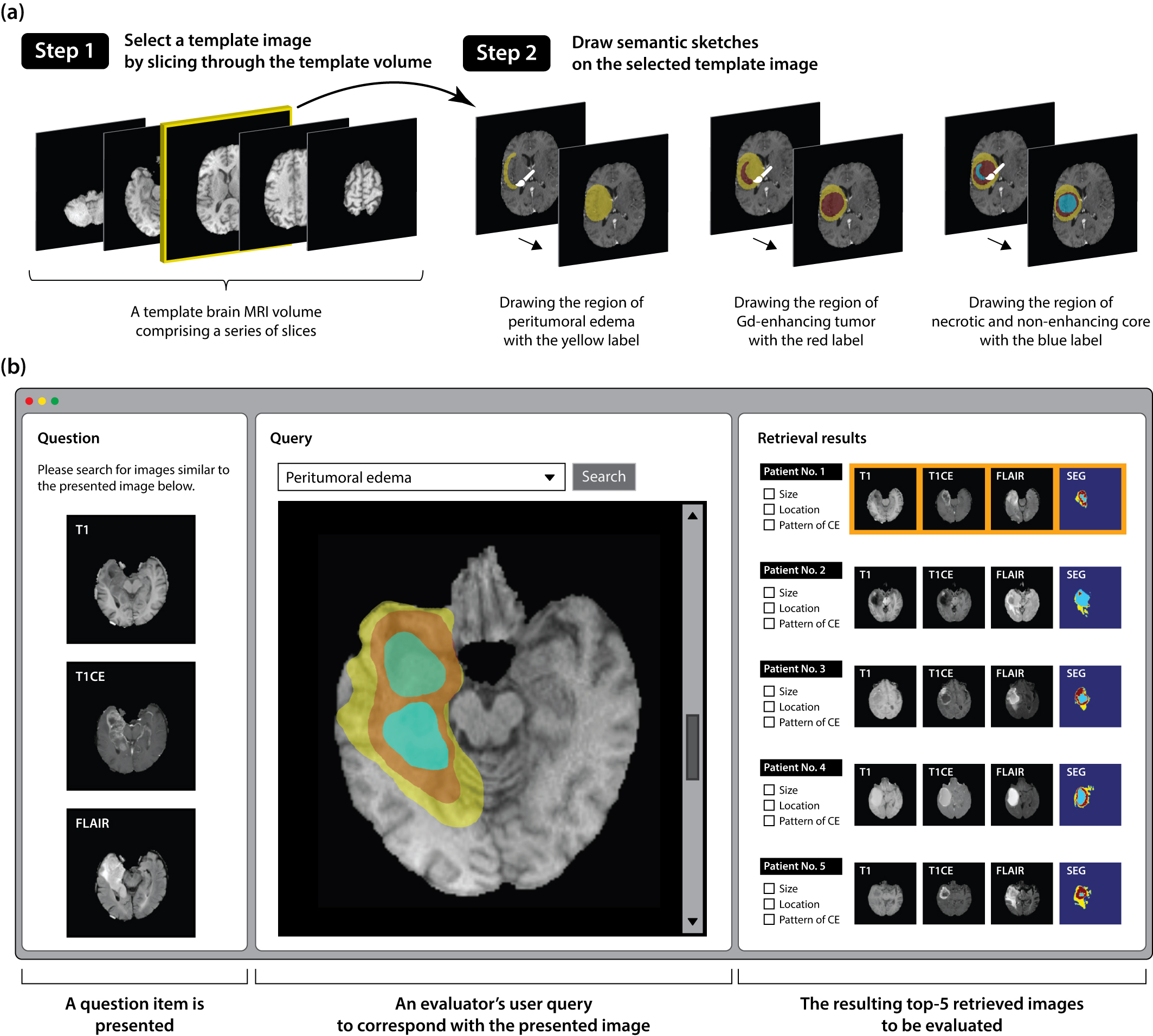}
  \caption{\textbf{Illustration of the software demonstrating our sketch-based medical image retrieval system.} \textbf{a} A user query is constructed through the following two-step user operation. First, a template image is selected to specify the location where a disease should exist, and second, a semantic segmentation label of the disease is sketched therein. \textbf{b} An easy-to-use graphical user interface was developed for the user tests. In the left field, an image is presented as a question item in Test-1. In the middle field, a ring-enhancing brain tumor surrounded by peritumoral edema located in the left temporal pole is drawn on a selected template image by an evaluator, the intention of which is to correspond with the presented image. The resulting top-5 retrieved images showing the corresponding reference images extracted from the testing dataset are listed in the right field. Each evaluator judges the consistency of the given criteria by checking the checkboxes. The orange highlighting of the retrieved images indicates that the image belongs to the same volume as the presented image (i.e., the same-volume image). CE, contrast enhancement; Gd, gadolinium; T1, T1-weighted sequence; T1CE, T1-weighted contrast-enhanced sequence; FLAIR, fluid-attenuated inversion recovery sequence; SEG, tumor-associated segmentation labels.}
  \label{fig:sbmir_system}
\end{figure*}

Our SBMIR system consists of a feature extraction module and a similarity calculation module (see \textbf{\cref{fig:query_by_sketch}c}). A query vector is calculated through the following two-step user operation (see the left part of \textbf{\cref{fig:query_by_sketch}c}). First, users select a two-dimensional (2D) template image by slicing through a three-dimensional (3D) template volume (see ``Step 1'' in \textbf{\cref{fig:sbmir_system}a} for an example). The 3D template volume comprises a series of 2D slices of a specific organ (e.g., brain) or anatomical region (e.g., chest). It is assumed that the variation in normal anatomy is limited to a small range in a given population, such that by selecting a slice as the template image that matches the area where users want the clinical findings to be present, users can specify the location of a disease. Second, users sketch semantic segmentation labels representing the disease on the selected template image (see ``Step 2'' in \textbf{\cref{fig:sbmir_system}a} for an example). These semantic segmentation labels are predefined and then learned for each specific clinical finding in medical images (e.g., a segmentation label representing a tumor region or a particular component of a disease, such as a necrotic region) as the image content that should be located therein. Then, the feature extraction module extracts a normal AC from the template image and an abnormal AC from the semantic sketch of the disease, both of which are summed to give the query vector according to \textbf{Eq. \ref{eq:decomposition}}. As a result, users can obtain reference images with the closest reference vector to the query vector, which is processed by the similarity calculation module (see the middle part of \textbf{\cref{fig:query_by_sketch}c}). These reference vectors are calculated in advance from reference images (see the right part of \textbf{\cref{fig:query_by_sketch}c}). Note that each of the retrieved top-$K$ similar images relative to the query vector belongs to different reference volumes (e.g., individual magnetic resonance imaging [MRI] and computed tomography [CT] scans) in order to avoid redundancy between consecutive 2D slices within a 3D volume.

For the model training and evaluation, a dataset consisting of brain MRI scans with gliomas and a dataset containing chest CT scans with lung cancers were used. To show that our SBMIR system is easy for healthcare professionals to use and that it can mitigate the usability and searchability problems of conventional CBMIR systems, 10 healthcare professionals with various clinical backgrounds participated in user tests based on a dedicated graphical user interface (GUI) (see \textbf{\cref{fig:sbmir_system}b}). The evaluators underwent a practice stage followed by three testing stages as follows: Test-1 demonstrated the image retrieval performance when example images were available, Test-2 revealed the image retrieval performance without example images, and Test-3 investigated the image retrieval performance for isolated samples. The \textbf{Supplementary Video 1} demonstrates our SBMIR system in action. Besides, we will soon release the source code and materials to show researchers how our SBMIR system works.

The main contributions of this study can be summarized into the following:
\begin{itemize}
  \item By extending the concept of the feature decomposition of medical images, we devised an algorithm to demonstrate the feasible query-by-sketch approach to construct a query vector, which requires neither an example image nor a detailed sketch of all the anatomical structures.
  \item We implemented the first SBMIR system that achieves flexible medical image retrieval on demand through an easy-to-use two-step user operation, the search results of which can change according to which template image is selected and how the disease is sketched.
  \item The user test showed that our SBMIR system could overcome the usability and searchability problems of conventional CBMIR systems through better image retrieval according to fine-grained image characteristics (Test-1), image retrieval without example images (Test-2), and image retrieval for isolated samples (Test-3). 
\end{itemize}

\section{Algorithm}\label{sec:algorithm}

Our SBMIR system has a unique feature extraction module at its core, which is based on feature decomposition of medical images extended with \emph{semantically organized latent space}. Such extension is critical to achieving the practical SBMIR system, which we will demonstrate in \textbf{\ref{app:eval_image_retrieval}}. Here, we describe the training algorithm of the feature extraction module, which is combined with the similarity calculation module to realize our real-time, large-scale image retrieval system. 

\begin{figure}[ht]
  \centering
  \includegraphics[]{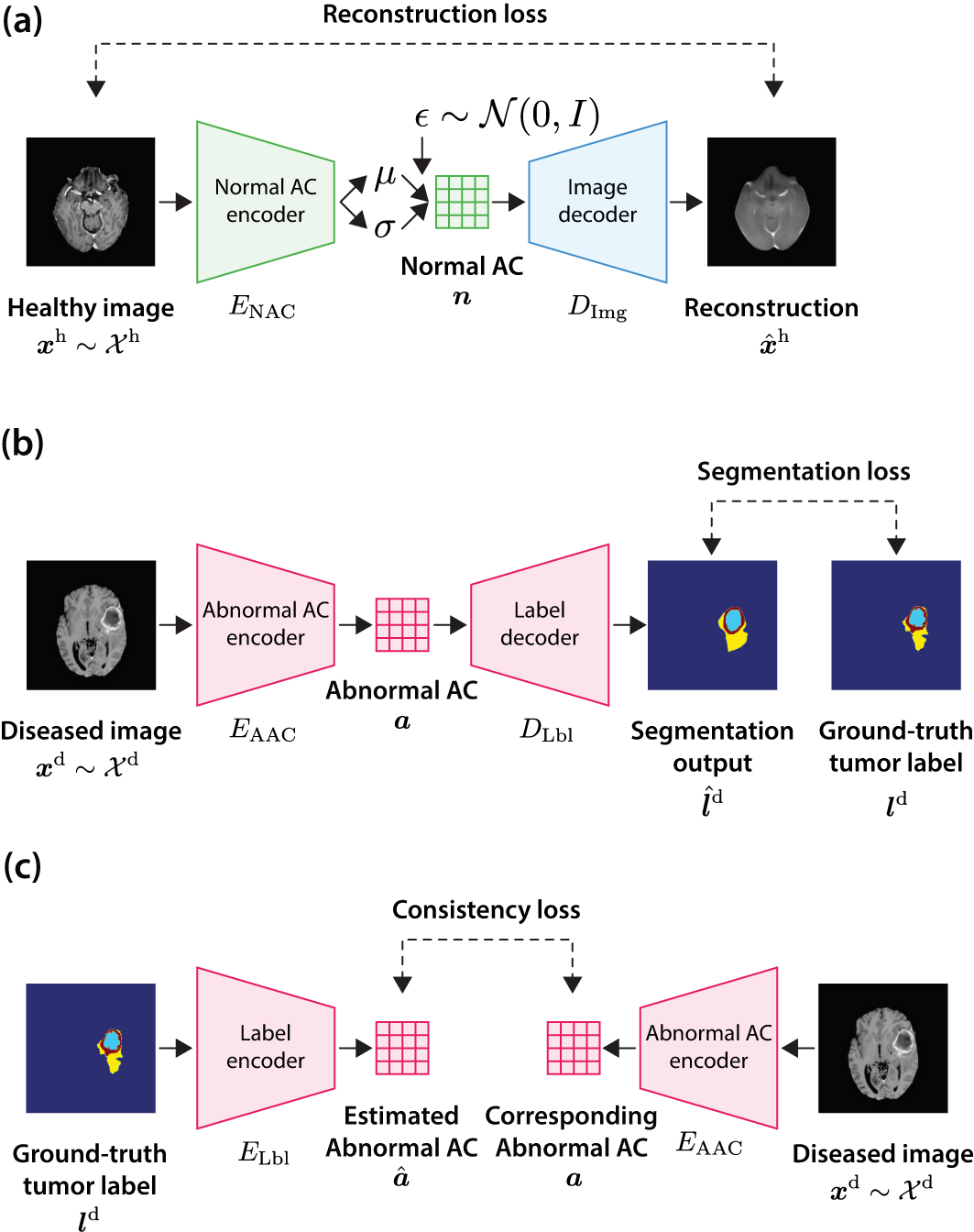}
  \caption{\textbf{Basic components of the feature extraction module.} The deep-learning framework for feature extraction consists of three parts. \textbf{a} A variational autoencoder, consisting of a normal anatomy code (AC) encoder and an image decoder, learns normal ACs using sub-batches comprising only healthy images. \textbf{b} An encoder--decoder type of segmentation network, consisting of an abnormal AC encoder and a label decoder, acquires abnormal ACs through the prediction of segmentation labels for the abnormal lesions. \textbf{c} An encoding network, consisting of a label encoder, maps ground-truth segmentation labels representing abnormal lesions in order to estimate the corresponding abnormal ACs that are output by the abnormal AC encoder using a corresponding diseased image as input.}
  \label{fig:basic_components}
\end{figure}

\subsection{Network architecture for feature decomposition}\label{sec:architect_feat_decomp}

The feature extraction module of our SBMIR system is a deep-learning framework that can perform feature decomposition of medical images \citep{Kobayashi2021} with extended mapping functions between the image space $\mathcal{X}$ and the label space $\mathcal{L}$ via the latent space $\mathcal{Z}$ ($\mathcal{X} \leftrightarrow \mathcal{Z} \leftrightarrow \mathcal{L}$). The training of the neural networks requires a dataset consisting of pairs of an image and a corresponding segmentation label of tumor-associated regions $(\bm{x}, \bm{l}) \in \mathcal{X} \times \mathcal{L}$. Below, we assume that the image space $\mathcal{X}$ has a subspace corresponding to healthy images $\mathcal{X}^\mathrm{h}$ and another subspace corresponding to diseased images $\mathcal{X}^\mathrm{d}$. Although feature decomposition of medical images often involves elaborate algorithms such as conditional generation using generative adversarial networks (GANs) \citep{Goodfellow2014, Liu2022}, our SBMIR system achieves this objective by combining three simple components, without utilizing adversarial training, to stabilize the training process.

\subsubsection{Variational auto-encoder for learning normal ACs}\label{sec:vae_component}

As illustrated in \textbf{\cref{fig:basic_components}a}, the first component of our feature extraction module is a variational autoencoder (VAE) \citep{Kingma2013, Rezende2014} that learns the mapping function of the normal ACs $\bm{n}$ between the image space and the latent space ($\mathcal{X} \xrightarrow{\bm{n}} \mathcal{Z} \xrightarrow{\bm{n}} \mathcal{X}$). This VAE consists of a pair of a normal AC encoder $E_\mathrm{NAC}$ and an image decoder $D_\mathrm{Img}$, which together take a healthy image $\bm{x}^\mathrm{h} \sim \mathcal{X}^\mathrm{h}$ as input and output a reconstructed image $\hat{\bm{x}}^\mathrm{h}$ by simultaneously producing a normal AC $\bm{n}$ in the latent space ($D_\mathrm{Img}(E_\mathrm{NAC} (\bm{x}^\mathrm{h})) = D_\mathrm{Img}(\bm{n}) = \hat{\bm{x}}^\mathrm{h}$). Based on the idea that the distribution of normal ACs $\bm{n}$ should be within a certain range reflecting the limited range of normal anatomic variation within a population, we impose an isotropic multivariate Gaussian $\mathcal{N}(\bm{n}; \bm{0}, \bm{I})$ as a prior distribution over the normal ACs ($p(\bm{n}) = \mathcal{N}(\bm{0}, \bm{I})$). As such, using the two output variables, $\bm{\mu}$ and $\bm{\sigma}$, a posterior distribution estimated by the encoder $E_\mathrm{NAC} (\bm{n} \rvert \bm{x}^\mathrm{h}) = \mathcal{N}(\bm{\mu} (\bm{x}^\mathrm{h}), \bm{\sigma} (\bm{x}^\mathrm{h}))$ is forced to be close to the prior by minimizing the Kullback--Leibler (KL)-divergence between the prior and the posterior distributions. 

\subsubsection{Segmentation network for learning abnormal ACs}\label{sec:segmentaion_network}

\textbf{\Cref{fig:basic_components}b} shows the second component of our feature extraction module, an encoder--decoder type of segmentation network that learns the mapping function for the abnormal ACs $\bm{a}$ alongside the sequence $\mathcal{X} \xrightarrow{\bm{a}} \mathcal{Z} \xrightarrow{\bm{a}} \mathcal{L}$ based on a pair of a diseased image $\bm{x}^\mathrm{d} \sim \mathcal{X}^\mathrm{d}$ and a segmentation label of tumor-associated regions $\bm{l}^\mathrm{d}$. The abnormal ACs $\bm{a}$ are acquired as the outputs of an abnormal AC encoder $E_\mathrm{AAC}$ and are decoded to a semantic segmentation label $\hat{\bm{l}}^\mathrm{d}$ through a label decoder $D_\mathrm{Lbl}$, as follows: $D_\mathrm{Lbl}(E_\mathrm{AAC}(\bm{x}^\mathrm{d})) = D_\mathrm{Lbl}(\bm{a}) = \hat{\bm{l}}^\mathrm{d}$. Note that there is no skip connection between the abnormal AC encoder $E_\mathrm{AAC}$ and the label decoder $D_\mathrm{Lbl}$. Therefore, through training using segmentation losses between $\hat{\bm{l}}^\mathrm{d}$ and $\bm{l}^\mathrm{d}$, the abnormal ACs $\bm{a}$ can be optimized to encode the semantic features particularly relevant to the tumor-associated regions. 

\subsubsection{Encoding network to estimate abnormal ACs from semantic segmentation labels}\label{sec:label_encoder}

\textbf{\Cref{fig:basic_components}c} describes the third component of our feature extraction module, a label encoder $E_\mathrm{Lbl}$ that enables the mapping function from the label space to the latent space with respect to the abnormal ACs $\bm{a}$ ($\mathcal{L} \xrightarrow{\bm{a}} \mathcal{Z}$). Specifically, the label encoder $E_\mathrm{Lbl}$ estimates an abnormal AC $\hat{\bm{a}}$ from a ground-truth segmentation label $\bm{l}^\mathrm{d}$ as input ($E_\mathrm{Lbl} (\bm{l}^\mathrm{d}) = \hat{\bm{a}}$), which is an inverse function of the label decoder ($D_\mathrm{Lbl} (\bm{a}) = \hat{\bm{l}}^\mathrm{d}$). In the training of the label encoder $E_\mathrm{Lbl}$, the corresponding abnormal AC $\bm{a}$ that gives the closest segmentation prediction to the ground-truth label through the label decoder $D_\mathrm{Lbl}$ can be the ground-truth for the estimated abnormal AC $\hat{\bm{a}}$. 

\subsubsection{The whole network architecture}\label{sec:whole_architect}

\begin{figure*}[t!]
  \centering
  \includegraphics[]{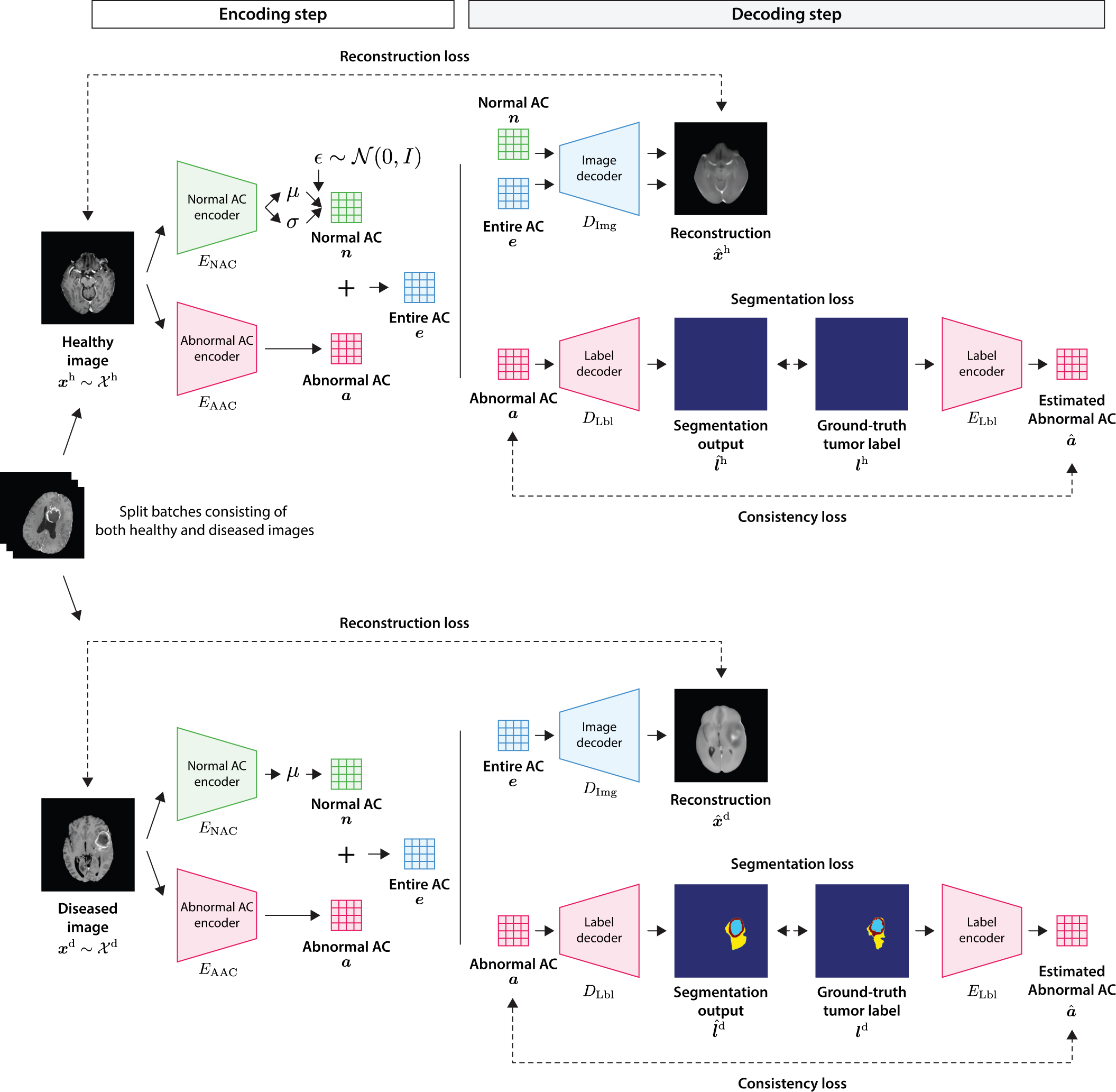}
  \caption{\textbf{Overall architecture of the feature extraction module for our sketch-based medical image retrieval system.} The deep-learning framework can learn normal anatomy codes (ACs), abnormal ACs, and entire ACs according to each semantic concept and mapping function between the image space and the label space via the latent space. The training is performed differently, depending on whether the input images are healthy or diseased. Particularly, the VAE component is switched to be trainable when healthy images are given, whereas its inference results are utilized when diseased images are given. This is reasonable because there are no ground-truth pseudo-normal images corresponding to the diseased images.}
  \label{fig:whole_architecture}
\end{figure*}

By combining these components, the deep-learning framework for the feature extraction module in our SBMIR system is trained as illustrated in \textbf{\cref{fig:whole_architecture}}. The training process is changed according to whether the input image is healthy $\bm{x}^\mathrm{h} \sim \mathcal{X}^\mathrm{h}$ or diseased $\bm{x}^\mathrm{d} \sim \mathcal{X}^\mathrm{d}$. Notably, the VAE component is trained only when healthy images $\bm{x}^\mathrm{h}$ are given as input, while its inference results are used when diseased images $\bm{x}^\mathrm{d}$ are given. 

When a healthy image $\bm{x}^\mathrm{h}$ is given as input, the VAE component is switched to be trainable (see the upper part of \textbf{\cref{fig:whole_architecture}}). At the encoding step, from the healthy image $\bm{x}^\mathrm{h}$, the normal AC encoder $E_\mathrm{NAC}$ estimates the posterior distribution of the normal AC ($E_\mathrm{NAC} (\bm{n} \rvert \bm{x}^\mathrm{h}) = \mathcal{N}(\bm{\mu} (\bm{x}^\mathrm{h}), \bm{\sigma} (\bm{x}^\mathrm{h}))$) and the abnormal AC encoder $E_\mathrm{AAC}$ outputs the abnormal AC $\bm{a}$ in the latent space ($E_\mathrm{AAC} (\bm{x}^\mathrm{h}) = \bm{a}$). A normal AC $\bm{n}$ can be sampled from $\mathcal{N}(\bm{n}; \bm{\mu} (\bm{x}^\mathrm{h}), \bm{\sigma} (\bm{x}^\mathrm{h}))$ using a reparameterization trick \citep{Kingma2013, Rezende2014}, that is, $\bm{n} = \bm{\mu} + \bm{\sigma} \odot \bm{\epsilon}$, where $\bm{\epsilon} \sim \mathcal{N}(\bm{0}, \bm{I})$ and $\odot$ indicates the Hadamard product. Then, an entire AC $\bm{e}$ is calculated as the sum of the normal AC $\bm{n}$ and the abnormal AC $\bm{a}$ ($\bm{e} = \bm{n} + \bm{a}$). Note that since the input image does not include any abnormality, the abnormal AC $\bm{a}$ is trained to be the zero vector ($\bm{a} \approx \bm{0}$) so as not to convey any abnormal information about the image, making the entire AC $\bm{e}$ and the normal AC $\bm{n}$ identical as a result ($\bm{e} \approx \bm{n}$). At the decoding step, the normal AC $\bm{n}$ and the entire AC $\bm{e}$ are independently fed into the image decoder $D_\mathrm{Img}$ to reconstruct the same healthy input image $\hat{\bm{x}}^\mathrm{h}$ ($D_\mathrm{Img} (\bm{n}) = \hat{\bm{x}}^\mathrm{h}$ and $D_\mathrm{Img} (\bm{e}) = \hat{\bm{x}}^\mathrm{h}$). Additionally, the abnormal AC $\bm{a}$, which is trained to be the zero vector, is taken by the label encoder $D_\mathrm{Lbl}$ as input to generate a segmentation label $\hat{\bm{l}}^\mathrm{h}$ that is encouraged to be similar to a ground-truth label $\bm{l}^\mathrm{h}$ filled with zeros ($D_\mathrm{Lbl} (\bm{a}) = \hat{\bm{l}}^\mathrm{h}$). Finally, the ground-truth label $\bm{l}^\mathrm{h}$ is fed into the label encoder $E_\mathrm{Lbl}$ that is trained to estimate the corresponding abnormal AC $\hat{\bm{a}}$, which should be the zero vector as well ($E_\mathrm{Lbl} (\bm{l}^\mathrm{h}) = \hat{\bm{a}}$). 

Conversely, when a diseased image $\bm{x}^\mathrm{d}$ is given as input, the VAE component is not used for learning, and only its inference results are utilized (see the lower part of \textbf{\cref{fig:whole_architecture}}). The assumption is that the normal ACs $\bm{n}$, which are trained only on the healthy images $\bm{x}^\mathrm{h}$ to encode the normal anatomical information, should be incapable of reconstructing abnormal lesions in the diseased images $\bm{x}^\mathrm{d}$ \citep{Schlegl2017}. Thus, the inference results from the diseased image $\bm{x}^\mathrm{d}$ can be the normal ACs $\bm{n}$ that match a pseudo-normal image corresponding to the input image. The rest of the training process is essentially analogous to the process starting with a healthy image. At the encoding step, from the diseased image $\bm{x}^\mathrm{d}$, the normal AC encoder $E_\mathrm{NAC}$ infers the normal AC $\bm{n}$ ($E_\mathrm{NAC} (\bm{x}^\mathrm{d}) = \bm{n}$), and the abnormal AC encoder $E_\mathrm{AAC}$ outputs the abnormal AC $\bm{a}$ ($E_\mathrm{AAC} (\bm{x}^\mathrm{d}) = \bm{a}$). Then, the entire AC $\bm{e}$ is calculated as the sum of the normal AC $\bm{n}$ and the abnormal AC $\bm{a}$ ($\bm{e} = \bm{n} + \bm{a}$). At the decoding step, the entire AC $\bm{e}$ is fed into the image decoder $D_\mathrm{Img}$ to reconstruct the whole input image with abnormal findings $\hat{\bm{x}}^\mathrm{d}$ ($D_\mathrm{Img} (\bm{e}) = \hat{\bm{x}}^\mathrm{d}$). Note that reconstruction from the normal AC $\bm{n}$ is not performed because there is no ground-truth for the pseudo-normal image corresponding to the input image. Then, the abnormal AC $\bm{a}$ is fed into the label decoder $D_\mathrm{Lbl}$ to predict the segmentation label of the abnormalities $\hat{\bm{l}}^\mathrm{d}$ that should be matched to the ground-truth segmentation label $\bm{l}^\mathrm{d}$ ($D_\mathrm{Lbl} (\bm{a}) = \hat{\bm{l}}^\mathrm{d}$). Finally, the ground-truth segmentation label $\bm{l}^\mathrm{d}$ is taken as input by the label encoder $E_\mathrm{Lbl}$ to estimate the corresponding abnormal AC $\hat{\bm{a}}$ ($E_\mathrm{Lbl} (\bm{l}^\mathrm{d}) = \hat{\bm{a}}$).

\subsection{Creating the semantically organized latent space}\label{sec:semantic_latent}

\begin{figure*}[ht]
  \centering
  \includegraphics[]{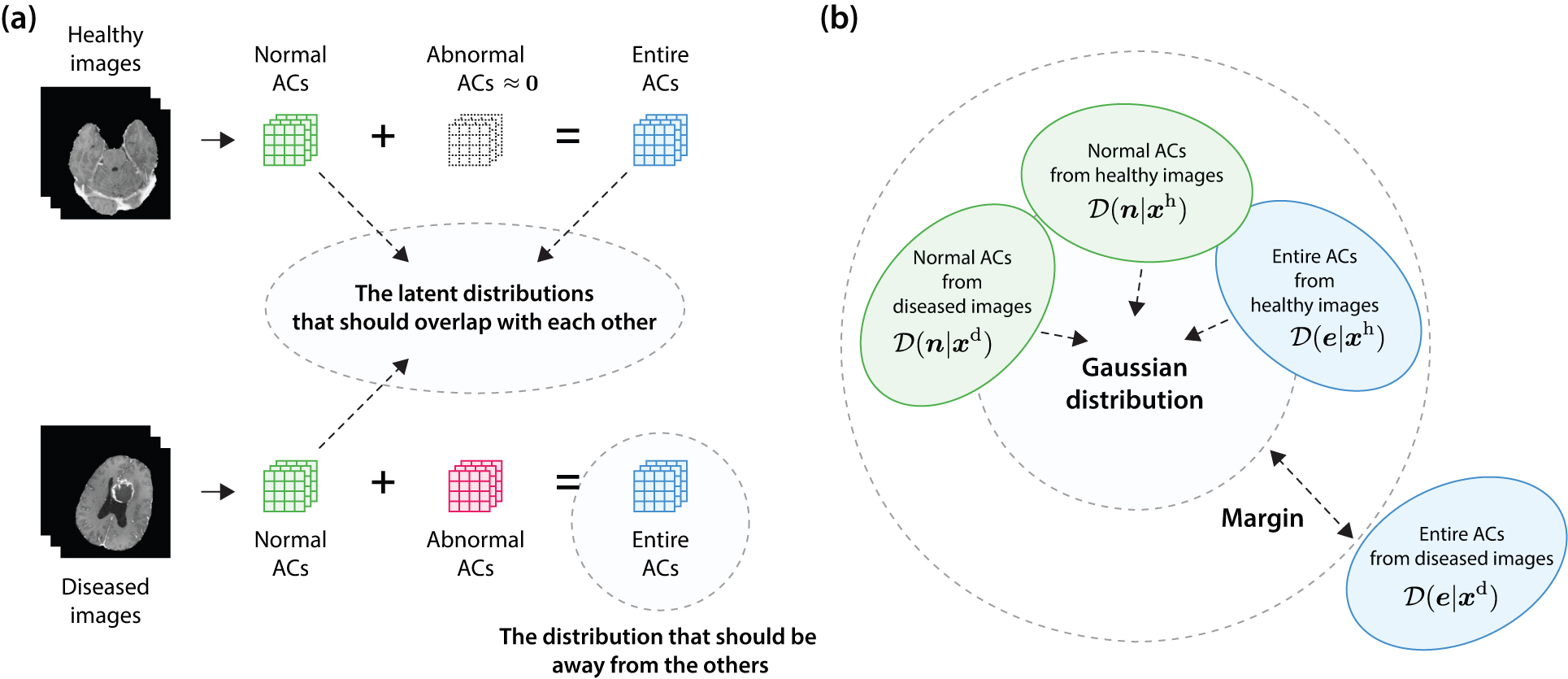}
  \caption{\textbf{Semantically organized latent space.} \textbf{a} The abnormal anatomy codes (ACs) extracted from healthy images should be close to zero vectors, such that the normal ACs and entire ACs should be identical. On the other hand, the abnormal ACs extracted from diseased images should contain meaningful information, such that the normal ACs and the entire ACs are distributed separately. \textbf{b} The normal ACs from healthy images, the entire ACs from healthy images, and the normal ACs from diseased images should overlap with each other, whereas the entire ACs from diseased images should be far away from the others by introducing a margin parameter.}
  \label{fig:latent_space}
\end{figure*}

The semantically organized latent space is critical for our SBMIR system to perform image retrieval based on semantics (i.e., whether an image is healthy or diseased). That is, a query vector conveying the information of a diseased region should retrieve only diseased images $\bm{x}^\mathrm{d}$ and that not conveying any disease region information should retrieve only healthy images $\bm{x}^\mathrm{h}$. We call this \emph{semantic consistency}, which will be quantitatively evaluated later in \textbf{\ref{app:semantic_consistency}}. To achieve this, the latent space $\mathcal{Z}$ needs to be separated into a subspace representing healthy images $\mathcal{Z}^\mathrm{h}$ (i.e., \emph{the healthy subspace}), and another subspace representing diseased images $\mathcal{Z}^\mathrm{d}$ (i.e., \emph{the diseased subspace}). In other words, our SBMIR system should enable the corresponding mapping based on semantics, between the image space and the latent space as follows: $\mathcal{X}^\mathrm{h} \leftrightarrow \mathcal{Z}^\mathrm{h}$ and $\mathcal{X}^\mathrm{d} \leftrightarrow \mathcal{Z}^\mathrm{d}$. Here, we explain how to configure the semantically organized latent space. 

\subsubsection{Four latent distributions in the latent space}\label{sec:four_latent_distrib}

How the information conveyed by an abnormal AC $\bm{a}$ changes according to the semantics is essential for configuring the semantically organized latent space. As shown in \textbf{\cref{fig:latent_space}a}, the entire AC $\bm{e}$ of a diseased image $\bm{x}^\mathrm{d}$ is represented as a sum of the normal AC $\bm{n}$ and the abnormal AC $\bm{a}$ ($\bm{e} = \bm{n} + \bm{a}$); in contrast, that of a healthy image $\bm{x}^\mathrm{h}$ can be approximated only by the normal AC $\bm{n}$ ($\bm{e} \approx \bm{n}$), as the abnormal AC $\bm{a}$ should be the zero vector reflecting the absence of abnormality ($\bm{a} \approx \bm{0}$). Therefore, there should be four latent distributions in the latent space: a distribution of entire ACs from healthy images $\mathcal{D}(\bm{e} \rvert \bm{x}^\mathrm{h})$, that of normal ACs from healthy images $\mathcal{D}(\bm{n} \rvert \bm{x}^\mathrm{h})$, that of entire ACs from diseased images $\mathcal{D}(\bm{e} \rvert \bm{x}^\mathrm{d})$, and that of normal ACs from diseased images $\mathcal{D}(\bm{n} \rvert \bm{x}^\mathrm{d})$. Considering the semantics, the healthy subspace $\mathcal{Z}^\mathrm{h}$ should enclose the distribution of entire ACs from healthy images $\mathcal{D}(\bm{e} \rvert \bm{x}^\mathrm{h})$, that of normal ACs from healthy images $\mathcal{D}(\bm{n} \rvert \bm{x}^\mathrm{h})$, and that of normal ACs from diseased images $\mathcal{D}(\bm{n} \rvert \bm{x}^\mathrm{d})$ ($\{ \mathcal{D}(\bm{e} \rvert \bm{x}^\mathrm{h}), \mathcal{D}(\bm{n} \rvert \bm{x}^\mathrm{h}), \mathcal{D}(\bm{n} \rvert \bm{x}^\mathrm{d}) \} \subset \mathcal{Z}^\mathrm{h}$), because all of these distributions represent healthy images. On the other hand, the diseased subspace $\mathcal{Z}^\mathrm{d}$ includes the remaining distribution of entire ACs from diseased images $\mathcal{D}(\bm{e} \rvert \bm{x}^\mathrm{d})$ ($\{\mathcal{D}(\bm{e} \rvert \bm{x}^\mathrm{d})\} \subset \mathcal{Z}^\mathrm{d}$) because only this distribution represents diseased images. Note that since the abnormal ACs $\bm{a}$ express only the amount of change from the healthy subspace $\mathcal{Z}^\mathrm{h}$ to the diseased subspace $\mathcal{Z}^\mathrm{d}$ (i.e., as shown in \textbf{\cref{fig:query_by_sketch}b}, an abnormal AC does not have enough information to reconstruct a whole image), the distributions of the abnormal ACs $\bm{a}$ are not taken into account.

\subsubsection{Configuration of the healthy subspace}\label{sec:config_healthy_subspace}

\textbf{\Cref{fig:latent_space}b} illustrates how the three distributions in the healthy subspace ($\{ \mathcal{D}(\bm{e} \rvert \bm{x}^\mathrm{h}), \mathcal{D}(\bm{n} \rvert \bm{x}^\mathrm{h}), \mathcal{D}(\bm{n} \rvert \bm{x}^\mathrm{d}) \} \subset \mathcal{Z}^\mathrm{h}$) are trained to overlap with each other. During the model training, the three distributions in the healthy subspace $\mathcal{Z}^\mathrm{h}$ are forced to follow the isotropic multivariate Gaussian distribution $\mathcal{N}(\bm{n}; \bm{0}, \bm{I})$ that is formulated by the VAE component to learn normal ACs $\bm{n}$. In particular, the distribution of normal ACs from healthy images $\mathcal{D}(\bm{n} \rvert \bm{x}^\mathrm{h})$ is directly optimized to follow the Gaussian distribution during the training of the VAE component (see the process starting with a healthy image in \textbf{\cref{fig:whole_architecture}}). Additionally, the distribution of normal ACs from diseased images $\mathcal{D}(\bm{n} \rvert \bm{x}^\mathrm{d})$ is ensured to follow the Gaussian distribution because each normal AC $\bm{n}$ from diseased images $\bm{x}^\mathrm{d}$ is sampled from the posterior distribution of the VAE component as an inference result (see the process starting with a diseased image in \textbf{\cref{fig:whole_architecture}}). The distribution of entire ACs from healthy images $\mathcal{D}(\bm{e} \rvert \bm{x}^\mathrm{h})$ can be indirectly optimized to follow the Gaussian distribution by forcing the abnormal ACs extracted from healthy images to the zero vector ($\bm{a} \approx \bm{0}$), as follows: $\mathcal{D}(\bm{e} \rvert \bm{x}^\mathrm{h}) = \mathcal{D}(\bm{n} + \bm{a} \rvert \bm{x}^\mathrm{h}) \approx \mathcal{D}(\bm{n} \rvert \bm{x}^\mathrm{h})$. Here, $\mathcal{D}(\bm{n} \rvert \bm{x}^\mathrm{h})$ is trained to be the Gaussian distribution by the learning objective of the VAE component. 

\subsubsection{Configuration of the diseased subspace}\label{sec:config_diseased_subspace}

\textbf{\Cref{fig:latent_space}b} also depicts how the distribution of entire ACs from diseased images in the diseased subspace ($\{\mathcal{D}(\bm{e} \rvert \bm{x}^\mathrm{d})\} \subset \mathcal{Z}^\mathrm{d}$) should be separated from the Gaussian distribution $\mathcal{N}(\bm{0}, \bm{I})$ that represents healthy images $\bm{x}^\mathrm{h}$. To achieve this, a \emph{margin parameter} is provided as a hyperparameter that determines how far apart the two subspaces should be, which is indicated by an arrow labeled ``Margin'' in \textbf{\cref{fig:latent_space}b}. The distance between the distribution of entire ACs from diseased images $\mathcal{D}(\bm{e} \rvert \bm{x}^\mathrm{d})$ and the Gaussian distribution $\mathcal{N}(\bm{0}, \bm{I})$ is measured by KL-divergence, and it is optimized to exceed the margin parameter, which is formulated as a margin loss as follows:
\begin{equation}
L_\mathrm{margin} = \max (0, m - \mathrm{KL}(\mathcal{D}(\bm{e} \rvert \bm{x}^\mathrm{d}) \| \mathcal{N}(\bm{0}, \bm{I}))).
\label{eq:margin}
\end{equation}
Here, $m$ is the margin parameter, and $\mathrm{KL}(\cdot \| \cdot)$ is the KL-divergence between two distributions. As the margin parameter increases, the distance, which is represented as the KL-divergence, between the two distributions is configured to be large.

\subsection{Learning objectives}\label{sec:learning_objectives}

To train the deep-learning framework, several loss functions are defined. Reconstruction loss $L_\mathrm{recon}$ is a composite loss function of both a perceptual loss function using the VGG network \citep{Simonyan2014} and a mean squared loss function, which forces the reconstructed images $\hat{\bm{x}}$ to be similar to the input images $\bm{x}$. When necessary, the reconstruction loss focusing on the tumor-associated regions is additionally calculated for the entire reconstruction to force the model to generate the abnormal regions more precisely (see \textbf{\cref{sec:hyparam_lung_cancer_dataset}}). Segmentation loss $L_\mathrm{seg}$ is a composite loss function combining a Dice loss function \citep{Dice1945} and a focal loss function \citep{Lin2017} between the predicted segmentation label for abnormal areas $\hat{\bm{l}}$ and its ground-truth label $\bm{l}$. Consistency loss $L_\mathrm{cons}$ calculates the L1 distance between the estimated abnormal ACs $\hat{\bm{a}}$ by the label encoder $E_\mathrm{Lbl}$ and the corresponding abnormal ACs $\bm{a}$, forcing the estimated abnormal ACs $\hat{\bm{a}}$ to be close to the corresponding abnormal ACs $\bm{a}$. Abnormality loss $L_\mathrm{abn}$ works only when a healthy image is given to force the norm of the abnormal ACs $\bm{a}$ to be zero. Regularization loss $L_\mathrm{reg}$ works only when a healthy image is given, matching the posterior distribution estimated by the encoder $E_\mathrm{NAC} (\bm{n} \rvert \bm{x}^\mathrm{h}) = \mathcal{N}(\bm{\mu} (\bm{x}^\mathrm{h}), \bm{\sigma} (\bm{x}^\mathrm{h}))$ and the prior distribution $\mathcal{N}(\bm{0}, \bm{I})$ by minimizing the KL-divergence between them, as follows:
\begin{equation}
L_\mathrm{reg} = \mathrm{KL} (\mathcal{N}(\bm{\mu} (\bm{x}^\mathrm{h}), \bm{\sigma} (\bm{x}^\mathrm{h}))\| \mathcal{N}(\bm{0}, \bm{I})).
\label{eq:reg}
\end{equation}
Finally, margin loss $L_\mathrm{margin}$ works only when a diseased image is given, imposing a distance between the distribution of the entire ACs from diseased images $\mathcal{D}(\bm{e} \rvert \bm{x}^\mathrm{d})$ and the Gaussian distribution $\mathcal{N}(\bm{0}, \bm{I})$, as formulated in \textbf{Eq. \ref{eq:margin}}. In the training process, the weighted sum of the abovementioned loss functions was set to be minimized:
\begin{equation}
\begin{split}
L_\mathrm{total} &= w_\mathrm{recon} L_\mathrm{recon} \\
                 &+ w_\mathrm{seg} L_\mathrm{seg} \\
                 &+ w_\mathrm{cons} L_\mathrm{cons} \\
                 &+ w_\mathrm{abn} L_\mathrm{abn} \\
                 &+ w_\mathrm{reg} L_\mathrm{reg} \\
                 &+ w_\mathrm{margin} L_\mathrm{margin}.
\label{eq:total}
\end{split}
\end{equation}
Here, $w_\mathrm{recon}$, $w_\mathrm{seg}$, $w_\mathrm{cons}$, $w_\mathrm{abn}$, $w_\mathrm{reg}$, and $w_\mathrm{margin}$ are weights for the reconstruction loss $L_\mathrm{recon}$, segmentation loss $L_\mathrm{seg}$, consistency loss $L_\mathrm{cons}$, abnormality loss $L_\mathrm{abn}$, regularization loss $L_\mathrm{reg}$, and margin loss $L_\mathrm{margin}$, respectively. The full algorithm for training the feature extraction module of our SBMIR system is summarized in \textbf{Algorithm \ref{alg:learning_algorithm}}. 

\begin{algorithm}[t]
\small
\SetAlgoLined
sg: a stop-gradient operator\\
\While{not converge}{
\tcc{Forward path for healthy images}
Sample a batch of healthy images $\bm{x}^\mathrm{h}$ and corresponding segmentation labels $\bm{l}^\mathrm{h}$ from the training dataset. \\
$\bm{\mu}, \bm{\sigma} \leftarrow E_\mathrm{NAC}(\bm{x}^\mathrm{h})$\\
$\bm{n} = \bm{\mu} + \bm{\sigma} \odot \bm{\epsilon}, \bm{\epsilon} \sim \mathcal{N}(\bm{0}, \bm{I})$\\
$\bm{a} = E_\mathrm{AAC}(\bm{x}^\mathrm{h})$\\
$\bm{e} = \bm{n} + \bm{a}$\\
$\hat{\bm{x}}^\mathrm{h} = D_\mathrm{Img} (\bm{e})$\\
$\hat{\bm{x}}^\mathrm{h} = D_\mathrm{Img} (\bm{n})$\\
$\hat{\bm{l}}^\mathrm{h} = D_\mathrm{Lbl} (\bm{a})$\\
$\hat{\bm{a}} = E_\mathrm{Lbl} (\bm{l}^\mathrm{h})$\\
Compute $L_\mathrm{recon} (\hat{\bm{x}}^\mathrm{h}, \bm{x}^\mathrm{h})$, $L_\mathrm{seg} (\hat{\bm{l}}^\mathrm{h}, \bm{l}^\mathrm{h})$, $L_\mathrm{cons} (\hat{\bm{a}}, \mathrm{sg}(\bm{a}))$, $L_\mathrm{abn} (\bm{a})$, and $L_\mathrm{reg} (\bm{\mu}, \bm{\sigma})$.\\
Update parameters of $E_\mathrm{NAC}$, $E_\mathrm{AAC}$, $E_\mathrm{Lbl}$, $D_\mathrm{Img}$, and $D_\mathrm{Lbl}$ to minimize $w_\mathrm{recon} L_\mathrm{recon} + w_\mathrm{seg} L_\mathrm{seg} + w_\mathrm{cons} L_\mathrm{cons} + w_\mathrm{abn} L_\mathrm{abn} + w_\mathrm{reg} L_\mathrm{reg}$ using stochastic gradient descent (e.g., Adam). \\

\tcc{Forward path for diseased images}
Sample a batch of diseased images $\bm{x}^\mathrm{d}$ and corresponding segmentation labels $\bm{l}^\mathrm{d}$ from the training dataset. \\
$\bm{\mu} \leftarrow E_\mathrm{NAC}(\bm{x}^\mathrm{d})$\\
$\bm{n} \leftarrow \mathrm{sg}(\bm{\mu})$\\
$\bm{a} = E_\mathrm{AAC}(\bm{x}^\mathrm{d})$\\
$\bm{e} = \bm{n} + \bm{a}$\\
$\hat{\bm{x}}^\mathrm{d} = D_\mathrm{Img} (\bm{e})$\\
$\hat{\bm{l}}^\mathrm{d} = D_\mathrm{Lbl} (\bm{a})$\\
$\hat{\bm{a}} = E_\mathrm{Lbl} (\bm{l}^\mathrm{d})$\\
Compute $L_\mathrm{recon} (\hat{\bm{x}}^\mathrm{d}, \bm{x}^\mathrm{d})$, $L_\mathrm{seg} (\hat{\bm{l}}^\mathrm{d}, \bm{l}^\mathrm{d})$, $L_\mathrm{cons} (\hat{\bm{a}}, \mathrm{sg}(\bm{a}))$, and $L_\mathrm{margin} (\bm{e})$.\\
Update parameters of $E_\mathrm{AAC}$, $E_\mathrm{Lbl}$, $D_\mathrm{Img}$, and $D_\mathrm{Lbl}$ to minimize $w_\mathrm{recon} L_\mathrm{recon} + w_\mathrm{seg} L_\mathrm{seg} + w_\mathrm{cons} L_\mathrm{cons} + w_\mathrm{margin} L_\mathrm{margin}$ using stochastic gradient descent (e.g., Adam). 
}
\caption{Training of the feature extraction module of our SBMIR system}
\label{alg:learning_algorithm}
\end{algorithm}

\section{Implementation}\label{sec:implementation}

\subsection{Datasets}\label{sec:datasets}

\begin{figure}[t!]
  \centering
  \includegraphics[width=\linewidth]{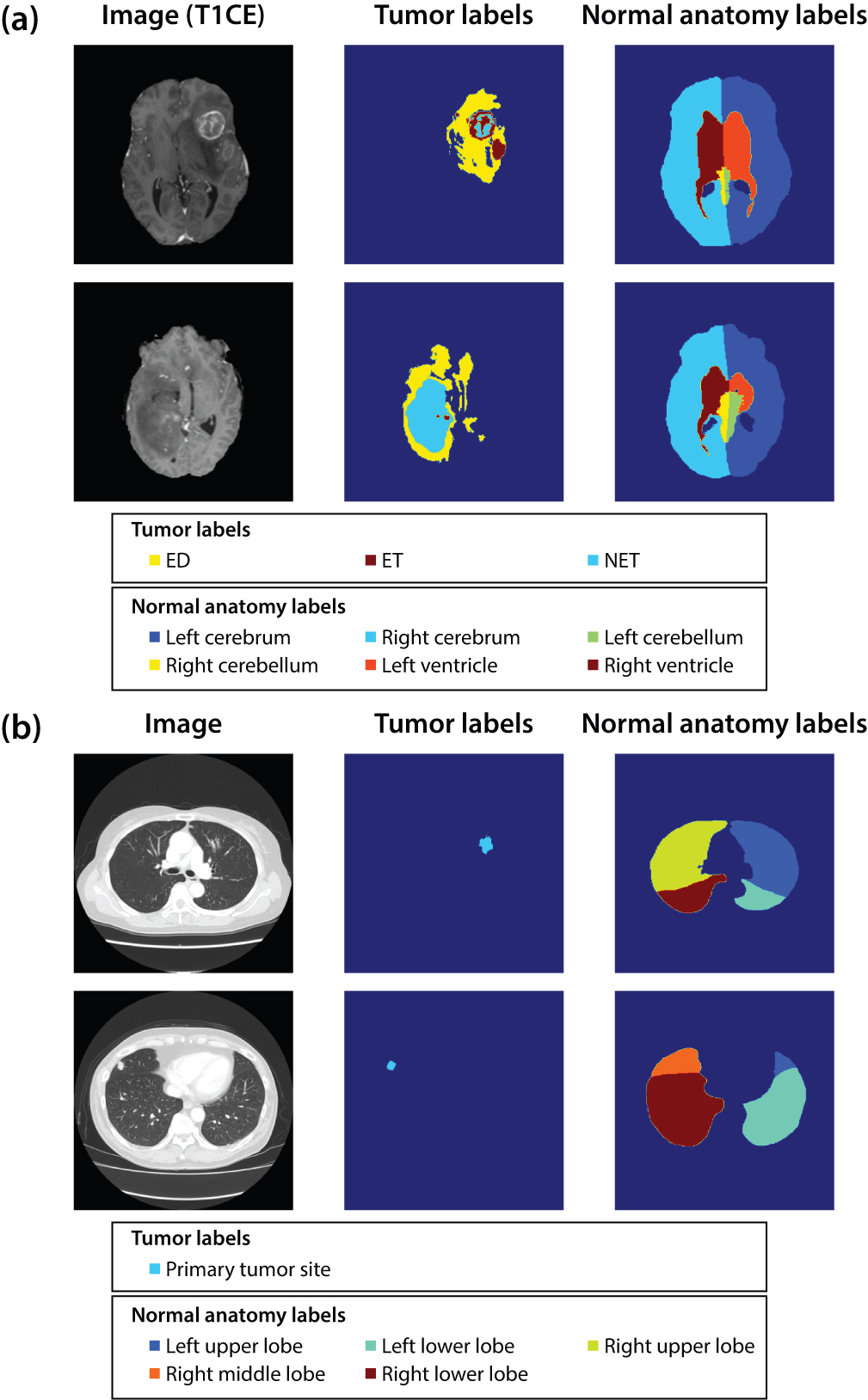}
  \caption{\textbf{Examples in the datasets.} \textbf{a} Example images with tumor-associated labels and normal-anatomy-associated labels in the glioma testing dataset are shown. \textbf{b} Example images with tumor-associated labels and normal-anatomy-associated labels in the lung cancer testing dataset are shown. Note that in the normal-anatomy-associated labels, regions annotated with tumor-associated labels were assigned normal anatomical classes that should be present therein and vice versa. ET, gadolinium-enhancing tumor; ED, peritumoral edema; NET, necrotic and non-enhancing tumor-core; T1CE, T1-weighted contrast-enhanced sequence.}
  \label{fig:dataset_examples}
\end{figure}

Two types of datasets, a \emph{glioma dataset} and a \emph{lung cancer dataset}, were used in the model training and evaluation, each of which was split into a training dataset and a testing dataset.

\subsubsection{Glioma dataset}\label{sec:glioma_dataset}

The glioma dataset initially aggregated three datasets comprising brain MRI scans with glioma obtained from the MICCAI 2019 BraTS Challenge \citep{Menze2015, Bakas2017, TCGAGBM, TCGALGG}, a dataset of 51,925 slices from 335 patients (MICCAI\_BraTS\_Training), a dataset of 19,375 slices from 125 patients (MICCAI\_BraTS\_Validation), and a dataset of 25,730 slices from 166 patients (MICCAI\_BraTS\_Testing). The patients were from multiple hospitals. Each MRI scan consists of T1-weighted (T1), T1-weighted contrast-enhanced (T1CE), T2-weighted, and fluid-attenuated inversion recovery (FLAIR) sequences. As shown in \textbf{\cref{fig:dataset_examples}a}, the tumor-associated segmentation labels include three classes: gadolinium (Gd)-enhancing tumor (ET), peritumoral edema (ED), and necrotic and non-enhancing tumor core (NET), while the normal-anatomy-associated labels include six classes: left cerebrum, right cerebrum, left cerebellum, right cerebellum, left ventricle, and right ventricle. Except for the tumor-associated labels in MICCAI\_BraTS\_Training, we supplemented each dataset with the tumor-associated labels according to the procedure described in a previous study \citep{Kobayashi2021}. Normal-anatomy-associated labels were generated using the software BrainSuite (version: 19a) \citep{Shattuck2002}. We then assigned each dataset for the use of training and testing our SBMIR system as follows: the \emph{glioma training dataset} ($N =$ 45,105 slices from 291 patients) consists of both MICCAI\_BraTS\_Validation and MICCAI\_BraTS\_Testing, and the \emph{glioma testing dataset} ($N =$ 51,925 slices from 335 patients) consists of MICCAI\_BraTS\_Training. Note that the original names of the datasets in the 2019 BraTS Challenge and the notations according to their purpose in this study are different. Following the previous study \citep{Kobayashi2021}, this difference in labeling was deemed appropriate for evaluating the performance of our SBMIR system based on data that have widely accepted and publicly available ground-truth tumor-associated labels (MICCAI\_BraTS\_Training) in order to ensure the objectivity and reproducibility. 

\subsubsection{Lung cancer dataset}\label{sec:lung_cancer_dataset}

The lung cancer dataset consists of chest CT scans from 1,000 patients with lung cancer collected from a single hospital. The study, data use, and data protection procedures were approved by the Ethics Committee of the National Cancer Center, Tokyo, Japan (protocol number 2016-496). The requirement for informed consent was waived in view of the retrospective nature of the study. All procedures followed applicable laws and regulations and the Declaration of Helsinki. As shown in \textbf{\cref{fig:dataset_examples}b}, the tumor-associated segmentation labels included one class, primary tumor (PT), the region of which was segmented by an expert radiation oncologist (K.K.). Other potential tumor-associated regions, such as lymph node metastases, were not annotated because the model training was conducted in the lung window, making it difficult to identify the diseased areas in soft tissues such as mediastinum. The normal-anatomy-associated labels included five classes: right upper lobe, right middle lobe, right lower lobe, left upper lobe, and left lower lobe. These five labels were generated by an off-the-shelf deep-learning model that is available from a public repository \citep{Hofmanninger2020}. Then, we split the patients randomly into the \emph{lung cancer training dataset} ($N =$ 49,696 slices from 600 patients) and the \emph{lung cancer testing dataset} ($N =$ 33,572 slices from 400 patients). Notably, the large testing dataset was beneficial for assessing the effectiveness of image search because it can ensure that corresponding images exist for individual user queries.

\subsection{Training settings}\label{sec:training_settings}

The detailed training settings of the models are described here. We first determined the hyperparameters using the glioma training dataset and then applied most of them to the lung cancer training dataset, except for the margin parameter. Note that the two types of datasets differed in terms of the spatial resolution of input images; the image size of the glioma training dataset was set to $256 \times 256$, and that of the lung cancer training dataset was set to $512 \times 512$. Also, it is important to note that the average voxel volume in the tumor-associated regions in the lung cancer testing dataset ($1.0 \times 10^4$) was much smaller than that in the glioma testing dataset ($4.1 \times 10^4$). 

\subsubsection{Preprocessing of the datasets}\label{sec:preproc_datasets}

For the glioma dataset (i.e., the glioma training dataset and the glioma testing dataset), T1, T1CE, and FLAIR sequences were concatenated into a three-channel MR volume. Then, $Z$-score normalization was applied channel-wise in a manner that was limited to the area inside the body. Subsequently, each 3D MR volume was decomposed into a collection of three-channel 2D axial slices, for which the size was resized to $3 \times 256 \times 256$. On the other hand, for the lung cancer dataset (i.e., the lung cancer training dataset and the lung cancer testing dataset), the voxel value was normalized into a lung window with a window width of 1500 and a window center of -550. Then, to set the number of channels to 3, which is similar to the glioma dataset, the CT volume was decomposed into a collection of three-channel 2D axial slices by concatenating each slice with the adjacent upper and lower slices. Thus, the size of the input tensor was $3 \times 512 \times 512$. This 2.5-dimensional approach for the CT slices is valid because the diagnosis of lung nodules usually requires investigation of adjacent slices to distinguish abnormal structures from normal structures such as vessels. Random rotation and random horizontal flipping were applied in the data augmentation of each image for training the models.

\subsubsection{Implementation of the neural networks}\label{sec:network_architecture}

All neural networks were implemented in Python 3.8 with PyTorch library 1.10.0 \citep{Paszke2019} using an NVIDIA Tesla V100 graphics processing unit with CUDA 10.2. We implemented almost the same neural network architecture for the two training datasets (i.e., the glioma training dataset and the lung cancer training dataset), the purpose of which was to maintain the same compression ratio from the input image to the ACs as latent representations. See \textbf{\ref{app:detailed_network_architecture}} for the detailed network architectures.

\subsubsection{Hyperparameters for the glioma dataset}\label{sec:hyparam_glioma_dataset}

For the glioma training dataset, the hyperparameters shared across the configurations were as follows: batch size = 200, the number of training epochs = 300, learning rate = $1.0 \times 10^{-4}$, weight decay = $1.0 \times 10^{-5}$, $w_\mathrm{recon} = 1.0$, $w_\mathrm{seg} = 10.0$, $w_\mathrm{cons} = 1.0$, $w_\mathrm{abn} = 0.1$, $w_\mathrm{reg} = 0.1$, $w_\mathrm{margin} = 0.1$, and $m = 10$. We determined these hyperparameters by grid search from the candidate values as follows: $w_\mathrm{recon} = \{1.0, 10.0, 100.0\}$, $w_\mathrm{seg} = \{1.0, 10.0\}$, $w_\mathrm{cons} = \{0.1, 1.0\}$, $w_\mathrm{abn} = \{0.1, 1.0\}$, $w_\mathrm{reg} = \{0.1, 1.0\}$, $w_\mathrm{margin} = \{0.1, 1.0\}$, and $m = \{0, 5, 10, 20, 40\}$.

\subsubsection{Hyperparameters for the lung cancer dataset}\label{sec:hyparam_lung_cancer_dataset}

For the lung cancer training dataset, the hyperparameters shared across the configurations were as follows: batch size = 144, the number of training epochs = 50, learning rate = $1.0 \times 10^{-4}$, weight decay = $1.0 \times 10^{-5}$, $w_\mathrm{recon} = 1.0$, $w_\mathrm{seg} = 10.0$, $w_\mathrm{cons} = 1.0$, $w_\mathrm{abn} = 0.1$, $w_\mathrm{reg} = 0.1$, $w_\mathrm{margin} = 0.1$, and $m = 40$. Owing to the relatively small areas of the tumor-associated regions, the reconstruction loss focusing on the area of tumor-associated regions was additionally calculated for the entire reconstruction. Almost all the hyperparameters above were determined as those optimized in the model training on the glioma training dataset, except for the margin parameter $m$ that was determined by grid search from the candidate values of $\{0, 20, 40, 60, 80\}$.

\subsection{Image retrieval pipeline}\label{sec:image_retrieval_pipeline}

\begin{figure*}[t!]
  \centering
  \includegraphics[width=\textwidth]{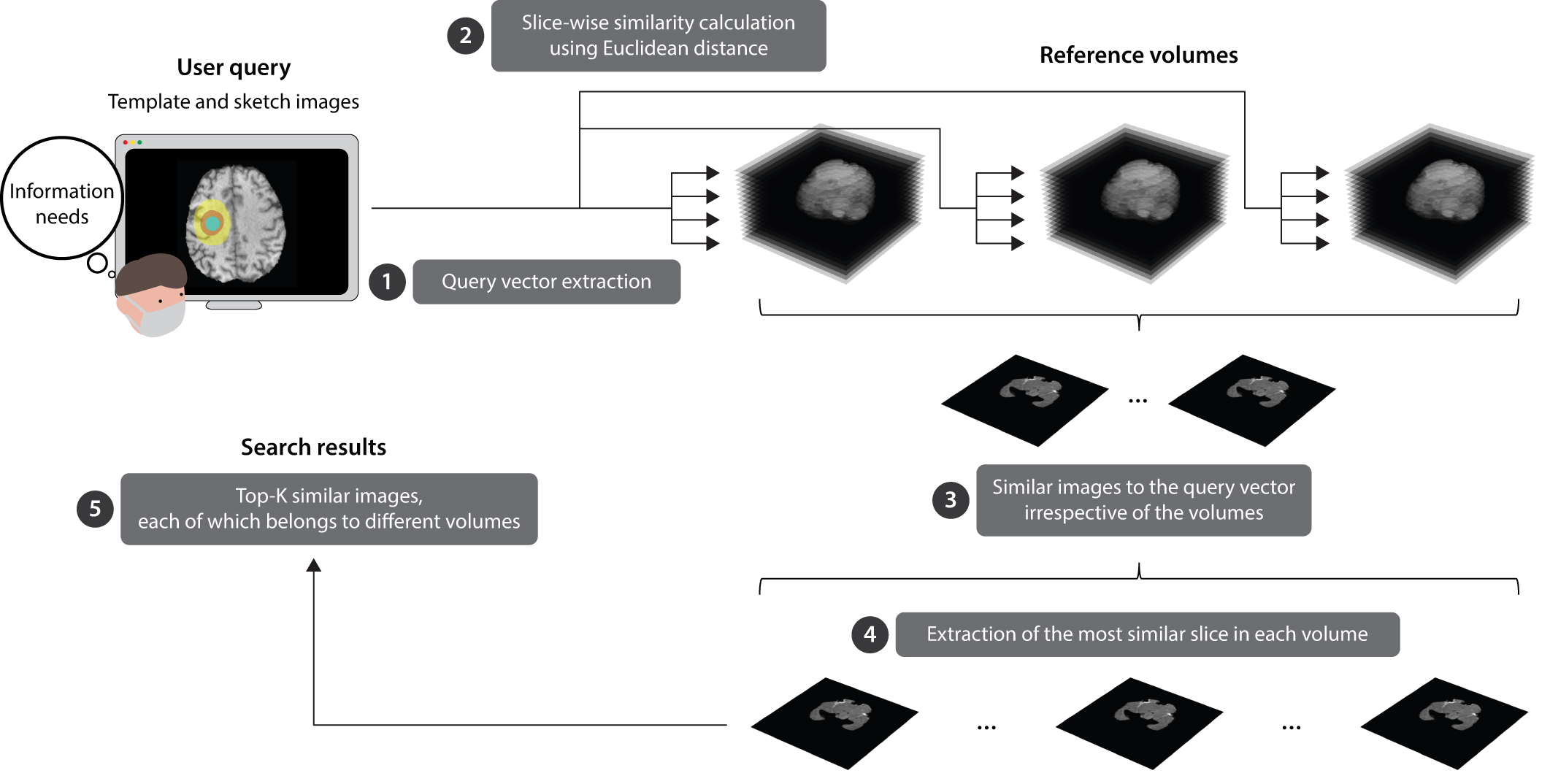}
  \caption{\textbf{Similarity calculation module of our sketch-based medical image retrieval (SBMIR) system.} The similarity calculation module of our SBMIR system is staged in five steps. First, a query vector extracted from a user query, which is the combination of a template image and a semantic sketch, is obtained from the feature extraction module. Second, the approximated Euclidean distances between the query vector and reference vectors of every slice in the reference volumes are computed. Third, the slices with reference vectors close to the query vector are identified irrespective of the reference volumes. Fourth, the extracted slices are rearranged to select the most similar slice in each reference volume. Fifth, the top-$K$ most similar images belonging to different reference volumes are presented to the user as the search results.}
  \label{fig:retrieval_pipeline}
\end{figure*}

After the model is trained, the whole SBMIR system combining the feature extraction module and the similarity calculation module can be implemented. Using the feature extraction module, reference images in a database were converted into reference vectors before user operation (see the right part of \textbf{\cref{fig:query_by_sketch}c}). Each reference vector was constructed from a reference image as the sum of a normal AC (through the normal AC encoder) and an abnormal AC (through the abnormal AC encoder). At the time of the image retrieval, the two-step user operation constructs a query vector to meet the information needs (see the left part of \textbf{\cref{fig:query_by_sketch}c}): first, selecting a template image to specify the location where the target image content should exist, and second, sketching the semantic segmentation label of the disease to express the image content therein. Then, the normal AC encoder extracts a normal AC from the template image, and the label encoder extracts an abnormal AC from the semantic sketch of the disease, both of which are summed as the query vector according to \textbf{Eq. \ref{eq:decomposition}}.

In the similarity calculation module (see the middle part of \textbf{\cref{fig:query_by_sketch}c}), the distance of the reference vectors to the query vector was calculated using approximated Euclidean distances by an algorithm called \emph{Annoy} \citep{annoy}. The reference images with close reference vectors to the query vector are then rearranged according to each reference volume. Finally, the top-$K$ similar images, each of which belongs to a different reference volume, are obtained. Note that this volume-wise similarity calculation is essential to avoid the redundancy resulting from consecutive slices with similar appearances in a single 3D volume. \textbf{\Cref{fig:retrieval_pipeline}} explains the details of the similarity calculation module to realize volume-wise image retrieval in our SBMIR system.

\subsection{Implementation of our SBMIR system on the glioma dataset}\label{sec:imple_glioma_dataset}

The feature extraction module was trained on the glioma training dataset. Then, the image retrieval pipeline was implemented on the glioma testing dataset for evaluation. The template volume was selected to be the image series with the minimum volume of tumor-associated regions in the glioma testing dataset, as the smaller the tumor volume, the smaller the deviation of normal structures can be. An ideal template volume may be a totally healthy image series; however, the glioma dataset did not contain an image series without abnormal findings. Thus, this may be the second-best setting for the template volume. For gliomas, we found that users can specify the following types of information regarding the target images to be retrieved: the location, shape, size, and internal characteristics of the tumor (see \textbf{\cref{sec:sbmir_fine_grained}}), which we call \emph{fine-grained characteristics} of the disease. Among these, the location, shape, and size of the target diseased area can be defined by selecting a template image and sketching the outer edge of the tumor. Furthermore, depending on which combination of tumor-associated labels (i.e., ET, ED, and NET) is used to sketch the tumor, the internal characteristics of the tumor (e.g., contrast enhancement effect, necrosis, presence of edema, etc.) can be expressed. 

\begin{table*}[t]
\centering
\caption{\textbf{Backgrounds of the evaluators and participation in the user test.} Ten and nine healthcare professionals participated in the evaluations of gliomas and lung cancers, respectively.}
\begin{tabular}{@{}lllll@{}}
\toprule
\textbf{Evaluators} & \textbf{Background}    & \textbf{Years of experience} & \textbf{Glioma} & \textbf{Lung cancer} \\ \midrule
\#1                 & Radiation oncologist   & 10 - 19                      & \checkmark               & \checkmark                    \\
\#2                 & Medical oncologist     & 10 - 19                      & \checkmark               & \checkmark                    \\
\#3                 & Diagnostic radiologist & 20 - 29                      & \checkmark               & \checkmark                    \\
\#4                 & Diagnostic radiologist & 20 - 29                      & \checkmark               & \checkmark                    \\
\#5                 & Neurosurgeon           & 20 - 29                      & \checkmark               &                               \\
\#6                 & Colorectal surgeon     & 10 - 19                      & \checkmark               & \checkmark                    \\
\#7                 & Thoracic surgeon       & 20 - 29                      & \checkmark               & \checkmark                    \\
\#8                 & Medical physicist      & 10 - 19                      & \checkmark               & \checkmark                    \\
\#9                 & General surgeon        & 1 - 9                        & \checkmark               & \checkmark                    \\
\#10                & Medical researcher     & 1 - 9                        & \checkmark               & \checkmark                    \\ \bottomrule
\end{tabular}
\label{tab:evaluators}
\end{table*}

\subsection{Implementation of our SBMIR system on the lung cancer dataset}\label{sec:imple_lung_cancer_dataset}

After training the feature extraction module using the lung cancer training dataset, the image retrieval pipeline of our SBMIR system was implemented on the lung cancer testing dataset for evaluation. The template image series was also selected to be the one with the minimum volume of tumor-associated regions in the lung cancer testing dataset, similar to the glioma dataset. For lung cancers, we found that users can identify fine-grained characteristics of the disease, including the location, shape, and size of a tumor (see \textbf{\cref{sec:sbmir_fine_grained}}); however, in contrast with gliomas, the internal characteristics of the tumor (e.g., ground-glass opacity and solid tumor components) are not explicitly expressed by the model because only a single-class tumor-associated segmentation label (i.e., PT) was given in the training dataset. 

\section{Evaluation}\label{sec:evaluation}

The present study comprehensively evaluates our SBMIR system from technological and clinical standpoints. As for technical evaluations, the training results of the feature decomposition, hyperparameter optimization focusing on the image retrieval performance, and ablation studies are described in \textbf{\ref{app:training_results}}, \textbf{\ref{app:eval_image_retrieval}}, and \textbf{\ref{app:ablation_studies}}, respectively. In this section, we explain the details of the clinical evaluation, which focuses on how our SBMIR system can help the information-seeking objectives of healthcare professionals.

In exchange for the flexibility of the query-by-sketch approach, standardizing the user query and preparing the ground truth for the retrieved images are challenges. Therefore, user testing is the most valid evaluation schema to assess the image retrieval performance of our SBMIR system. A group of healthcare professionals with various clinical backgrounds, including radiologists, physicians, surgeons, medical physicists, and researchers, participated in the user tests as the evaluators (see \textbf{\cref{tab:evaluators}}). The user tests were conducted using a dedicated GUI (see \textbf{\cref{fig:sbmir_system}b}). 

\subsection{Definition of clinical similarity}\label{sec:dif_clinical_similarity}

 We developed criteria for the evaluators to determine whether the retrieved image was clinically similar to the features specified by the user query for each dataset. 
 
 \subsubsection{Clinical similarity of glioma images}\label{sec:dif_clinical_similarity_glioma}

For gliomas, our SBMIR system can specify the location, shape, size, and internal characteristics of the tumor, as demonstrated in \textbf{\cref{sec:sbmir_fine_grained}}. Among these characteristics, we considered that determining whether the shape is consistent or not can be too subjective to standardize among evaluators. Hence, the clinically similar images were defined to be the images that met the following three criteria: (1) a difference in the maximum tumor diameter of within 2 cm; (2) the same location of the tumor according to the brain lobe (i.e., right frontal lobe, right parietal lobe, right occipital lobe, right temporal lobe, left frontal lobe, left parietal lobe, left occipital lobe, and left temporal lobe); (3) the same pattern of the contrast enhancement (e.g., the presence of contrast-enhancement and tumor necrosis). When a user intention included a more detailed location (e.g., the relative position in a brain lobe), the evaluators were required to judge whether the retrieved images matched the detailed location. Each criterion was judged with a score of 0 or 1, and a maximum score of 3 points was possible for each image, which we call the \emph{similarity score}. When the similarity score was 3/3, the retrieved image was considered clinically similar to the query. 

 \subsubsection{Clinical similarity of lung cancer images}\label{sec:dif_clinical_similarity_lung_cancer}

For lung cancers, our SBMIR system can specify the location, shape, and size of the tumor, as demonstrated in \textbf{\cref{sec:sbmir_fine_grained}}. Note that the internal characteristics of the tumor are not explicitly expressed, as only a single class of tumor-associated segmentation labels (i.e., PT) was applied to the datasets. We also concluded that determining the concordance in the shape of lung cancer can be too subjective. Therefore, the clinical similarity was defined according to the following two criteria: (1) a difference in the maximum tumor diameter of within 2 cm; (2) the same location of the tumor according to the lung lobe (i.e., right upper lobe, right middle lobe, right lower lobe, left upper lobe, left lower lobe). When a user intention included a more detailed location (e.g., the lung segment, the apex of the lung, the pleural contact, and the hilar region), the evaluators were asked to judge whether the retrieved images matched the detailed location. Each criterion was evaluated with a score of 0 or 1, and a maximum similarity score of 2 points was possible for each image. When the similarity score was 2/2, the retrieval image was considered clinically similar to the query. 

\begin{figure*}[t!]
  \centering
  \includegraphics[]{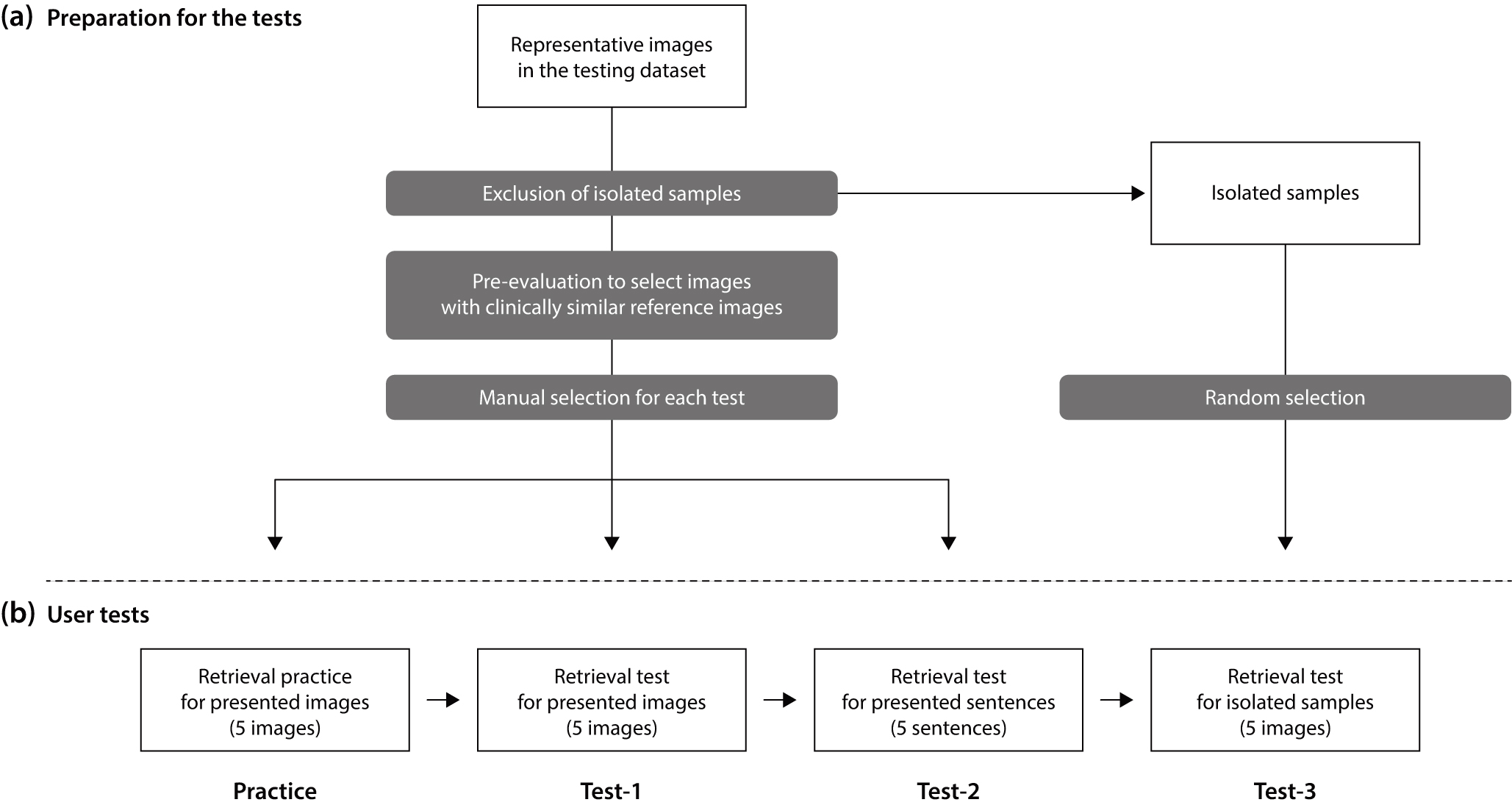}
  \caption{\textbf{Evaluation flow for our sketch-based medical image retrieval (SBMIR) system.} \textbf{a} To prepare question items for each test, we conducted the following steps. First, a set of representative images was identified in a testing dataset. Second, the isolated samples were excluded from the set, from which five isolated samples were randomly assigned to Test-3. Third, from the remaining representative images, those with at least one clinically similar image were extracted in the pre-evaluation process based on the direct comparison of Dice similarities. Fourth, expert radiologists selected representative images for the question items of the practice stage, Test-1, and Test-2, in order to vary the clinical features as much as possible. \textbf{b} In the user tests, the healthcare professionals first learned how to operate our SBMIR system in the practice stage. After that, retrieval tests for presented images (Test-1), presented sentences (Test-2), and isolated samples (Test-3) were conducted sequentially.}
  \label{fig:evaluation_flow}
\end{figure*}

\subsection{Evaluation metrics for the retrieved images}\label{sec:eval_metrics}

We devised two types of evaluation metrics for the retrieved images -- \emph{user-oriented} and \emph{objective metrics}. Every evaluation metric was averaged among the evaluators ($N=10$ for gliomas and $N=9$ for lung cancers, as shown in \textbf{\cref{tab:evaluators}}), and the mean $\pm$ standard deviation of each metric was reported. 

\subsubsection{User-oriented evaluation metrics}\label{sec:user_oriented_metrics}

The user-oriented metrics included precision@K, reciprocal rank, and normalized discounted cumulative gain (nDCG), which reflect how the retrieved results were clinically similar to the intent of the query based on the judgment of the evaluators. These were evaluated according to a 2D slice basis; that is, when evaluating the similarity score to calculate these metrics, each evaluator interpreted the consistency between the intention of the query and retrieved images without considering the adjacent slices of each retrieved image. Because the framework of the user tests is characterized by the large size of the datasets ($N = 51,925$ images in the glioma testing dataset and $N = 33,572$ images in the lung cancer testing dataset) and the fine-grained attributes that user queries can specify, it is straightforward to assess how many of the retrieved top-$K$ images contain corresponding images that satisfy the intent of the query by the user-oriented metrics. We defined the following three user-oriented metrics.

Precision@K \citep{Shirahatti2005} is the ratio between the number of clinically similar images in the top-$K$ retrieved images and the number of retrieved images $K$, which is formulated as
\begin{equation}\label{eq:precision}
\mathrm{Precision@K} = \frac{\rvert \mathcal{S} \bigcap \mathcal{L}_K \rvert}{K},
\end{equation}
where $\mathcal{S}$ is the set of clinically similar images judged by an evaluator, $\mathcal{L}_K$ is the list of the top-$K$ retrieved images, and $K$ is the number of retrieved images. 

The reciprocal rank \citep{Pedronette2015} is calculated from the inverse of the rank position of the first clinically similar image in the top-$K$ retrieved images, which is defined as 
\begin{equation}\label{eq:mrr}
\mathrm{Reciprocal\,rank} = \frac{1}{k},
\end{equation}
where $k$ is the rank position of the first clinically similar image in the top-$K$ retrieved images. When there is no clinically similar image in the retrieval list, the reciprocal rank is set to 0. 

The nDCG \citep{Wang2013} is the total similarity score of the retrieved images in the order defined as 
\begin{equation}\label{eq:ndcg}
\mathrm{nDCG} = \frac{1}{\mathrm{DCG}_\mathrm{perfect}} \left( s_1 + \sum^K_{k = 2} \frac{s_k}{\log_2 k} \right),
\end{equation}
where $s_i$ indicates the score of the $i$-th ranked image, $k$ is the rank position, and $\mathrm{DCG}_\mathrm{perfect}$ represents the maximum DCG in an ideal retrieval result to normalize the value within a range from 0 to 1. The role of the log functions is to discount the score of retrieved images that ranked lower. 

Note that precision@5 can be influenced by how many clinically similar images for each user query were originally included in the testing dataset. As such, precision@5 for rare images can be small even when the image retrieval works successfully. Hence, we prepared the user tests to guarantee that the lower bound of precision@5 will be 0.2 when the image retrieval works properly, which will be described in detail in \textbf{\cref{sec:prep_user_test}}.

\subsubsection{Objective evaluation metrics}\label{sec:objective_metrics}

Recall@K was defined as an objective metric that can be assessed independently of the evaluator's interpretation of the retrieval results. This objective metric was determined according to a 3D volume basis, reflecting whether the images belonging to the same volume as the presented image itself, which we call \emph{the same-volume images}, are listed among the top-$K$ retrieved images. 

Recall@K \citep{Shirahatti2005} is automatically calculated based on whether the same-volume images are acquired in the top-$K$ retrieved images, which is formulated as
\begin{eqnarray}\label{eq:recall}
\mathrm{Recall@K}
 =
  \begin{cases}
    1 & ( i \in \mathcal{L}_K ) \\
    0 & ( i \notin \mathcal{L}_K ),
  \end{cases}
\end{eqnarray}
where $i$ is the same-volume image, and $\mathcal{L}_K$ is a list of the top-$K$ retrieved images. 

\subsection{Preparation for the user tests}\label{sec:prep_user_test}

As illustrated in \textbf{\cref{fig:evaluation_flow}a}, we prepared question items for the practice stage and the three stages of user tests, including Test-1, Test-2, and Test-3. Each stage included five question items, consisting of either images or text descriptions, presented to the evaluators to retrieve medical images consistent with the clinical characteristics in each question item. To prepare question items, we focused on \emph{representative images}. A representative image is defined as the 2D axial slice containing the largest tumor-associated region in each 3D volume, which is usually considered to be the image that best characterizes the clinical meaning of the volume.

Each question item was developed based on a specific representative image assigned in the following steps. First, from a set of representative images in each testing dataset, we identified isolated samples that were ``not'' included in the 5-nearest neighbors (NN) groups of any other images using a fine-tuned ResNet-101 (see \textbf{\ref{app:isolated_samples}}) for exclusion from Test-1 and Test-2. Five of these isolated samples were randomly selected for Test-3. Then, \emph{pre-evaluation} was conducted on the remaining representative images. The purpose of the pre-evaluation was to identify a set of representative images that were certain to have at least one clinically similar image in the testing dataset. To confirm this, we directly compared Dice similarities \citep{Dice1945} to maximize the average overlap of the tumor-associated labels and the normal-anatomy-associated labels (see \textbf{\cref{fig:dataset_examples}}). As described previously \citep{Kobayashi2021}, a CBMIR based on the direct comparison of Dice similarities can be considered an oracle for retrieving images similar to a query image. The Dice similarities were computed between each representative image and all the other images in the dataset, and the similar images were then rearranged volume-wise, in the same manner as shown in \textbf{\cref{fig:retrieval_pipeline}}. Subsequently, whether the top-5 retrieved images based on the Dice similarities included at least one clinically similar image according to the similarity score was evaluated from a clinical perspective, and the representative images that did not meet this criterion were excluded (i.e., the lower bound of precision@5 can be 0.2). Lastly, the \emph{manual selection} was performed to assign as diverse a selection of disease phenotypes as possible in each testing stage, considering the tumor location and other disease characteristics. 

After assigning five representative images for each stage, the practice stage, Test-1, and Test-3 used the assigned images as question items presented to the evaluators. For Test-2, text descriptions as question items expressing specific clinical findings, to simulate radiology reports, were made based on the assigned images. Because it is generally difficult to fully communicate the image characteristics of a medical image by text, the originally assigned representative images were only used as references for making the text descriptions and were not used in the evaluation process.

\subsection{Flow of the user tests}\label{sec:flow_user_test}

\textbf{\Cref{fig:evaluation_flow}b} illustrates the flow of the user tests, consisting of the practice stage, Test-1, Test-2, and Test-3. For each of the testing datasets, the evaluators first underwent the practice stage. In the practice stage, five sets of a representative image and a ground-truth tumor-associated label were consecutively presented to the evaluators. By referring to the ground-truth tumor-associated labels as guidance for sketching the diseases, they were able to learn how to construct a user query to specify the content of medical images. Also, they received automated feedback on whether the same-volume image was listed in the top-5 retrieved images to confirm the validity of the user query. This practice stage is important to demonstrate that our SBMIR system can be learned easily with minimum practice. Then, three types of user tests were conducted as follows. Test-1 demonstrated the image retrieval performance when example images were available (see \textbf{\cref{sec:sbmir_with_examples}}). Test-2 revealed the image retrieval performance without example images (see \textbf{\cref{sec:sbmir_without_examples}}). Test-3 investigated the image retrieval performance for isolated samples (see \textbf{\cref{sec:sbmir_isolated_samples}}). Each test stage contained five question items indicated by images or text descriptions, which is described in detail in \textbf{\ref{app:description_question_items}}. User-oriented metrics were independently evaluated by each evaluator in Test-1 and Test-2, while an objective metric was automatically assessed in Test-1 and Test-3. 

\subsection{Comparison with conventional CBMIR methods}\label{sec:comp_conventional_cbmir}

A conventional CBMIR system using the query-by-example approach was implemented for the comparison purpose. The fine-tuned ResNet-101 was used as a feature extractor (see \textbf{\ref{app:resnet_feature}}), and the image retrieval pipeline was built in a similar manner to our SBMIR system (see \textbf{\cref{fig:retrieval_pipeline}}). In Test-1, where example images were available, the image retrieval results were evaluated based on the user-oriented metrics. An expert diagnostic radiologist (M.M.) and an expert radiation oncologist (K.K.) were responsible for the aforementioned processes that required clinical perspectives, including the pre-evaluation and manual selection in \textbf{\cref{sec:prep_user_test}}. 

\section{Results and discussion}\label{sec:results}

\begin{figure*}[htp!]
  \centering
  \includegraphics[]{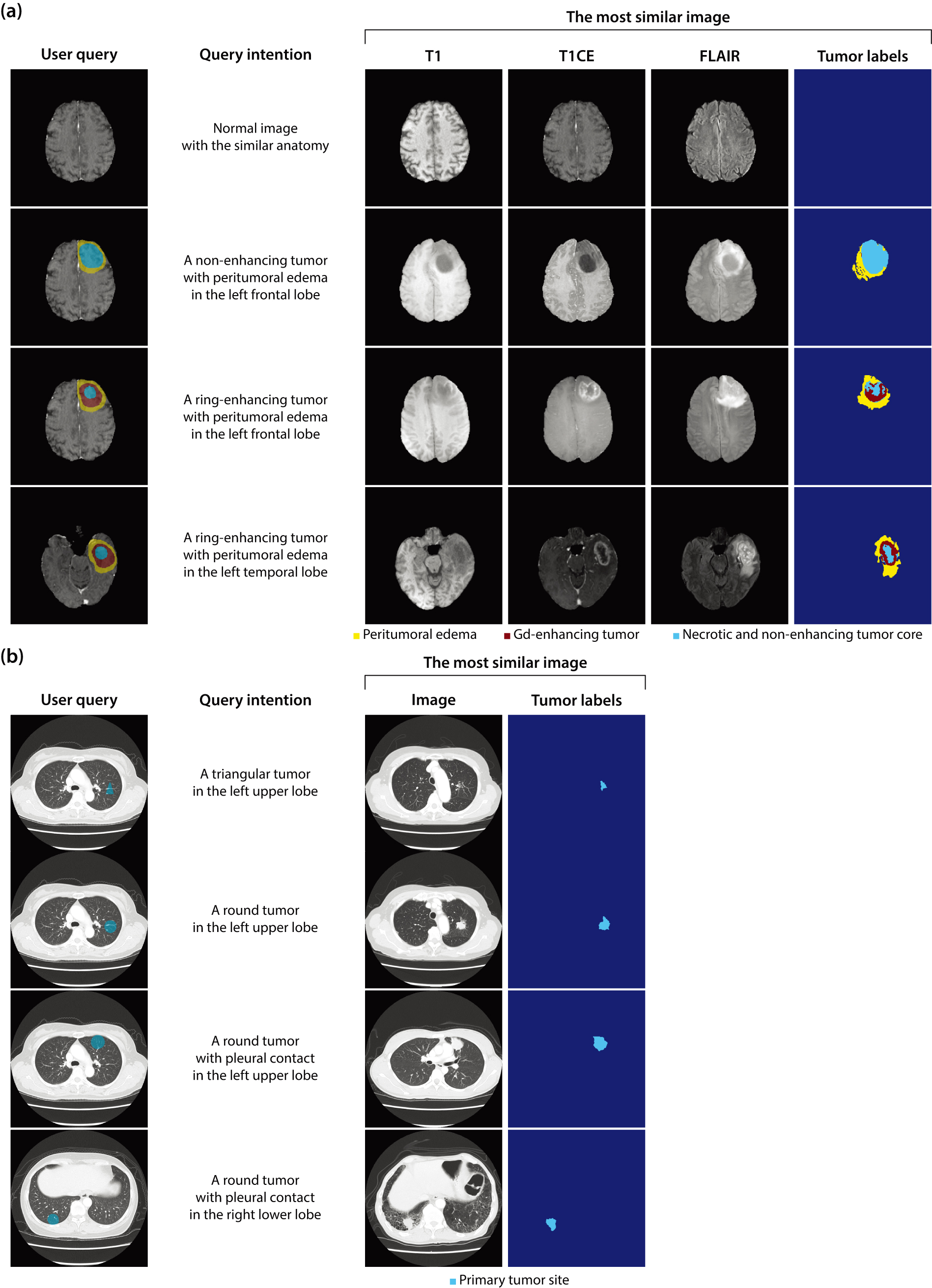}
  \caption{\textbf{Image retrieval performed according to fine-grained characteristics.} \textbf{a} Example search results from our sketch-based medical image retrieval (SBMIR) system on the glioma testing dataset. \textbf{b} Example search results of our SBMIR system on the lung cancer testing dataset. These results show that the fine-grained characteristics of the retrieved images change depending on how the tumor is sketched and which template image is selected. Gd, gadolinium; T1, T1-weighted sequence; T1CE, T1-weighted contrast-enhanced sequence; FLAIR, fluid-attenuated inversion recovery sequence.}
  \label{fig:how_sbmir_works}
\end{figure*}

\subsection{SBMIR retrieved medical images according to fine-grained imaging characteristics}\label{sec:sbmir_fine_grained}

Here, we demonstrate that our SBMIR system enables medical image retrieval to be performed according to fine-grained characteristics, including the location, shape, size, and internal characteristics of a disease. Because a query vector can be specified based on which 2D template image is selected from the 3D template volume and how the disease is sketched, we observed how the search results changed when each piece of information in a user query was varied. 

\textbf{\Cref{fig:how_sbmir_works}a} demonstrates medical image retrieval performed according to the fine-grained characteristics of gliomas. The radiological components of gliomas were categorized into three classes using tumor-associated segmentation labels-- ET, ED, and NET--all of which can be sketched on a selected template image. We started with a user query, consisting of only a template image without a sketch, to retrieve a normal image with similar anatomy to the template image (see the first row). This result was consistent with the assumption that normal ACs convey only information characterizing normal anatomy. We then drew a semantic sketch representing a non-enhancing tumor surrounded by mild peritumoral edema in the left frontal lobe on the same template image, which successfully retrieved an image containing the intended characteristics (see the second row). When we changed the internal characteristics of the tumor to exhibit ring enhancement, the retrieval results were again changed according to the intention of the query (see the third row). Finally, we sketched a similar disease in a different location on a different template image, the left temporal lobe, whereby the retrieved image also accompanied a tumor with the specified features at the intended location (see the fourth row). The retrieved image in the second row suggests a low-grade glioma, and those in the third and fourth rows suggest high-grade gliomas \citep{TCGAGlioma2015}. Hence, by configuring the three types of tumor-associated segmentation labels to specify the image content, even an image with a particular differential diagnosis can be retrieved. 

\textbf{\Cref{fig:how_sbmir_works}b} shows that flexible medical image retrieval can also be successfully performed for lung cancers. Only a single class of tumor-associated segmentation label representing PT can be sketched on a selected template image because, in contrast with gliomas, the internal characteristics of lung cancers were not explicitly modeled. We started by sketching a tumor with a triangular shape in the left upper lobe on a template image, successfully retrieving an image with a tumor exhibiting the corresponding shape and location (see the first row). Subsequently, we altered the shape of the tumor to round, changing the retrieval results to include an image with a relatively round tumor located in the same region (see the second row). The detailed anatomical location of the semantic sketch in the same template image can also affect the search results. To demonstrate this, we sketched a round tumor at a different position on the same template image to contact the pleura, thus retrieving a tumor with pleural contact (see the third row). Lastly, we sketched a similar tumor in the right lower lobe on a different template image, changing the location as intended (see the fourth row). Hence, the detailed characteristics regarding the size, shape, and anatomical location of lung cancers can be reflected in the search results.

\begin{figure*}[ht!]
  \centering
  \includegraphics[]{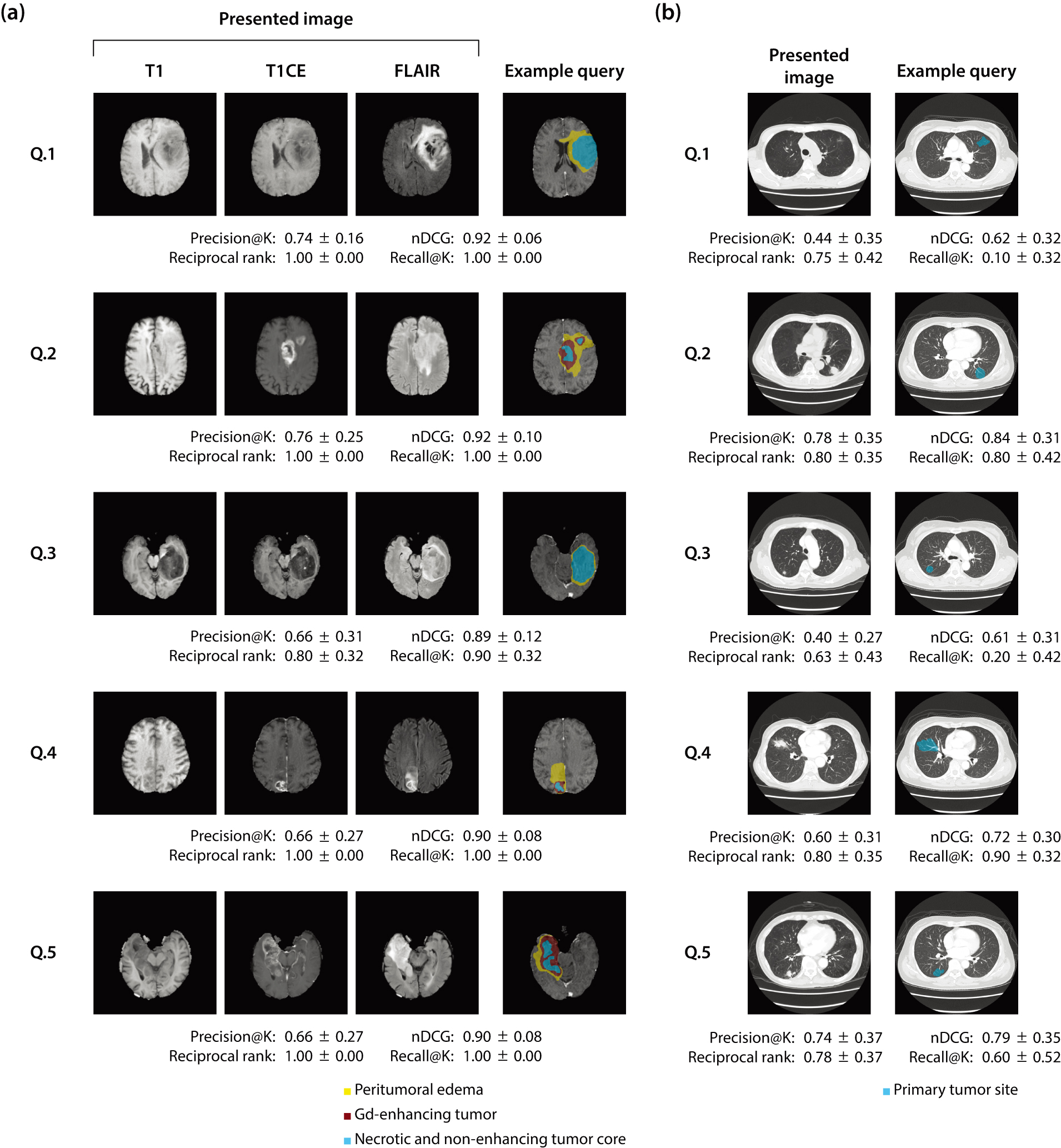}
  \caption{\textbf{Five question items and their evaluation metrics in Test-1.} \textbf{a} Based on the glioma testing dataset, question items Q.1 to Q.5 were presented to the evaluators in Test-1 to reproduce the image characteristics of the presented images in user queries for retrieving clinically similar images, including the presented one. The means $\pm$ standard deviations of the evaluation metrics, including precision@5, reciprocal rank, normalized discounted cumulative gain (nDCG), and recall@5, were calculated among the evaluators ($N = 10$). Example queries that successfully retrieved at least one clinically similar image are shown. \textbf{b} The five question items in Test-1 using the lung cancer testing dataset are shown with example queries. The means $\pm$ standard deviations of the same evaluation metrics were calculated among the evaluators ($N = 9$). Image retrieval based on our sketch-based medical image retrieval system was successful in terms of clinical similarity according to the fine-grained characteristics in most cases for both gliomas and lung cancers. Gd, gadolinium; T1, T1-weighted sequence; T1CE, T1-weighted contrast-enhanced sequence; FLAIR, fluid-attenuated inversion recovery sequence.}
  \label{fig:test_1}
\end{figure*}

\begin{figure*}[ht!]
  \centering
  \includegraphics[]{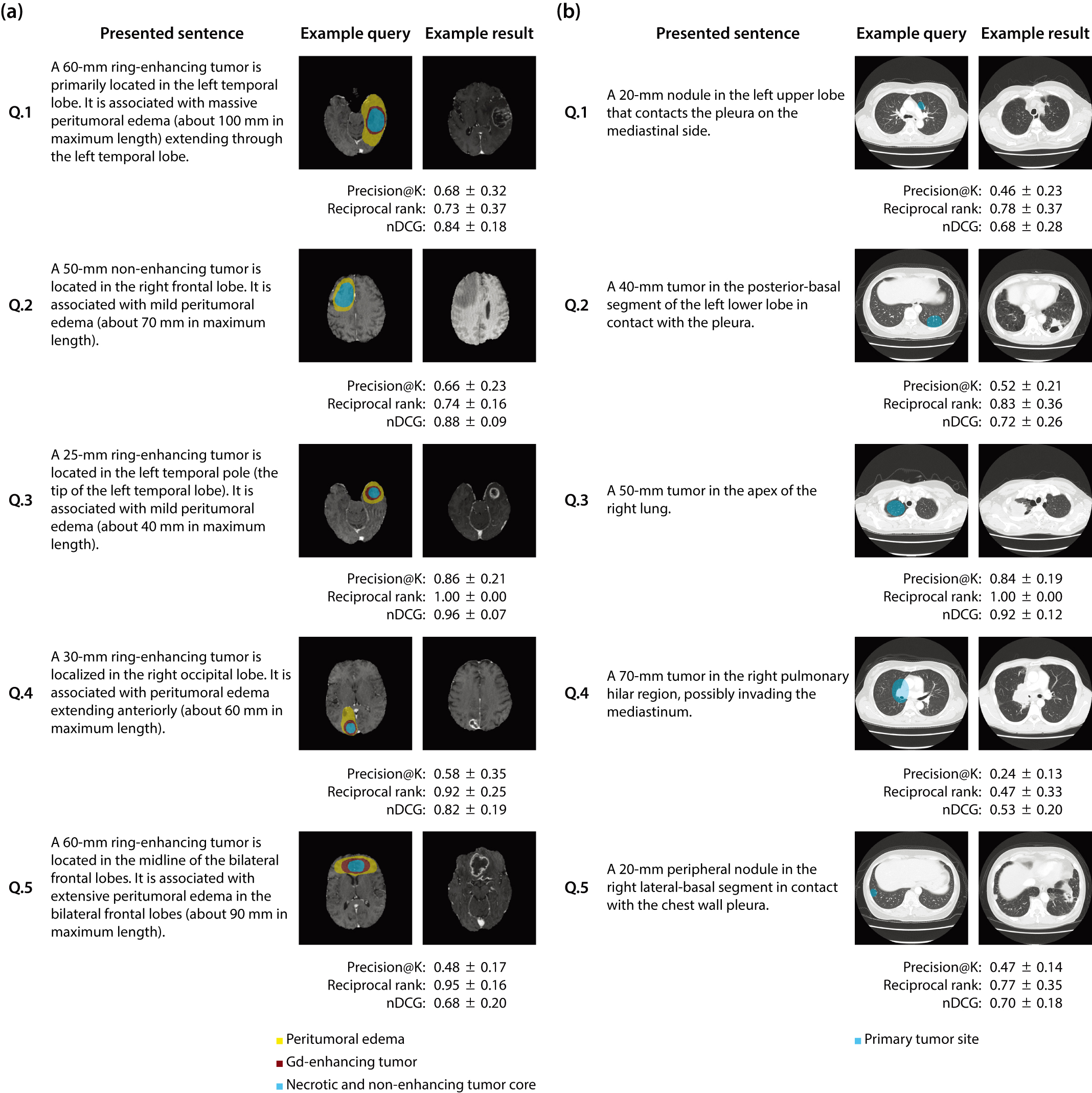}
  \caption{\textbf{Five question items and their evaluation metrics in Test-2.} \textbf{a} The five question items in Test-2 on the glioma testing dataset are shown with example user queries and retrieved images. The retrieved images are shown as T1-weighted contrast-enhanced sequences. The means $\pm$ standard deviations of the evaluation metrics, including precision@5, reciprocal rank, and normalized discounted cumulative gain (nDCG), were calculated among the evaluators ($N = 10$). \textbf{b} The five question items in Test-2 on the lung cancer testing dataset are shown with example user queries and retrieved images. The means $\pm$ standard deviations of the same evaluation metrics were calculated among the evaluators ($N = 9$). At least one or more clinically similar images were obtained for each statement in both datasets. Hence, our sketch-based medical image retrieval system enables medical image retrieval even without example images. Gd, gadolinium.}
  \label{fig:test_2}
\end{figure*}

\begin{figure*}[ht!]
  \centering
  \includegraphics[]{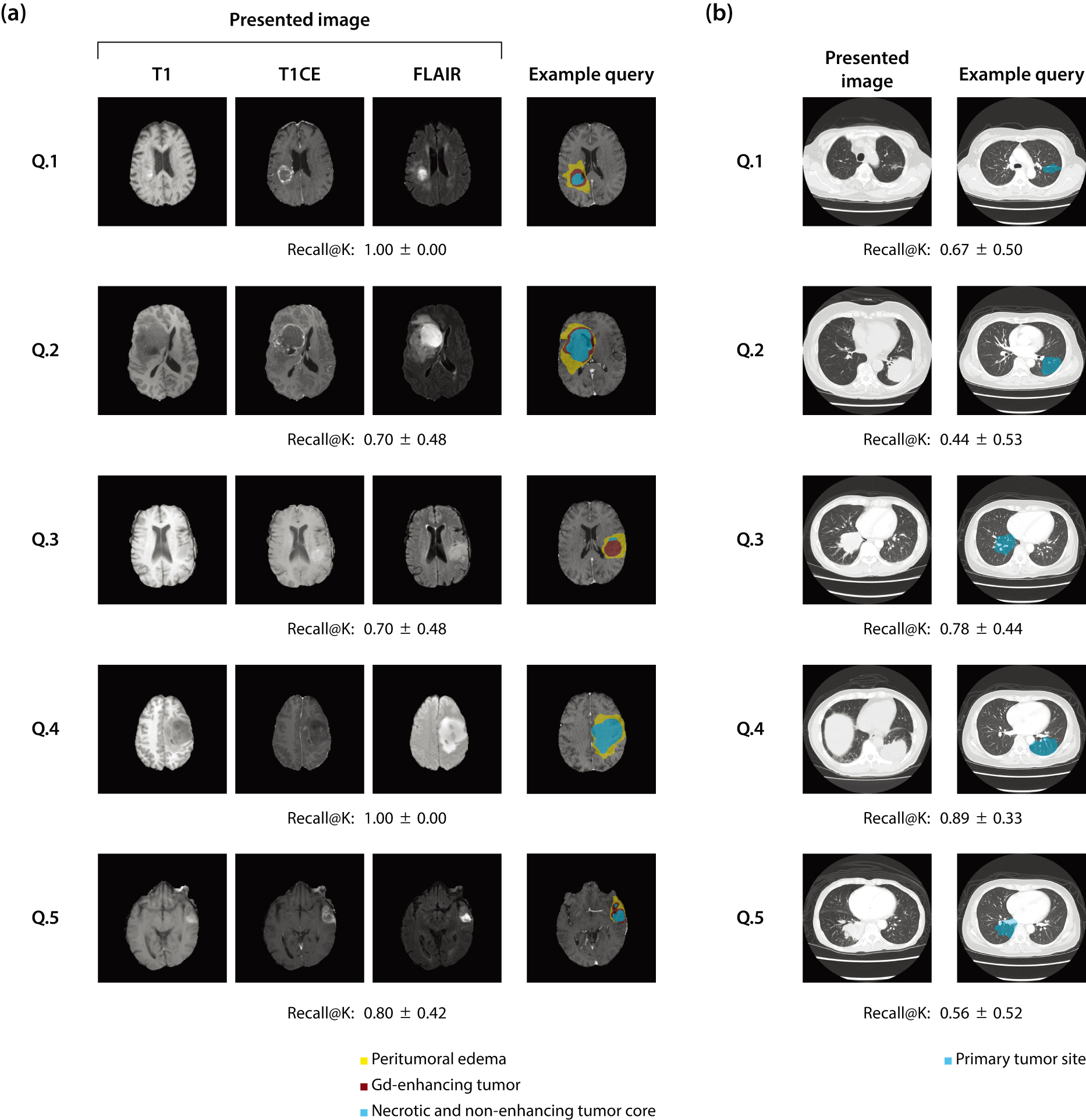}
  \caption{\textbf{Five question items and their evaluation metrics in Test-3.} \textbf{a} The five question items, which present isolated samples to retrieve in Test-3 on the glioma testing dataset and their evaluation metrics using recall@5, are shown. The means $\pm$ standard deviations of recall@5 were calculated among the evaluators ($N = 10$). \textbf{b} The five question items in Test-3 on the lung cancer testing dataset and their evaluation metrics are shown. The means $\pm$ standard deviations of recall@5 were calculated among the evaluators ($N = 9$). For both datasets, images belonging to the same volume as the isolated samples (i.e., same-volume images) were successfully retrieved by our sketch-based medical image retrieval system. Gd, gadolinium; T1, T1-weighted sequence; T1CE, T1-weighted contrast-enhanced sequence; FLAIR, fluid-attenuated inversion recovery sequence.}
  \label{fig:test_3}
\end{figure*}

\subsection{SBMIR outperformed the existing CBMIR even when example images were available}\label{sec:sbmir_with_examples}

In Test-1, the evaluators were asked to translate the characteristics of a presented image into a user query to retrieve clinically similar images, including the images belonging to the same volume as the presented one, which we call the same-volume images. There were two primary purposes of this test. First, by observing whether the same-volume images were retrievable or not, the representation power of user queries was assessed. Second, the image retrieval performance of our SBMIR system was compared with that of the existing CBMIR system based on the query-by-example approach, which used the presented images as example images (see \textbf{\cref{sec:comp_conventional_cbmir}}). 

For gliomas, five images with various sizes, locations, and patterns of contrast enhancement, were presented to the evaluators (see \textbf{\cref{fig:test_1}a} and \textbf{\ref{app:test-1_questions}}). The average recall@5 among the evaluators was consistently high, above 0.9. Such high recall@5 values indicate that the user queries could capture most of the characteristics of the presented images and thereby successfully retrieve the same-volume images. Before the user tests, we had thought that it might be difficult for some evaluators to construct appropriate user queries based on their understanding of the presented image; however, such a \emph{skill-based limitation} hindering image retrieval performance was not substantial for gliomas. Furthermore, all presented images had a precision@5 averaging above 0.6, indicating that at least three clinically similar images were listed on average among the top-5 retrieved images (see \textbf{\cref{sec:dif_clinical_similarity}} for the definition of clinical similarity). For comparison, the means $\pm$ standard deviations of precision@5, reciprocal rank, nDCG, and recall@5 for the retrieval results among the question items based on the query-by-example approach were $0.32 \pm 0.18$, $1.00 \pm 0.00$, $0.68 \pm 0.07$, and $1.00 \pm 0.00$, respectively. As the presented images were also subject to image retrieval, it is natural that recall@5 and reciprocal rank values were as high as 1.00, reflecting the fact that all the presented images were the top ranked images. Nevertheless, the consistently higher precision@5 of our SBMIR system (i.e., all precision@5 values in \textbf{\cref{fig:test_1}a} are above 0.32) supports its superior image retrieval performance according to fine-grained image characteristics even when example images are available. \textbf{\Cref{fig:test_1_glioma}a} shows the results for Q.2, whose recall@5 ($1.00 \pm 0.00$) was the highest, while \textbf{\cref{fig:test_1_glioma}b} shows the results for Q.3, whose recall@5 ($0.90 \pm 0.32$) was the lowest. These examples demonstrate the robustness of our system, as most of the retrieved images were judged to be clinically similar, even though each evaluator sketched the disease with unique details on the same or slightly different template images. 

For lung cancers, five images with different sizes and locations were presented to the evaluators (see \textbf{\cref{fig:test_1}b} and \textbf{\ref{app:test-1_questions}}). Notably, recall@5 varied among the question items. A low recall@5 suggests that the representation power of the user query was insufficient to retrieve the same-volume images. To examine the causes of this, we compared the user queries in Q.4 exhibiting the highest recall@5 ($0.90 \pm 0.32$) (see \textbf{\cref{fig:test_1_lung_cancer}a}) with those in Q.1 showing the lowest recall@5 ($0.10 \pm 0.32$) (see \textbf{\cref{fig:test_1_lung_cancer}b}). Because the user queries resembled each other among the evaluators for each question item, we concluded that skill-based limitations were not the primary cause. Instead, a \emph{sketch-based limitation}, which can be troublesome when a semantic sketch has insufficient representation power to specify the image characteristics of the disease phenotype, might have affected the variability of recall@5. This was because the single class of tumor-associated labels (i.e., PT) seemed insufficient for differentiating the internal characteristics of lung cancers, including ground-glass opacity and solid components. Meanwhile, the means $\pm$ standard deviations of precision@5, reciprocal rank, nDCG, and recall@5 of the query-by-example approach were $0.28 \pm 0.11$, $1.00 \pm 0.00$, $0.66 \pm 0.03$, and $1.00 \pm 0.00$, respectively. The consistently higher precision@5 (i.e., all precision@5 values in \textbf{\cref{fig:test_1}b} are above 0.28) suggests that our SBMIR system's overall performance for retrieving clinically similar images was still superior to the conventional CBMIR system despite the possible sketch-based limitation for lung cancers. 

\subsection{SBMIR enabled image retrieval without example images}\label{sec:sbmir_without_examples}

Test-2 demonstrated that our SBMIR system could retrieve clinically similar images even without example images. This would be impossible using a conventional CBMIR system based on the query-by-example approach, posing the usability problem. Five descriptions of clinical findings, which simulate radiology reports, were presented to the evaluators. Each evaluator constructed a user query according to clinical findings recalled by interpreting the description and evaluated the clinical similarity of retrieved images to the intention of their query using the user-oriented metrics.

For gliomas, the five descriptions included various types of tumors (see \textbf{\cref{fig:test_2}a} and \textbf{\ref{app:test-2_questions}}). The mean precision@5 ranged from 0.48 in Q.5 to 0.86 in Q.3. The mean reciprocal rank and mean nDCG were intermediate-to-high, exceeding 0.6 for all question items. Thus, most top-5 retrieved images were given high similarity scores, including more than two clinically similar images ranked high on average (see \textbf{\cref{sec:dif_clinical_similarity}} for the definitions of the similarity scores). As can be seen in the example results of Q.3 (see \textbf{\cref{fig:test_2_glioma}a}), even though each evaluator was presented with the description independently, the user queries resembled each other, retrieving clinically similar images as intended. Notably, our SBMIR system was effective in retrieving images with distinctive yet rare characteristics, for example, the so-called butterfly glioma in Q.5 (see \textbf{\cref{fig:test_2_glioma}b}). These promising results were obtained in a situation that required sufficiently detailed descriptions for the evaluators to recall common clinical findings, which makes the evaluation stricter and more complex than in Test-1. 

For lung cancers, in a more challenging task, we presented descriptions that specify the detailed anatomical location of the disease (see \textbf{\cref{fig:test_2}b} and \textbf{\ref{app:test-2_questions}}), including the lung segment, the apex of the lung, the pleural contact, and the hilar region. This is because the radiological phenotypes of lung cancers are so diverse that the information about the size and lung lobes alone would be insufficient for the evaluators to specify their user queries. Except for Q.4, the mean precision@5 (ranging from 0.46 in Q.1 to 0.84 in Q.3) and mean reciprocal rank (ranging from 0.77 in Q.5 to 1.00 in Q.3) showed intermediate-to-high values, suggesting that at least two clinically similar images were successfully ranked higher on average. See \textbf{\cref{fig:test_2_lung_cancer}a} for the results of successful image retrieval with relatively similar user queries in Q.3. In contrast, for particular cases, such as the central type of lung cancer in Q.4 (see \textbf{\cref{fig:test_2_lung_cancer}b}), even though the mean precision@5 of 0.24 exceeded 0.2 (i.e., as described in \textbf{\cref{sec:user_oriented_metrics}}, the lower bound of precision@5 will be 0.2 when image retrieval works properly), the mean reciprocal rank of 0.47 was still low. The detailed anatomical location, such as the hilar region, may not have been fully learned by the model, impeding localization of the area by normal ACs from the template images. We call this a \emph{template-image-based limitation}, which can hinder the image retrieval performance of our SBMIR system. 

\subsection{SBMIR enabled image retrieval for isolated samples}\label{sec:sbmir_isolated_samples}

Test-3 confirmed that our SBMIR system could obtain the same-volume images even when the isolated samples were presented, overcoming the searchability problem. See \textbf{\ref{app:isolated_samples}} for the details of how we identified the isolated samples, which were defined as images that are ``not'' included within the $k$-NN of any other images in the database. Five randomly selected isolated samples, identified based on the fine-tuned ResNet-101 features, were presented to the evaluators. Recall@5 was automatically evaluated for each isolated sample. 

For gliomas, the five isolated samples were presented to the evaluators (see \textbf{\cref{fig:test_3}a} and \textbf{\ref{app:test-3_questions}}). The mean recall@5 was more than or equal to 0.7, implying that most evaluators succeeded in retrieving the same-volume images. Example results in Q.1 with the highest recall@5 ($1.00 \pm 0.00$) and those in Q.2 with the lowest recall@5 ($0.70 \pm 0.48$) are presented in \textbf{\cref{fig:test_3_glioma}a} and \textbf{\cref{fig:test_3_glioma}b}, respectively. Despite the slightly different user queries, the same-volume images were effectively retrieved within the top-5 retrieved images, as highlighted by the red boxes. 

For lung cancers, we presented the five isolated samples with the evaluators (see \textbf{\cref{fig:test_3}b} and \textbf{\ref{app:test-3_questions}}). Recall@5 exceeded 0.5 for all question items, except for Q.2, indicating that more than half of the evaluators successfully retrieved the same-volume images. The example results in Q.4 showing the highest recall@5 ($0.89 \pm 0.33$) are shown in \textbf{\cref{fig:test_3_lung_cancer}a}. In contrast, the isolated sample in Q.2 was difficult to retrieve, as shown by the recall@5 of $0.44 \pm 0.53$ (see \textbf{\cref{fig:test_3_lung_cancer}b}). The low recall@5 for Q.2 may be related to a template-image-based limitation based on the observation that the body and chest wall of the presented image seems much larger than the corresponding template image, possibly hindering the representation power of the template image to encompass such individual anatomical differences. 

\section{Conclusion}\label{sec:conclusion}

Despite tremendous advancements in CBMIR \citep{Haq2021, Zhong2021, Fang2021, Rossi2021}, the practical aspect of information-seeking objectives of healthcare professionals has not been paid much attention, overlooking practical usability and searchability limitations. To overcome this, we developed the SBMIR system, which does not require preparing example images or sketching the entire anatomical appearance. The most innovative aspect is the simple two-step user operation (see \textbf{\cref{fig:sbmir_system}a}), which can specify fine-grained characteristics of image content. The user test showed that our SBMIR system could overcome previous limitations through better image retrieval performance according to fine-grained image characteristics (see \textbf{\cref{fig:test_1}}), image retrieval without example images (see \textbf{\cref{fig:test_2}}), and image retrieval for isolated samples (see \textbf{\cref{fig:test_3}}). Our SBMIR system enables users to retrieve images of interest on demand, expanding the utility of medical image databases.

Three possible sources of limitations of our SBMIR system were observed. Skill-based limitations seemed to be minimized by the practice stage. Sketch-based and template-image-based limitations are associated with the algorithm, indicating room for improvement. As malignant tumors are characterized not only by shape and size but also by internal characteristics \citep{Aerts2014}, learning internal characteristics may mitigate sketch-based limitations, which were particularly evident for lung cancers. Besides, modeling normal anatomy using detailed segmentation labels could reduce template-image-based limitations, while information about normal anatomy was learned in a self-supervised manner in the present study. Because our SBMIR system can be applied to other diseases as long as lesions are segmentable, the development of medical image retrieval technologies will hopefully be accelerated by considering these issues.

The present study reminds us that database searching is an inherently interactive process. Indeed, the bidirectional communication between the user and the system to refine user queries for better results has been a fundamental concern for practical information access \citep{Kherfi08, Miao2021}. However, in conventional query-by-example approaches, there has been no room for such interaction in the whole process. In our SBMIR system, on the other hand, users can seek a better way to express their user intention by observing how the search results change according to which template image is selected and how the disease is sketched (see \textbf{\cref{fig:how_sbmir_works}} and the \textbf{Supplementary Video 1}). Such human-machine interaction has the potential to establish reliable and trustworthy data-driven applications in medicine \citep{Cutillo2020, Liang2022}. 

\section*{Acknowledgement}
The authors thank the members of the Division of Medical AI Research and Development of the National Cancer Center Research Institute for their kind support. The RIKEN AIP Deep Learning Environment (RAIDEN) supercomputer system was used in this study to perform computations. 

\subsection*{Funding}
This work was supported by JST CREST (Grant Number JPMJCR1689), JST AIP-PRISM (Grant Number JPMJCR18Y4), JSPS Grant-in-Aid for Scientific Research on Innovative Areas (Grant Number JP18H04908), and JSPS KAKENHI (Grant Number JP22K07681).

\subsection*{Competing interests}
Kazuma Kobayashi and Ryuji Hamamoto have received research funding from Fujifilm Corporation.

\subsection*{Contributions}
K.K. conceived the study, devised the algorithms, developed the software, coordinated the user tests, performed the technical evaluation, and analyzed the results. K.K., L.G., and R. Hataya prepared the manuscript. K.K. and M.M. prepared the datasets and the question items for the user tests. K.K, T.M., M.M., and M.T. designed the framework of the user tests. K.K., T.M., M.M., H.W., M.T., Y.T., Y.Y., S.N., N.K., and A.B. participated in the user tests. Y.K. provided technical advice. T.H. and R. Hamamoto supervised the research.

\subsection*{Data availability}
The glioma dataset is available on the website of the MICCAI BraTS Challenge \citep{Menze2015, Bakas2017, TCGAGBM, TCGALGG}. Note that the collected chest CT scans from the hospital that were utilized in the lung cancer dataset remain under their custody. 

\subsection*{Code availability}
All source codes for the training and evaluation of the present study will be publicly available on GitHub (\url{https://github.com/Kaz-K/sketch-based-medical-image-retrieval}). 

\appendix

\section{Preparation of ResNet-101 to extract image-level features}\label{app:resnet_feature}
\setcounter{figure}{0}

We prepared two types of feature extractors that represent the characteristics of entire images: ImageNet-trained ResNet-101 and fine-tuned ResNet-101. ResNet-101 is a 101-layered deep neural network that produces a feature vector with 2,048 dimensions from a layer just before the final fully connected layer \citep{He2016}. Because ImageNet contains 1.28 million natural training images consisting of 1,000 classes \citep{Russakovsky2015}, ImageNet-trained ResNet-101 can be expected to take general image features into account; however, there are substantial differences between ImageNet classification and medical image diagnosis \citep{Raghu2019}. Hence, we fine-tuned the ImageNet-trained ResNet-101 using the training datasets (see \textbf{\cref{sec:datasets}}). For fine-tuning based on the training dataset, the final 1,000-node classification layer of the ResNet-101 was removed and replaced by a one-node layer to output a probability of whether the input image contains abnormal findings or not. As our datasets contain tumor-associated labels corresponding to a relatively small portion of areas in a large image of a body region (see \textbf{\cref{fig:dataset_examples}}), we expected that fine-tuning would force the model to focus on the local pathological change in the image rather than a global subject of the image. All the model parameters were optimized according to the training settings described previously \citep{Kobayashi2021}. For each training dataset, the models were trained for 100 epochs. The resultant image-wise classification performances for the abnormality of the fine-tuned ResNet-101 were as follows. The means $\pm$ standard deviations of accuracy, precision, recall, and specificity were $0.94 \pm 0.07$, $0.94 \pm 0.20$, $0.84 \pm 0.24$, and $0.97 \pm 0.08$ for the glioma testing dataset and $0.97 \pm 0.04$, $0.81 \pm 0.30$, $0.86 \pm 0.29$, and $0.98 \pm 0.04$ for the lung cancer testing dataset. 

\section{Identification of isolated samples in the conventional query-by-example approach}\label{app:isolated_samples}
\setcounter{figure}{0}

\begin{figure*}[t!]
  \centering
  \includegraphics[]{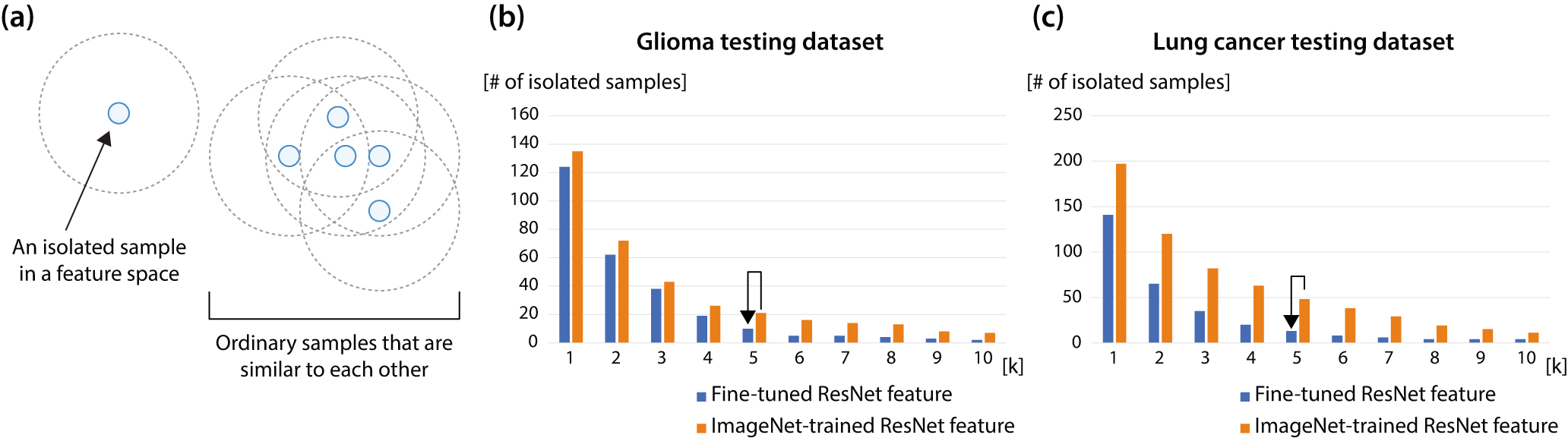}
  \caption{\textbf{Isolated samples in a database as a fundamental limitation of conventional query-by-example approaches.} \textbf{a} In a latent space, an ordinary sample can be surrounded by other ordinary samples, while an isolated sample with unique image characteristics is considered to be located far away from the others. Thus, when such isolated samples exist in a database, one may not be able to find them using the query-by-example approach because it is challenging to prepare an example image with similar unique characteristics. This can be quantitatively evaluated by counting the number of samples that do not appear in the $k$-nearest neighbors ($k$-NN) of any other samples in the latent space. \textbf{b} The number of isolated samples in the glioma testing dataset among $k$-NN samples. The fine-tuning of ResNet-101 reduced the number of isolated samples. For example, the number of isolated samples in the 5-NN group was reduced from 21 to 10 (indicated by the arrow). \textbf{c} The number of isolated samples in the lung cancer testing dataset among $k$-NN samples. Similarly, the number of isolated samples in the 5-NN group was reduced from 48 to 13 by fine-tuning (indicated by the arrow). Nevertheless, a substantial number of isolated samples existed in both databases, even when the fine-tuned feature extractor was used.}
  \label{fig:isolated_samples}
\end{figure*}

We defined an isolated sample as an image that is ``not'' included among the $k$-nearest neighbors ($k$-NN) of any other images (see \textbf{\cref{fig:isolated_samples}a}). To count the isolated samples in the testing datasets, we focused on representative images. A representative image is defined as the 2D axial slice containing the largest tumor-associated region in each 3D volume. This is because 3D volume data, such as CT and MRI scans, usually contain similar abnormal findings in consecutive slices, which can raise a concern about duplicative counting of similar images. We decided that the evaluation using the representative images is valid because the slice that best characterizes the clinical significance of 3D volume data is often the one in which a lesion appears largest. After identifying the representative images from each 3D volume, a feature extractor (i.e., ImageNet-trained ResNet-101 or fine-tuned ResNet-101, as described in \textbf{\ref{app:resnet_feature}}) extracted an image-level feature from each representative image. Then, the L2 distances with all the other representative images based on the feature vectors were calculated to obtain $k$-NN samples for each representative image. The number of representative images that were ``not'' included in the $k$-NN of all the other representative images was counted. The numbers of isolated samples in the glioma and lung cancer testing datasets are shown in \textbf{\cref{fig:isolated_samples}b} and \textbf{\cref{fig:isolated_samples}c}, respectively.

\section{Detailed network architecture}\label{app:detailed_network_architecture}
\setcounter{table}{0}
\setcounter{figure}{0}

For the glioma training dataset, the input image size was $3 \times 256 \times 256 (= 196,608)$ and the segmentation label size was $4 \times 256 \times 256$, which indicates three classes of tumor-associated regions (i.e., ET, ED, and NET) plus a background class. \textbf{\cref{tab:vae_encoder_network}} shows the detailed implementation of the normal AC encoder, which mainly consists of a repeated structure with residual blocks \citep{He2016} and a strided convolution (ResBlock + StridedConv). For the abnormal AC encoder, the network architecture demonstrated in \textbf{\cref{tab:encoder_network}} was employed. Almost the same network architecture as the \textbf{\cref{tab:encoder_network}} was used for the label encoder, except for its input image size of $4 \times 256 \times 256$, which is consistent with the segmentation label size. The image decoder employed the neural network architecture shown in \textbf{\cref{tab:decoder_network}}. For the label decoder, almost the same architecture in \textbf{\cref{tab:decoder_network}} was used, except for its final output size, which was adjusted to be $4 \times 256 \times 256$. 

For the lung cancer training dataset, the network architectures were the same, but the spatial resolutions of the output shape in \textbf{\cref{tab:vae_encoder_network}}, \textbf{\cref{tab:encoder_network}}, and \textbf{\cref{tab:decoder_network}} were doubled because the spatial resolution of the input image was doubled, to be $3 \times 512 \times 512 (= 786,432)$. Additionally, the segmentation label size was $2 \times 512 \times 512$, reflecting one class of tumor-associated region (i.e., PT) plus a background class. Because the input image was concatenated with adjacent upper and lower slices to reach a channel size of 3, the segmentation label corresponding to the center slice was set to be the learning objective. Consequently, the sizes of the ACs were $512 \times 2 \times 2 = 2,048$ for the glioma dataset and $512 \times 4 \times 4 = 8,192$ for the lung cancer dataset, maintaining the same compression ratio ($\nicefrac{2,048}{196,608} = \nicefrac{8,192}{786,432} =\nicefrac{1}{96}$) across the datasets irrespective of the difference in the spatial resolution of images. 

\begin{table}[t!]
\centering
\caption{\textbf{Basic architecture of the encoder networks of the variational autoencoder component}}
\resizebox{0.8\linewidth}{!}{%
\begin{tabular}{lcc}
\hline
\textbf{Module} & Activation & Output shape \\ \hline

\begin{tabular}[c]{@{}l@{}}Input image \\ Conv \\ StridedConv \\ \end{tabular}  & 
\begin{tabular}[c]{@{}c@{}}\\ $\begin{bmatrix} 3 \times 3 & 32\end{bmatrix}$\\ $\begin{bmatrix} 3 \times 3 & 64\end{bmatrix}$ \\ \end{tabular} & 
\begin{tabular}[c]{@{}c@{}}$3 \times 256 \times 256$ \\ $32 \times 256 \times 256$\\ $64 \times 128 \times 128$ \\ \end{tabular}
\\ \hline

\begin{tabular}[c]{@{}l@{}}ResBlock \\ \\ StridedConv \\ \end{tabular}  &
\begin{tabular}[c]{@{}c@{}}$\begin{bmatrix} 3 \times 3 & 64 \\ 3 \times 3 & 64\end{bmatrix}$\\ 
                           $\begin{bmatrix} 3 \times 3 & 128\end{bmatrix}$ \\ \end{tabular} &
\begin{tabular}[c]{@{}c@{}}$64 \times 128 \times 128$\\ \\ $128 \times 64 \times 64$ \\ \end{tabular}   
\\ \hline

\begin{tabular}[c]{@{}l@{}}ResBlock \\ \\ StridedConv \\ \end{tabular}  &
\begin{tabular}[c]{@{}c@{}}$\begin{bmatrix} 3 \times 3 & 128 \\ 3 \times 3 & 128\end{bmatrix}$\\ 
                           $\begin{bmatrix} 3 \times 3 & 256\end{bmatrix}$ \\ \end{tabular} &
\begin{tabular}[c]{@{}c@{}}$128 \times 64 \times 64$\\ \\ $256 \times 32 \times 32$ \\ \end{tabular}   
\\ \hline

\begin{tabular}[c]{@{}l@{}}ResBlock \\ \\ StridedConv \\ \end{tabular}  &
\begin{tabular}[c]{@{}c@{}}$\begin{bmatrix} 3 \times 3 & 256 \\ 3 \times 3 & 256\end{bmatrix}$\\ 
                           $\begin{bmatrix} 3 \times 3 & 512\end{bmatrix}$ \\ \end{tabular} &
\begin{tabular}[c]{@{}c@{}}$256 \times 32 \times 32$\\ \\ $512 \times 16 \times 16$ \\ \end{tabular}   
\\ \hline

\begin{tabular}[c]{@{}l@{}}ResBlock \\ \\ StridedConv \\ \end{tabular}  &
\begin{tabular}[c]{@{}c@{}}$\begin{bmatrix} 3 \times 3 & 512 \\ 3 \times 3 & 512\end{bmatrix}$\\ 
                           $\begin{bmatrix} 3 \times 3 & 512\end{bmatrix}$ \\ \end{tabular} &
\begin{tabular}[c]{@{}c@{}}$512 \times 16 \times 16$\\ \\ $512 \times 8 \times 8$ \\ \end{tabular}   
\\ \hline

\begin{tabular}[c]{@{}l@{}}ResBlock \\ \\ StridedConv \\ \end{tabular}  &
\begin{tabular}[c]{@{}c@{}}$\begin{bmatrix} 3 \times 3 & 512 \\ 3 \times 3 & 512\end{bmatrix}$\\ 
                           $\begin{bmatrix} 3 \times 3 & 512\end{bmatrix}$ \\ \end{tabular} &
\begin{tabular}[c]{@{}c@{}}$512 \times 8 \times 8$\\ \\ $512 \times 4 \times 4$ \\ \end{tabular}   
\\ \hline

\begin{tabular}[c]{@{}l@{}}ResBlock \\ \\ StridedConv \\ Split \\ \end{tabular}  &
\begin{tabular}[c]{@{}c@{}}$\begin{bmatrix} 3 \times 3 & 512 \\ 3 \times 3 & 512\end{bmatrix}$\\ 
                           $\begin{bmatrix} 3 \times 3 & 512\end{bmatrix}$ \\ - \\ \end{tabular} &
\begin{tabular}[c]{@{}c@{}}$512 \times 4 \times 4$\\ \\ $512 \times 2 \times 2$ \\ \\ \end{tabular}   
\\ \hline

\begin{tabular}[c]{@{}l@{}}Conv\\ \end{tabular}  &
\begin{tabular}[c]{@{}c@{}}$\begin{bmatrix} 1 \times 1 & 512 \end{bmatrix}$, $\begin{bmatrix} 1 \times 1 & 512 \end{bmatrix}$\\ \end{tabular} &
\begin{tabular}[c]{@{}c@{}}$512 \times 2 \times 2$, $512 \times 2 \times 2$ \\ \end{tabular}   
\\ \hline

\end{tabular}%
}
\label{tab:vae_encoder_network}
\end{table} 

\begin{table}[t!]
\centering
\caption{\textbf{Basic architecture of the encoder networks}}
\resizebox{0.55\linewidth}{!}{%
\begin{tabular}{lcc}
\hline
\textbf{Module} & Activation & Output shape \\ \hline

\begin{tabular}[c]{@{}l@{}}Input image \\ Conv \\ StridedConv \\ \end{tabular}  & 
\begin{tabular}[c]{@{}c@{}}\\ $\begin{bmatrix} 3 \times 3 & 32\end{bmatrix}$\\ $\begin{bmatrix} 3 \times 3 & 64\end{bmatrix}$ \\ \end{tabular} & 
\begin{tabular}[c]{@{}c@{}}$3 \times 256 \times 256$ \\ $32 \times 256 \times 256$\\ $64 \times 128 \times 128$ \\ \end{tabular}
\\ \hline

\begin{tabular}[c]{@{}l@{}}ResBlock \\ \\ StridedConv \\ \end{tabular}  &
\begin{tabular}[c]{@{}c@{}}$\begin{bmatrix} 3 \times 3 & 64 \\ 3 \times 3 & 64\end{bmatrix}$\\ 
                           $\begin{bmatrix} 3 \times 3 & 128\end{bmatrix}$ \\ \end{tabular} &
\begin{tabular}[c]{@{}c@{}}$64 \times 128 \times 128$\\ \\ $128 \times 64 \times 64$ \\ \end{tabular}   
\\ \hline

\begin{tabular}[c]{@{}l@{}}ResBlock \\ \\ StridedConv \\ \end{tabular}  &
\begin{tabular}[c]{@{}c@{}}$\begin{bmatrix} 3 \times 3 & 128 \\ 3 \times 3 & 128\end{bmatrix}$\\ 
                           $\begin{bmatrix} 3 \times 3 & 256\end{bmatrix}$ \\ \end{tabular} &
\begin{tabular}[c]{@{}c@{}}$128 \times 64 \times 64$\\ \\ $256 \times 32 \times 32$ \\ \end{tabular}   
\\ \hline

\begin{tabular}[c]{@{}l@{}}ResBlock \\ \\ StridedConv \\ \end{tabular}  &
\begin{tabular}[c]{@{}c@{}}$\begin{bmatrix} 3 \times 3 & 256 \\ 3 \times 3 & 256\end{bmatrix}$\\ 
                           $\begin{bmatrix} 3 \times 3 & 512\end{bmatrix}$ \\ \end{tabular} &
\begin{tabular}[c]{@{}c@{}}$256 \times 32 \times 32$\\ \\ $512 \times 16 \times 16$ \\ \end{tabular}   
\\ \hline

\begin{tabular}[c]{@{}l@{}}ResBlock \\ \\ StridedConv \\ \end{tabular}  &
\begin{tabular}[c]{@{}c@{}}$\begin{bmatrix} 3 \times 3 & 512 \\ 3 \times 3 & 512\end{bmatrix}$\\ 
                           $\begin{bmatrix} 3 \times 3 & 512\end{bmatrix}$ \\ \end{tabular} &
\begin{tabular}[c]{@{}c@{}}$512 \times 16 \times 16$\\ \\ $512 \times 8 \times 8$ \\ \end{tabular}   
\\ \hline

\begin{tabular}[c]{@{}l@{}}ResBlock \\ \\ StridedConv \\ \end{tabular}  &
\begin{tabular}[c]{@{}c@{}}$\begin{bmatrix} 3 \times 3 & 512 \\ 3 \times 3 & 512\end{bmatrix}$\\ 
                           $\begin{bmatrix} 3 \times 3 & 512\end{bmatrix}$ \\ \end{tabular} &
\begin{tabular}[c]{@{}c@{}}$512 \times 8 \times 8$\\ \\ $512 \times 4 \times 4$ \\ \end{tabular}   
\\ \hline

\begin{tabular}[c]{@{}l@{}}ResBlock \\ \\ StridedConv \\ \end{tabular}  &
\begin{tabular}[c]{@{}c@{}}$\begin{bmatrix} 3 \times 3 & 512 \\ 3 \times 3 & 512\end{bmatrix}$\\ 
                           $\begin{bmatrix} 3 \times 3 & 512\end{bmatrix}$ \\ \end{tabular} &
\begin{tabular}[c]{@{}c@{}}$512 \times 4 \times 4$\\ \\ $512 \times 2 \times 2$ \\ \end{tabular}   
\\ \hline

\end{tabular}%
}
\label{tab:encoder_network}
\end{table} 

\begin{table}[t!]
\centering
\caption{\textbf{Basic architecture of the decoder networks}}
\resizebox{0.65\linewidth}{!}{%
\begin{tabular}{lcc}
\hline
\textbf{Module} & Activation & Output shape \\ \hline

\begin{tabular}[c]{@{}l@{}}Latent representation \\ \end{tabular}  & 
\begin{tabular}[c]{@{}c@{}} - \\ \end{tabular} & 
\begin{tabular}[c]{@{}c@{}}$512 \times 2 \times 2$ \\ \end{tabular}
\\ \hline

\begin{tabular}[c]{@{}l@{}}Conv \\ ResBlock \\ \\ \end{tabular}  &
\begin{tabular}[c]{@{}c@{}}$\begin{bmatrix} 3 \times 3 & 512\end{bmatrix}$\\ 
                           $\begin{bmatrix} 3 \times 3 & 512 \\ 3 \times 3 & 512\end{bmatrix}$ \\ \end{tabular} &
\begin{tabular}[c]{@{}c@{}}$512 \times 2 \times 2$\\ $512 \times 2 \times 2$ \\ \\ \end{tabular}   
\\ \hline

\begin{tabular}[c]{@{}l@{}}Conv \\ Upsample \\ ResBlock \\ \\ \end{tabular}  &
\begin{tabular}[c]{@{}c@{}}$\begin{bmatrix} 1 \times 1 & 512\end{bmatrix}$\\ - \\
                           $\begin{bmatrix} 3 \times 3 & 512 \\ 3 \times 3 & 512\end{bmatrix}$ \\ \end{tabular} &
\begin{tabular}[c]{@{}c@{}}$512 \times 2 \times 2$\\ $512 \times 4 \times 4$ \\ $512 \times 4 \times 4$ \\ \\ \end{tabular}   
\\ \hline

\begin{tabular}[c]{@{}l@{}}Conv \\ Upsample \\ ResBlock \\ \\ \end{tabular}  &
\begin{tabular}[c]{@{}c@{}}$\begin{bmatrix} 1 \times 1 & 512\end{bmatrix}$\\ - \\
                           $\begin{bmatrix} 3 \times 3 & 512 \\ 3 \times 3 & 512\end{bmatrix}$ \\ \end{tabular} &
\begin{tabular}[c]{@{}c@{}}$512 \times 4 \times 4$\\ $512 \times 8 \times 8$ \\ $512 \times 8 \times 8$ \\ \\ \end{tabular}   
\\ \hline

\begin{tabular}[c]{@{}l@{}}Conv \\ Upsample \\ ResBlock \\ \\ \end{tabular}  &
\begin{tabular}[c]{@{}c@{}}$\begin{bmatrix} 1 \times 1 & 256\end{bmatrix}$\\ - \\
                           $\begin{bmatrix} 3 \times 3 & 256 \\ 3 \times 3 & 256\end{bmatrix}$ \\ \end{tabular} &
\begin{tabular}[c]{@{}c@{}}$256 \times 8 \times 8$\\ $256 \times 16 \times 16$ \\ $256 \times 16 \times 16$ \\ \\ \end{tabular}   
\\ \hline

\begin{tabular}[c]{@{}l@{}}Conv \\ Upsample \\ ResBlock \\ \\ \end{tabular}  &
\begin{tabular}[c]{@{}c@{}}$\begin{bmatrix} 1 \times 1 & 128\end{bmatrix}$\\ - \\
                           $\begin{bmatrix} 3 \times 3 & 128 \\ 3 \times 3 & 128\end{bmatrix}$ \\ \end{tabular} &
\begin{tabular}[c]{@{}c@{}}$128 \times 16 \times 16$\\ $128 \times 32 \times 32$ \\ $128 \times 32 \times 32$ \\ \\ \end{tabular}   
\\ \hline

\begin{tabular}[c]{@{}l@{}}Conv \\ Upsample \\ ResBlock \\ \\ \end{tabular}  &
\begin{tabular}[c]{@{}c@{}}$\begin{bmatrix} 1 \times 1 & 64\end{bmatrix}$\\ - \\
                           $\begin{bmatrix} 3 \times 3 & 64 \\ 3 \times 3 & 64\end{bmatrix}$ \\ \end{tabular} &
\begin{tabular}[c]{@{}c@{}}$64 \times 32 \times 32$\\ $64 \times 64 \times 64$ \\ $64 \times 64 \times 64$ \\ \\ \end{tabular}   
\\ \hline

\begin{tabular}[c]{@{}l@{}}Conv \\ Upsample \\ ResBlock \\ \\ \end{tabular}  &
\begin{tabular}[c]{@{}c@{}}$\begin{bmatrix} 1 \times 1 & 32\end{bmatrix}$\\ - \\
                           $\begin{bmatrix} 3 \times 3 & 32 \\ 3 \times 3 & 32\end{bmatrix}$ \\ \end{tabular} &
\begin{tabular}[c]{@{}c@{}}$32 \times 64 \times 64$\\ $32 \times 128 \times 128$ \\ $32 \times 128 \times 128$ \\ \\ \end{tabular}   
\\ \hline

\begin{tabular}[c]{@{}l@{}}Conv \\ Upsample \\ ResBlock \\ \\ \end{tabular}  &
\begin{tabular}[c]{@{}c@{}}$\begin{bmatrix} 1 \times 1 & 3\end{bmatrix}$\\ - \\
                           $\begin{bmatrix} 3 \times 3 & 3 \\ 3 \times 3 & 3\end{bmatrix}$ \\ \end{tabular} &
\begin{tabular}[c]{@{}c@{}}$3 \times 128 \times 128$\\ $3 \times 256 \times 256$ \\ $3 \times 256 \times 256$ \\ \\ \end{tabular}   
\\ \hline

\end{tabular}%
}
\label{tab:decoder_network}
\end{table} 

\section{Training results of the feature extraction module of our SBMIR system}\label{app:training_results}
\setcounter{figure}{0}

We trained the feature extraction module of our SBMIR system using the glioma training dataset and the lung cancer training dataset independently (see \textbf{\cref{sec:hyparam_glioma_dataset}} and \textbf{\cref{sec:hyparam_lung_cancer_dataset}}). We qualitatively and quantitatively evaluated its training results to verify the algorithm from technical viewpoints. 

\subsection{Qualitative evaluation of the image reconstruction}\label{app:qual_eval_image_recon}

\begin{figure}[t!]
  \centering
  \includegraphics[]{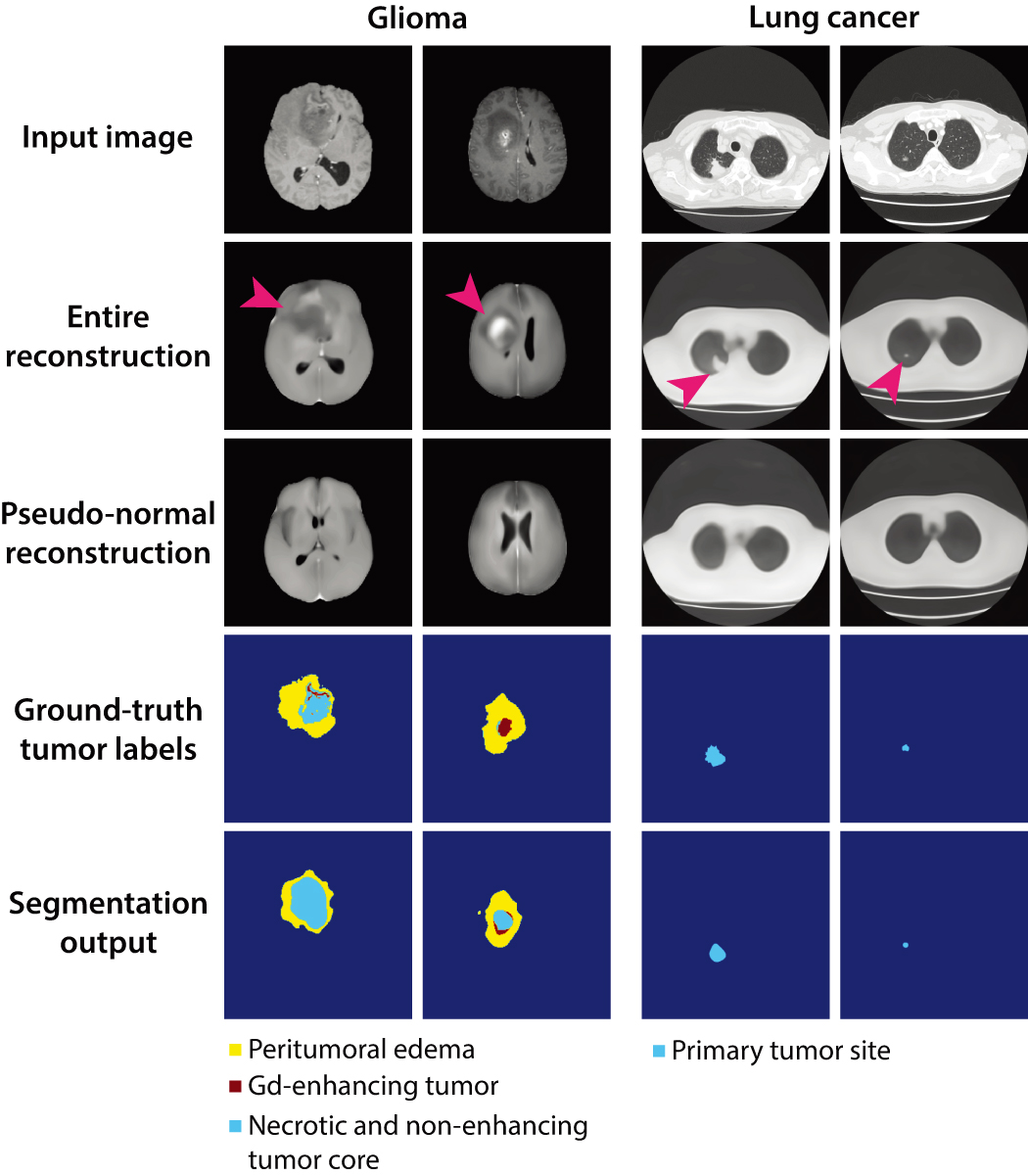}
  \caption{\textbf{Training results of the image reconstruction and segmentation.} The input images, the entire reconstructions from entire anatomy codes (ACs), and the pseudo-normal reconstructions from normal ACs are presented in the first, second, and third rows, respectively. The ground-truth and predicted segmentation labels are shown in the fourth and fifth rows, respectively. Arrows indicate the diseased areas in the entire reconstructions, which are diminished in the pseudo-normal reconstructions. Gd, gadolinium.}
  \label{fig:results_image_reconstruction}
\end{figure}

How the feature decomposition was achieved can be assessed by visualizing images generated by the image decoder and the label decoder. Recall that the reconstructed images from the normal ACs should be pseudo-normal images, while those from the entire ACs should be entire images with some abnormalities if they exist (see \textbf{\cref{fig:query_by_sketch}b}). Additionally, the abnormal ACs produce segmentation labels for the tumor-associated regions when they are inputted into the label decoder. 

\textbf{\Cref{fig:results_image_reconstruction}} presents the images and segmentation labels decoded from the ACs using the glioma testing dataset and the lung cancer testing dataset. The first row shows the input images. The second row demonstrates the entire reconstructions of the input images, which were decoded by the image decoder taking the entire ACs as input. The third row indicates the pseudo-normal reconstructions, which were decoded by the image decoder taking the normal ACs as input. Note that the abnormal imaging features that appear both in the input images and in the entire reconstructions (see the arrows in \textbf{\cref{fig:results_image_reconstruction}}) are diminished in the pseudo-normal images, recovering the normal anatomy that should have existed therein. These results suggest successful feature decomposition. Moreover, the fourth row presents the ground-truth segmentation labels for the tumor-associated regions, and the fifth row shows the predicted segmentation labels that were decoded by the label decoder taking the abnormal ACs as input. Note that the segmentation was trained only for the tumor-associated regions and not for the normal anatomy-associated regions, as shown in the results.

One may argue that the quality of the reconstructed images and segmentation labels was insufficient, as observed from the blurred and rounded appearance that did not recover the detailed image characteristics. These tendencies are reasonable because the image information is compressed because of the limited size of the latent representation, which raises a trade-off between the reconstruction qualities and latent size \citep{Razavi2019, Kobayashi2021}. Indeed, we did not pursue the generation quality of the reconstructions as a primary purpose because the lower dimension of the latent representations can be advantageous for computational efficiency in similarity search. Besides, although the detailed part of the image characteristics was not perfectly reproduced, it was still sufficient for recognizing the anatomical location and presence of abnormalities in the reconstructed images. 

\subsection{Qualitative evaluation of the semantically organized latent space}\label{app:qual_eval_semantic_latent_space}

\begin{figure*}[t!]
  \centering
  \includegraphics[]{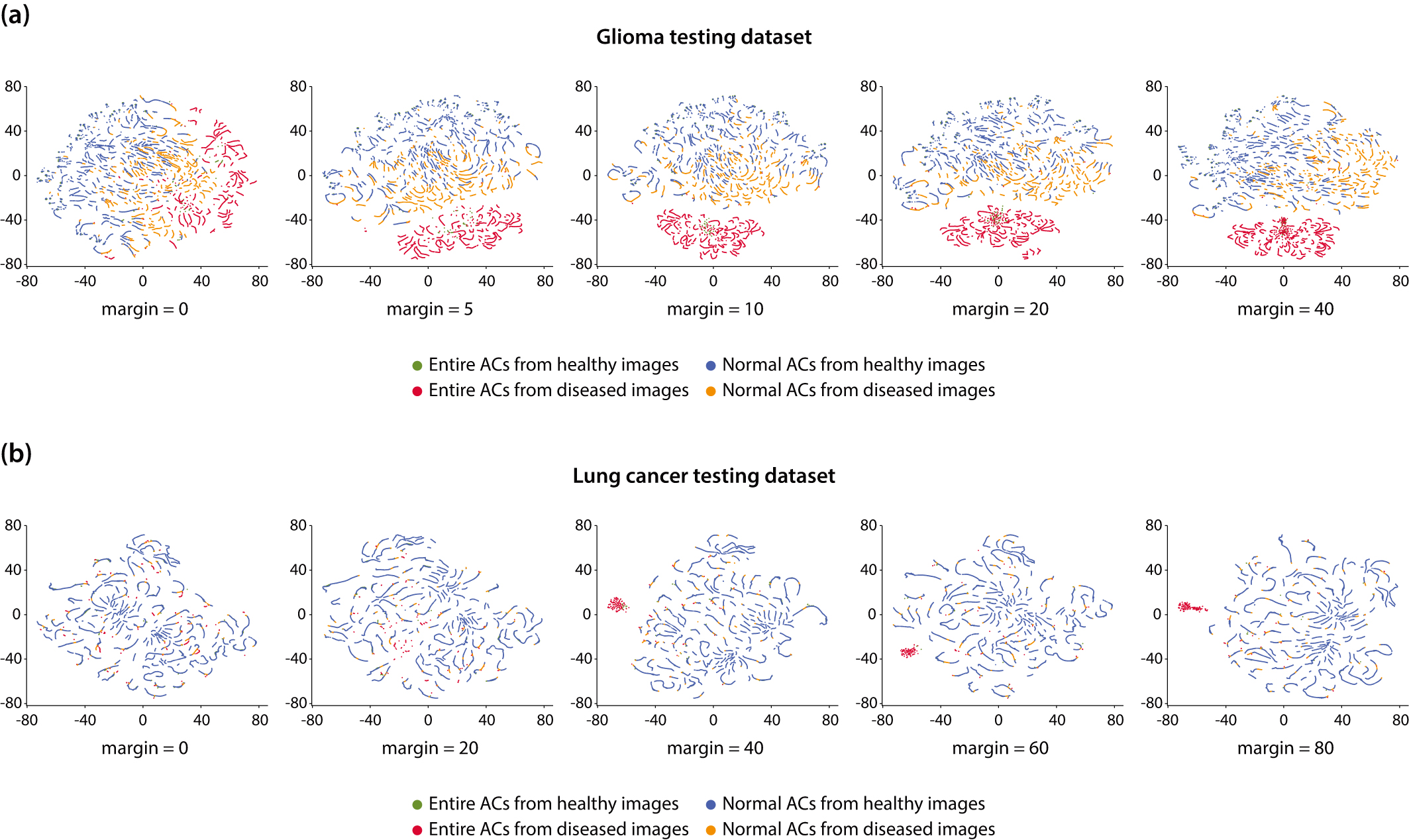}
  \caption{\textbf{Relationship between the margin parameter and cluster formation in the latent space.} Four latent distributions, the entire anatomy codes (ACs) from healthy images (green dots), the normal ACs from healthy images (blue dots), the normal ACs from diseased images (orange dots), and the entire ACs from diseased images (red dots) are visualized using t-distributed stochastic neighbor embedding (t-SNE) plotting. \textbf{a} In the glioma testing dataset, two clusters emerged with a margin parameter of more than 5. \textbf{b} In the lung cancer testing dataset, two clusters emerged when the margin parameter was more than 40. Note that the entire ACs from diseased images (red dots) were intermingled with the other data points, particularly with a margin parameter of 0, possibly reflecting the relatively small area of each lung cancer.}
  \label{fig:latent_space_organization}
\end{figure*}

We applied t-distributed stochastic neighbor embedding (t-SNE) analysis \citep{Laurens2008} to evaluate how the latent space was organized according to the semantics by changing the margin parameter (see \textbf{\cref{sec:config_diseased_subspace}}). We randomly selected 50 individual volumes from each dataset and extracted entire ACs and normal ACs in an image-wise manner. The once randomly selected samples were used repeatedly in the following t-SNE analyses for the purpose of comparison, particularly in \textbf{\ref{app:ablation_studies}}.

The results using the glioma testing dataset and the lung cancer testing dataset are shown in \textbf{\cref{fig:latent_space_organization}a} and \textbf{\cref{fig:latent_space_organization}b}, respectively. When the margin parameter was set to 0, there was no visible cluster in the latent space, where the different ACs intermingled with each other (see the leftmost images in \textbf{\cref{fig:latent_space_organization}a} and \textbf{\cref{fig:latent_space_organization}b}). This is particularly evident in \textbf{\cref{fig:latent_space_organization}b}, possibly reflecting that the primary site of lung cancer usually occupies such a small region that a latent feature representing an abnormality does not convey meaningful information when no margin parameter is assigned. On the other hand, when the margin parameter was increased, two clusters appeared and moved away from each other, particularly when the margin parameter was more than 10 for the glioma testing dataset and more than 40 for the lung cancer testing dataset. One cluster consisting of the entire ACs from healthy images (green dots), the normal ACs from healthy images (blue dots), and the normal ACs from diseased images (orange dots) can be interpreted as corresponding to the healthy subspace. In contrast, the other cluster consisting of the entire ACs from diseased images (red dots) can be considered the diseased distribution. Therefore, sufficiently large margin parameters are necessary for configuring the semantically organized latent space (see \textbf{\cref{fig:latent_space}b}).

\subsection{Quantitative evaluation of the image reconstruction}\label{app:quant_eval_image_recon}

Given entire ACs, the image decoder was trained to generate entire reconstructions $\hat{\bm{x}}$ of input images $\bm{x}$. Then, the reconstruction error between the entire reconstructions $\hat{\bm{x}}$ and input images $\bm{x}$ was evaluated as $\| \hat{\bm{x}} - \bm{x} \|_{2}^{2}$. The means $\pm$ standard deviations of the reconstruction error were $0.22 \pm 0.23$ and $0.16 \pm 0.05$ in the glioma testing dataset and lung cancer testing dataset, respectively. 

\subsection{Quantitative evaluation of the segmentation}\label{app:quant_eval_segmentation}

The label decoder predicts segmentation labels of tumor-associated regions $\hat{\bm{l}}$, which should be close to the ground-truth label $\bm{l}$. The segmentation performance was evaluated based on the Dice score with respect to the tumor-associated labels. To calculate the Dice score for each volume, the segmentation outputs of each 2D axial image were concatenated into a 3D volume. The means $\pm$ standard deviations of the Dice score were $0.37 \pm 0.21$, $0.60 \pm 0.14$, and $0.44 \pm 0.21$ for NET, ED, and ET, respectively, in the glioma testing dataset. The mean $\pm$ standard deviation of the Dice score was $0.64 \pm 0.18$ for PT in the lung cancer testing dataset. The intermediate levels of these Dice scores were reasonable because the model did not have skip-connections, which is essential for precise medical image segmentation \citep{Drozdzal2016}, in order to concentrate the information on the diseased regions in the abnormal ACs.

\subsection{Quantitative evaluation of the mapping function from the label space to the latent space}\label{app:quant_eval_mapping_func}

We evaluated the performance of the label encoder taking the ground-truth labels as input to estimate the corresponding abnormal ACs $\hat{\bm{a}}$, which should be similar to the output of the abnormal AC encoder $\bm{a}$ taking the corresponding image as input. The means $\pm$ standard deviations of L2 distance between the output of the label encoder and the corresponding abnormal ACs $\| \hat{\bm{a}} - \bm{a} \|_{2}^{2}$ were $3.1 \times 10^{-2} \pm 4.5 \times 10^{-2}$ and $0.7 \times 10^{-2} \pm 3.6\times 10^{-2}$ in the glioma testing dataset and the lung cancer testing dataset, respectively. As the L2 distance is small, the inverse mapping from the semantic labels to the corresponding abnormal ACs can be precise, enabling users to specify the characteristics of abnormal findings of interest by drawing semantic sketches. 

\subsection{Quantitative evaluation of the separability in latent space}\label{app:quant_eval_separability}

To quantitatively evaluate how well the latent space $\mathcal{Z}$ can be separated into the healthy subspace $\mathcal{Z}^\mathrm{h}$ and the diseased subspace $\mathcal{Z}^\mathrm{d}$ (see \textbf{\cref{fig:latent_space}b} and \textbf{\cref{fig:latent_space_organization}}), we trained support vector machines (SVMs) to learn the support vector between the entire ACs from healthy images $\mathcal{D}(\bm{e} \rvert \bm{x}^\mathrm{h})$ and the entire ACs from diseased images $\mathcal{D}(\bm{e} \rvert \bm{x}^\mathrm{d})$ based on the training datasets (i.e., the glioma training dataset and the lung cancer training dataset). Because the SVM is expected to learn the hyperplane that has the maximum margin between the two distributions, its classification performance reflects how well the two distributions are separated. In the glioma testing dataset, the classification performance of the SVM was as follows: accuracy, 0.94; precision, 0.96; recall, 0.89; F-score, 0.92. In the lung cancer testing dataset, the classification performance of the SVM was as follows: accuracy, 0.97; precision, 0.90; recall, 0.82; F-score, 0.86. We consider that the high classification performance of the SVMs implies that the configuration of the semantically organized latent space was successfully achieved. 

\section{Technological evaluation focusing on the image retrieval performance}\label{app:eval_image_retrieval}
\setcounter{figure}{0}

Here, we demonstrate that extending the feature decomposition of medical images by imposing the semantically organized latent space is critical to achieving the practical SBMIR system. The practical SBMIR system should realize image retrieval according to the similarity of images while simultaneously reflecting the semantics. Note that in image retrieval reflecting semantics, a query vector conveying the information about a diseased region should retrieve only diseased images and that not conveying the information about a diseased region should retrieve only healthy images. From this perspective, we devised two evaluation methods to objectively demonstrate how the retrieved image features can preserve information regarding both the semantics and similarity of images. The image retrieval performance was observed to be affected by the margin parameter for configuring the semantically organized latent space (see \textbf{\cref{fig:latent_space}b}). 

\subsection{Semantic consistency in image retrieval}\label{app:semantic_consistency}

\begin{figure*}[t!]
  \centering
  \includegraphics[]{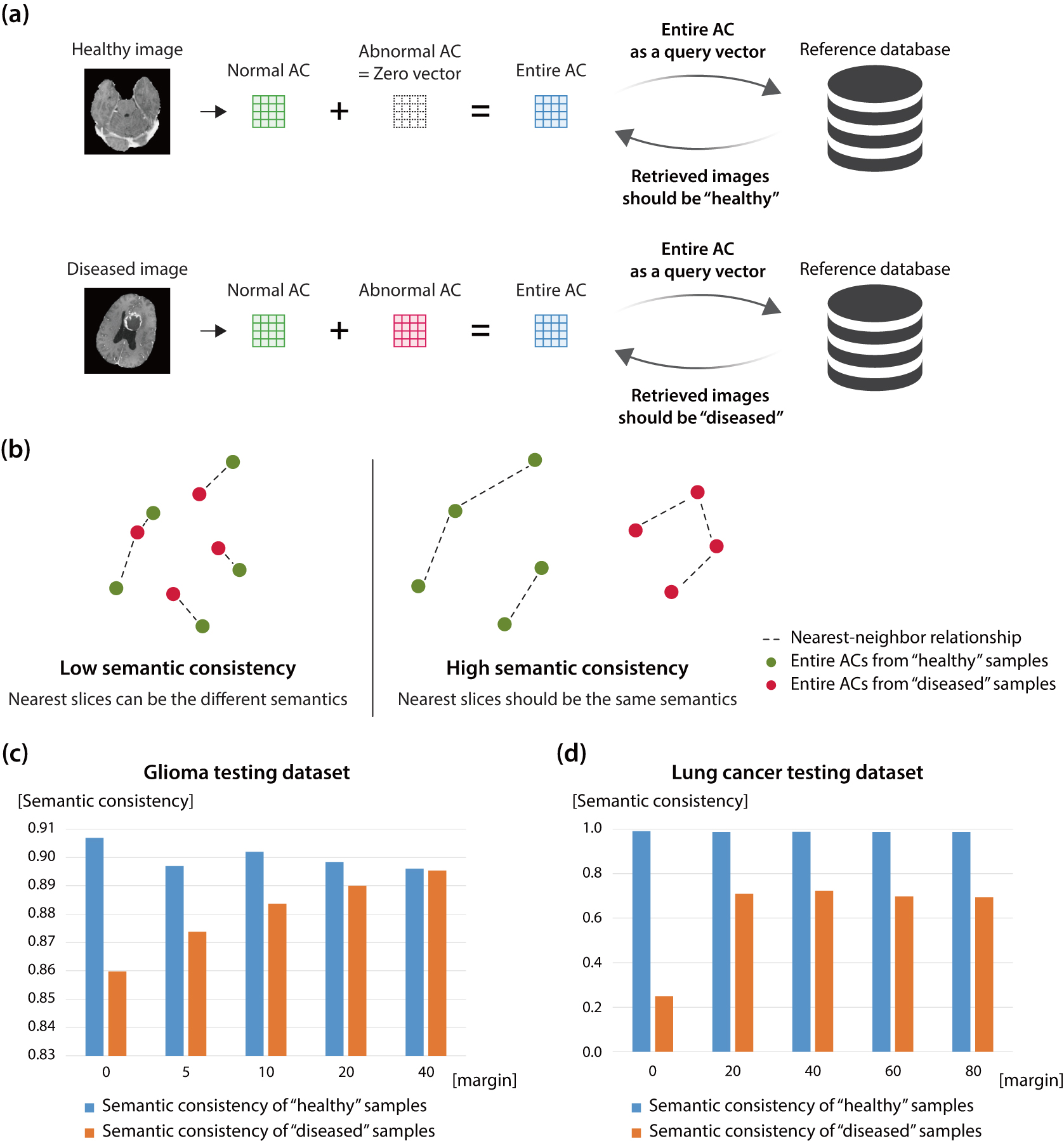}
  \caption{\textbf{Evaluation of the semantic consistency in image retrieval.} \textbf{a} When image retrieval can be realized according to the semantics of an image, healthy or diseased, a query vector as an entire anatomy code (AC) from a healthy or diseased image should retrieve healthy or diseased images, respectively. \textbf{b} To achieve semantic consistency, the latent space should be semantically organized, with data points with similar semantics located close to each other. \textbf{c} The semantic consistency in the glioma testing dataset. \textbf{d} The semantic consistency in the lung cancer testing dataset. Note that a higher margin parameter is essential for maintaining semantic consistency, as is particularly evident in the lung cancer testing dataset.}
  \label{fig:semantic_consistency}
\end{figure*}

\begin{figure*}[t!]
  \centering
  \includegraphics[]{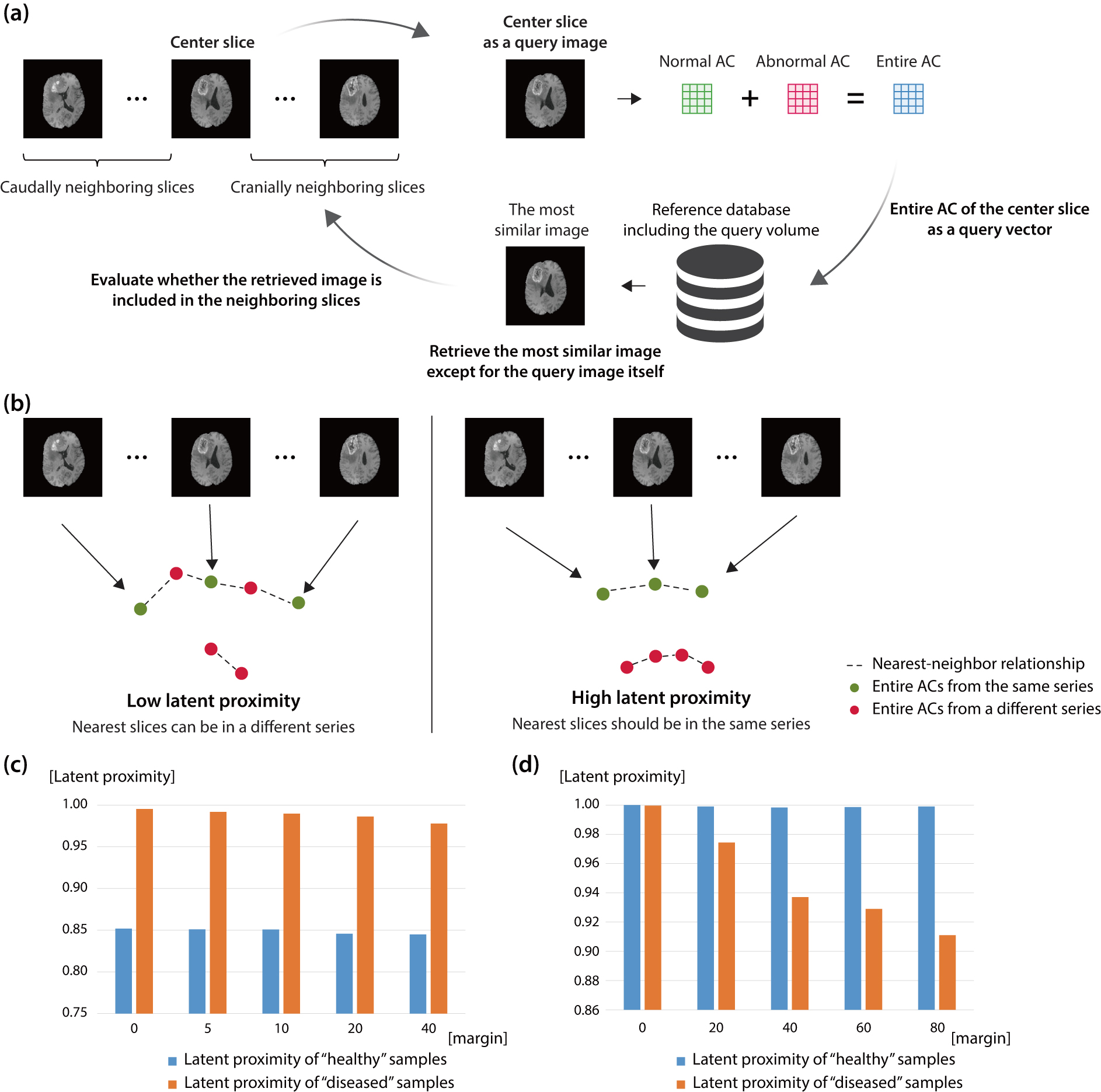}
  \caption{\textbf{Evaluation of latent proximity in image retrieval.} Efficient medical image retrieval requires similar images to be near each other in latent space. \textbf{a} To quantitatively evaluate latent proximity, we noted that the contiguous slices in an image volume are similar to each other. Then, whether a query vector as an entire anatomy code (AC) from a center image can retrieve the slices that were cranially or caudally consecutive to the center image was evaluated. \textbf{b} Low latent proximity indicates that images that are originally similar to each other are mixed in the latent space with dissimilar images, whereas high latent proximity suggests the similarity in the image space is also preserved in the latent space. \textbf{c} The latent proximity was evaluated using the glioma testing dataset. \textbf{d} The latent proximity was evaluated using the lung cancer testing dataset. Too large a margin parameter was found to adversely affect latent proximity, especially for the lung cancer testing dataset.}
  \label{fig:latent_proximity}
\end{figure*}

To evaluate how well the image retrieval is performed according to semantics, we developed an evaluation method to quantify semantic consistency. As shown in \textbf{\cref{fig:semantic_consistency}a}, image retrieval reflecting semantics should retrieve healthy or diseased images with an entire AC from a healthy or diseased image, respectively, when the entire AC is used as a query vector. To achieve this, each sample in the latent space should be grouped with samples with the same semantics rather than samples with different semantics (see \textbf{\cref{fig:semantic_consistency}b}), as is expected to be configured by the margin parameter. Therefore, we evaluated semantic consistency based on the following procedure. First, each image in a testing dataset was considered as a query image, and the corresponding entire AC was calculated as a query vector. Using each query vector, the five closest slices from different volumes other than the one the query vector belongs to were retrieved. Then, the ratio of retrieved images whose semantics are consistent with the query vector was evaluated. The resultant ratio was reported as the semantic consistency value, which was calculated according to the semantics of the query images (i.e., healthy images or diseased images). 

In the glioma testing dataset (see \textbf{\cref{fig:semantic_consistency}c}), the semantic consistency of diseased images increased from 0.86 to 0.88 as the margin parameter increased from 0 to 10. More remarkable effects were observed in the lung cancer testing dataset (see \textbf{\cref{fig:semantic_consistency}d}). When the margin parameter was set to 0, the semantic consistency of diseased images was as small as 0.25. This indicates that even if the user intention is to retrieve images with abnormalities, the retrieval results can be mixed with healthy images. In contrast, the semantic consistency was improved to 0.72 when the margin parameter was increased to 40, showing that the margin parameter had a meaningful effect on the semantic consistency of diseased images. As the average tumor volume in the lung cancer testing dataset was significantly smaller than that in the glioma testing dataset (see \textbf{\cref{sec:training_settings}}), we inferred that the latent features representing such small diseased areas could be overwhelmed by those representing the large image of a body region. Consequently, the margin parameter is essential for image retrieval with semantic consistency, effectively conveying the imaging features relevant to abnormal findings in medical image retrieval. 

\subsection{Latent proximity in image retrieval}\label{app:latent_proximity}

To evaluate how the similarity in the image space is preserved in latent space, we devised an index called \emph{latent proximity} based on the notion that similar images should also be nearby in the latent space, which was evaluated as follows (see \textbf{\cref{fig:latent_proximity}a}). First, each image in a testing dataset was considered as a query image, and the corresponding entire AC was calculated as a query vector. Based on each query vector, the closest slice from all volumes, including the original volume the query vector belongs to, was retrieved. Then, whether the retrieved image was included in the five cranially or caudally consecutive slices in the original volume was assessed according to the semantics of the query image. Finally, the ratio between the number of retrieved images not included in the consecutive slices and those included in the consecutive slices was calculated as the latent proximity value. As shown in \textbf{\cref{fig:latent_proximity}b}, low latent proximity indicates that even similar images are not in close proximity in latent space, which can cause unintended search results with dissimilar images. In contrast, high latent proximity can guarantee that similarity in image space is also reproduced in latent space, leading to retrieval results that are faithful to the user intention. The evaluation was repeated by varying the margin parameter in the analysis of both the glioma and lung cancer testing datasets. 

As shown in \textbf{\cref{fig:latent_proximity}c}, the latent proximity of the glioma testing dataset does not differ dramatically according to the magnitude of the margin parameter. On the other hand, the latent proximity of diseased images in the lung cancer testing dataset was clearly changed by the margin parameter (see \textbf{\cref{fig:latent_proximity}d}). Therefore, a margin parameter that is too large may worsen the correspondence between the image space and the latent space, hindering image retrieval performance according to the similarity of images. 

\section{Ablation studies of the feature extraction module of our SBMIR system}\label{app:ablation_studies}
\setcounter{figure}{0}

Among several loss functions used in the training of the feature extraction module (i.e., the reconstruction loss, segmentation loss, consistency loss, abnormality loss, regularization loss, and margin loss) (see \textbf{\cref{sec:learning_objectives}}), the necessity and the optimal values of the margin parameter are demonstrated in the previous sections. Here, we confirm that the abnormality loss and regularization loss are also essential for semantically organized latent space. For simplicity, only the results based on the glioma testing dataset are shown.

\subsection{Ablation study of abnormality loss}\label{app:ablation_abn_loss}

\begin{figure}[t!]
  \centering
  \includegraphics[]{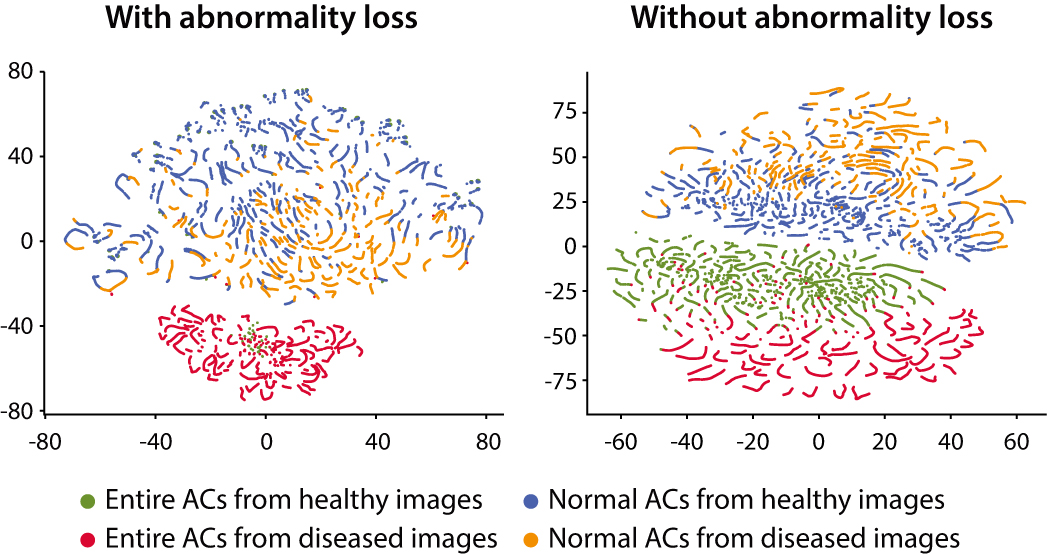}
  \caption{\textbf{Effects of abnormality loss on the organization of latent space.} Four latent distributions--namely, the entire anatomy codes (ACs) from healthy images (green dots), the normal ACs from healthy images (blue dots), the normal ACs from diseased images (orange dots), and the entire ACs from diseased images (red dots)--are visualized using t-distributed stochastic neighbor embedding (t-SNE) plotting. Without abnormality loss, the entire ACs from healthy images (green dots) moved away from the Gaussian distribution representing normal anatomy, rendering the latent space disorganized according to the image semantics.}
  \label{fig:abnormality_loss}
\end{figure}

Abnormality loss forces the abnormal AC encoder that takes healthy images to output zero vectors, as the abnormal ACs from healthy images should not contain any relevant information (see \textbf{\cref{fig:latent_space}a}). Because segmentation output from the subsequent label encoder is also trained to be zero vectors reflecting the absence of the tumor-associated regions, one can argue that the abnormality loss is duplicative. However, when training the model without abnormality loss, the entire ACs from healthy images (green dots) moved away from the Gaussian distribution as shown in \textbf{\cref{fig:abnormality_loss}}. This may be because the abnormal ACs, even from healthy images, had significant norms when trained without abnormality loss. In this case, a discrepancy emerged between the entire ACs from healthy images and the normal ACs from healthy images, which weakened the rationale that the former distribution represents the normal anatomy, undermining the assumption of the semantically organized latent space (see \textbf{\cref{fig:latent_space}b}).

\subsection{Ablation study of regularization loss}\label{app:ablation_reg_loss}

\begin{figure}[t!]
  \centering
  \includegraphics[]{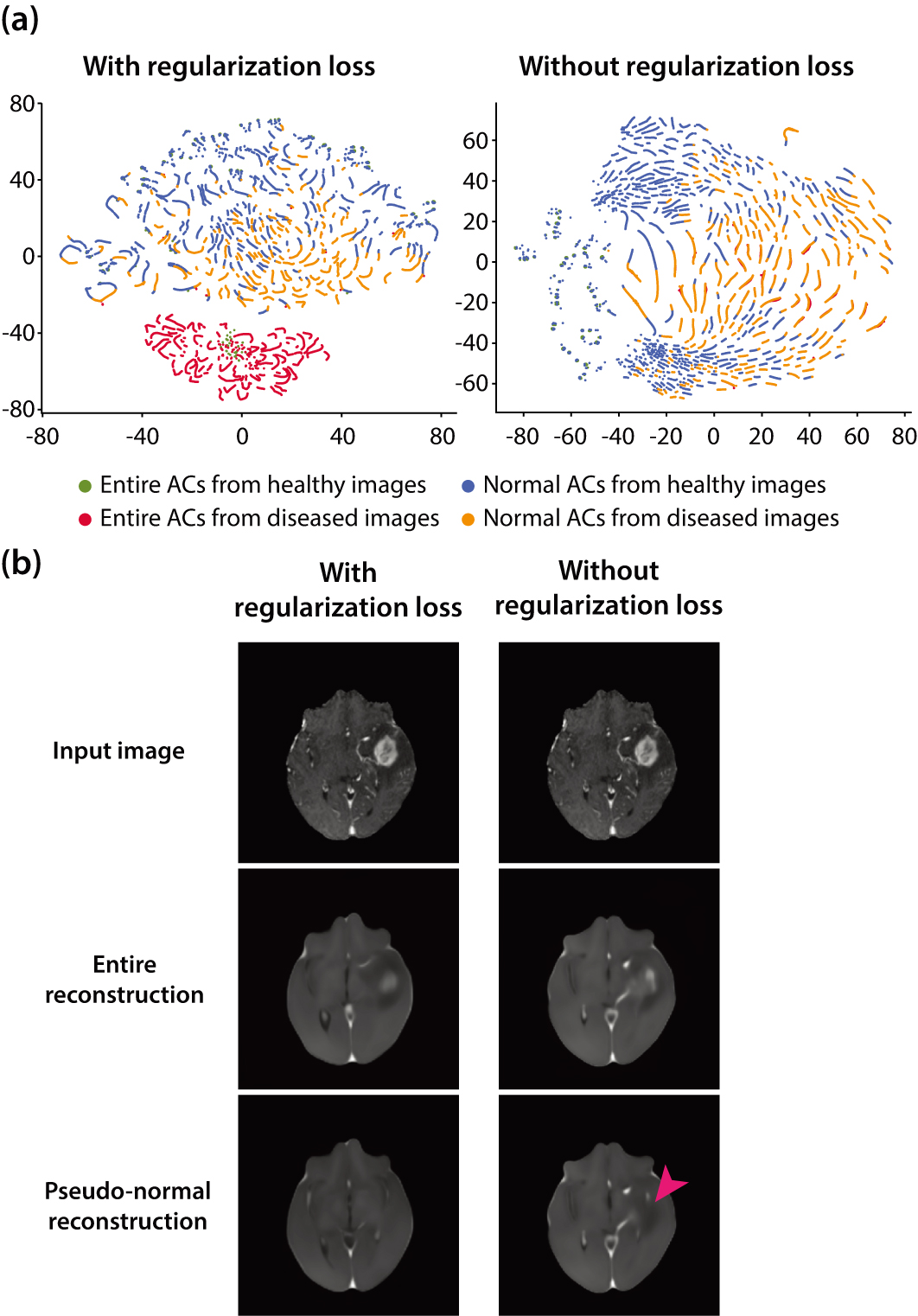}
  \caption{\textbf{Effects of regularization loss.} \textbf{a} Four latent distributions--namely, the entire anatomy codes (ACs) from healthy images (green dots), the normal ACs from healthy images (blue dots), the normal ACs from diseased images (orange dots), and the entire ACs from diseased images (red dots)--are visualized using t-distributed stochastic neighbor embedding (t-SNE) plotting. Without regularization loss to force the distribution of normal ACs from healthy images to follow the Gaussian distribution, the latent space lost its organization according to the image semantics. \textbf{b} Due to the lack of the prior distribution, the pair of the normal AC encoder and the image decoder acted like an identity function. Thereby, the reconstructed image, even from the normal ACs, contained abnormal imaging features (see the arrowhead).}
  \label{fig:regularization_loss}
\end{figure}

Regularization loss plays a role in regularizing the distribution of the normal ACs from healthy images to follow the Gaussian distribution, which is essential for training the VAE component (see \textbf{\cref{fig:basic_components}a}). One assumption of the VAE component is that normal anatomic variation can be modeled according to the Gaussian distribution, which is analogous to the fact that many medical indicators, such as body heights, are known to follow the Gaussian distribution. Here, we also show that the assumption is essential for the semantically organized latent space (see \textbf{\cref{fig:latent_space}b}). When the regularization loss was not imposed on the distribution of normal ACs, the latent space did not contain clustering as demonstrated in \textbf{\cref{fig:regularization_loss}a}. Moreover, owing to the lack of the prior distribution, the pair of the normal AC encoder and the image decoder (i.e., the neural networks trained as the VAE component) acted like an identity function, resulting in image reconstructions with abnormal findings even from the normal ACs (see the arrowhead in \textbf{\cref{fig:regularization_loss}b}). In summary, the assumption of the Gaussian distribution for the normal ACs is critical in our implementation not only for the semantically organized latent space but also for the feature decomposition of medical images. 

\section{Description of question items for the user tests}\label{app:description_question_items}
\setcounter{figure}{0}

\subsection{Question items in Test-1}\label{app:test-1_questions}

For gliomas, the clinical characteristics of the five presented images are as follows (see \textbf{\cref{fig:test_1}a}): 
\begin{itemize}\small
  \item Q.1: A 95-mm non-enhancing tumor with peritumoral edema in the left frontal lobe.
  \item Q.2: A 90-mm ring-enhancing tumor with multiple cores with peritumoral edema in the left frontal lobe.
  \item Q.3: A 70-mm non-enhancing tumor in the left temporal lobe.
  \item Q.4: A 50-mm ring-enhancing tumor with peritumoral edema in the right parietal lobe.
  \item Q.5: An 85-mm ring-enhancing tumor with peritumoral edema in the right temporal lobe.
\end{itemize}

For lung cancers, the clinical characteristics of the five presented images are as follows (see \textbf{\cref{fig:test_1}b}): 
\begin{itemize}\small
  \item Q.1: A 25-mm part-solid nodule in the left upper lobe.
  \item Q.2: A 30-mm tumor in the left lower lobe.
  \item Q.3: A 10-mm solid nodule in the right upper lobe.
  \item Q.4: A 45-mm tumor in the right middle lobe.
  \item Q.5: A 35-mm tumor in the right lower lobe.
\end{itemize}

\subsection{Question items in Test-2}\label{app:test-2_questions}

For gliomas, the five descriptions presented to the evaluators are as follows (see \textbf{\cref{fig:test_2}a}): 
\begin{itemize}\small
  \item Q.1: A 60-mm ring-enhancing tumor is located primarily in the left temporal lobe. It is associated with massive peritumoral edema (about 100 mm in maximum length) extending through the left temporal lobe.
  \item Q.2: A 50-mm non-enhancing tumor is located in the right frontal lobe. It is associated with mild peritumoral edema (about 70 mm in maximum length).
  \item Q.3: A 25-mm ring-enhancing tumor is located in the left temporal pole (the tip of the left temporal lobe). It is associated with mild peritumoral edema (about 40 mm in maximum length).
  \item Q.4: A 30-mm ring-enhancing tumor is localized in the right occipital lobe. It is associated with peritumoral edema extending anteriorly (about 60 mm in maximum length).
  \item Q.5: A 60-mm ring-enhancing tumor is located in the midline of the bilateral frontal lobes. It is associated with extensive peritumoral edema in the bilateral frontal lobes (about 90 mm in maximum length).
\end{itemize}

For lung cancers, the five descriptions presented to the evaluators are as follows (see \textbf{\cref{fig:test_2}b}): 
\begin{itemize}\small
  \item Q.1: A 20-mm nodule in the left upper lobe that contacts the pleura on the mediastinal side.
  \item Q.2: A 40-mm tumor in the posterior-basal segment of the left lower lobe in contact with the pleura.
  \item Q.3: A 50-mm tumor in the apex of the right lung.
  \item Q.4: A 70-mm tumor in the right pulmonary hilar region, possibly invading the mediastinum.
  \item Q.5: A 20-mm peripheral nodule in the right lateral-basal segment in contact with the chest wall pleura.
\end{itemize}

\subsection{Question items in Test-3}\label{app:test-3_questions}

For gliomas, the clinical characteristics of the five isolated samples presented to the evaluators are as follows (see \textbf{\cref{fig:test_3}a}): 
\begin{itemize}\small
  \item Q.1: A 65-mm ring-enhancing tumor with peritumoral edema located in the deep white matter in the right parietal lobe.
  \item Q.2: A 90-mm ring-enhancing tumor with extensive edema located primarily in the right frontal lobe.
  \item Q.3: A 45-mm enhancing tumor with peritumoral edema located primarily in the left insular lobe.
  \item Q.4: A 70-mm non-enhancing tumor with peritumoral edema in the left frontal lobe.
  \item Q.5: A 35-mm ring-enhancing tumor with peritumoral edema located at the anterior edge of the left temporal lobe.
\end{itemize}

For lung cancers, the clinical characteristics of the five isolated samples presented to the evaluators are as follows (see \textbf{\cref{fig:test_3}b}): 
\begin{itemize}\small
  \item Q.1: A 25-mm part-solid nodule in the left upper lobe.
  \item Q.2: A 65-mm tumor in the left lower lobe.
  \item Q.3: A 65-mm tumor in the right lower lobe.
  \item Q.4: A 70-mm tumor in the left lower lobe.
  \item Q.5: A 55-mm tumor in the right lower lobe.
\end{itemize}

\section{Example results of the user tests}\label{app:example_results}
\setcounter{figure}{0}

\subsection{Example results of Test-1}

\begin{figure*}[p!]
  \centering
  \includegraphics[]{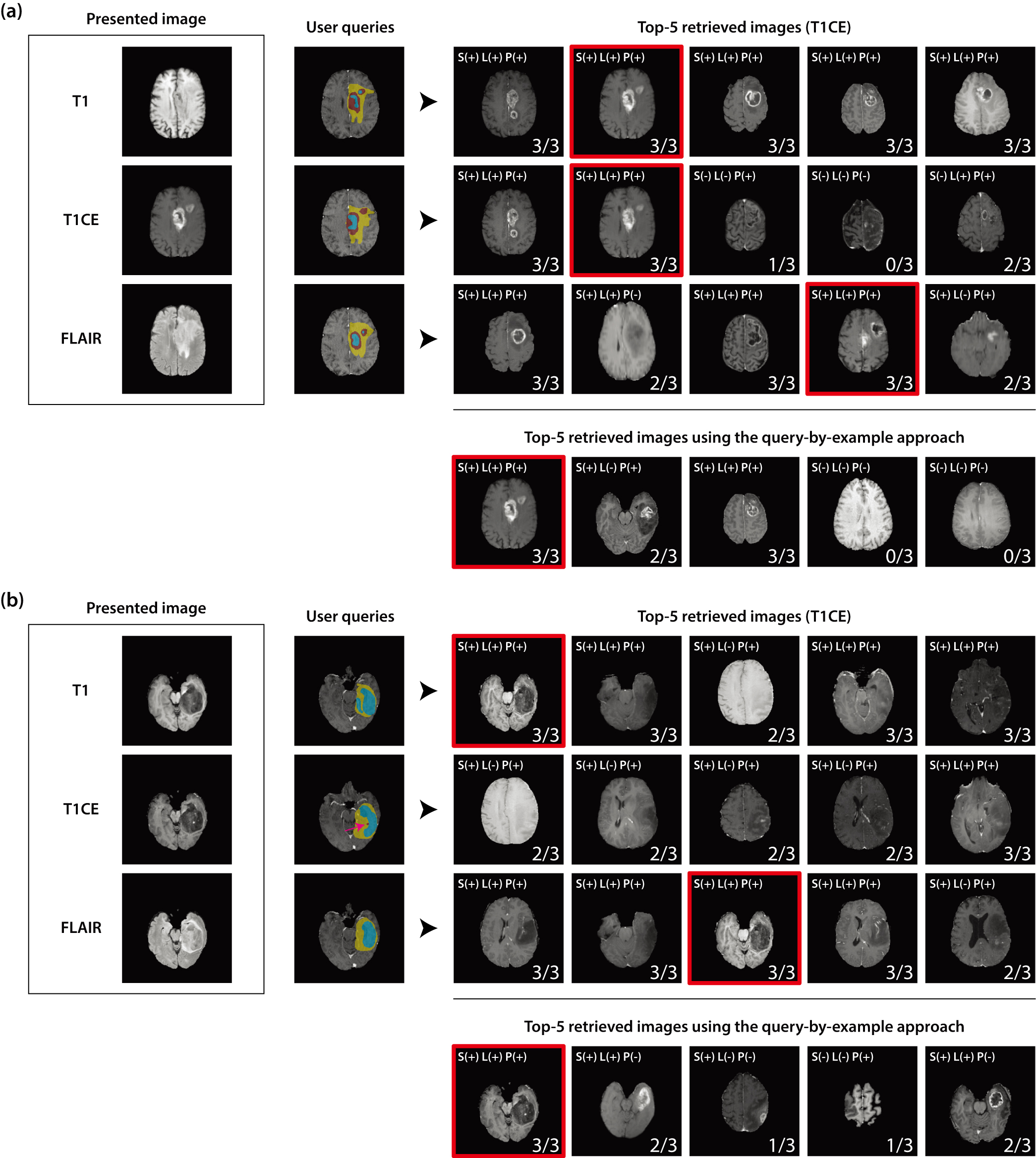}
  \caption{\textbf{Example results of Test-1 for gliomas.} \textbf{a} The presented image in Q.2 of Test-1 showed the highest recall@5 for gliomas. Three example user queries and corresponding top-5 retrieved images are shown. The retrieved images are arranged from left to right, starting with the most similar image. The letters S, L, and P in the upper left corner of each retrieved image indicate the judged consistencies of the size (S), location (L), and pattern of contrast enhancement (P), respectively. The number in the lower right corner indicates the given similarity score, with a maximum score of 3/3. The retrieved images highlighted with red boxes are the images belonging to the same volume as the presented images (i.e., the same-volume images). The corresponding retrieval result by the query-by-example approach is shown at the bottom. \textbf{b} The presented image in Q.3 of Test-1 showed the lowest recall@5 for gliomas. A failed case is shown in the middle row, where none of the retrieved images are highlighted in a red box. This is presumably the effect of overdiagnosis of the point-like contrast-enhanced region in the tumor drawn only in the middle user query (see the arrow), consistent with a skill-based limitation. T1, T1-weighted sequence; T1CE, T1-weighted contrast-enhanced sequence; FLAIR, fluid-attenuated inversion recovery sequence.}
  \label{fig:test_1_glioma}
\end{figure*}

\begin{figure*}[p!]
  \centering
  \includegraphics[]{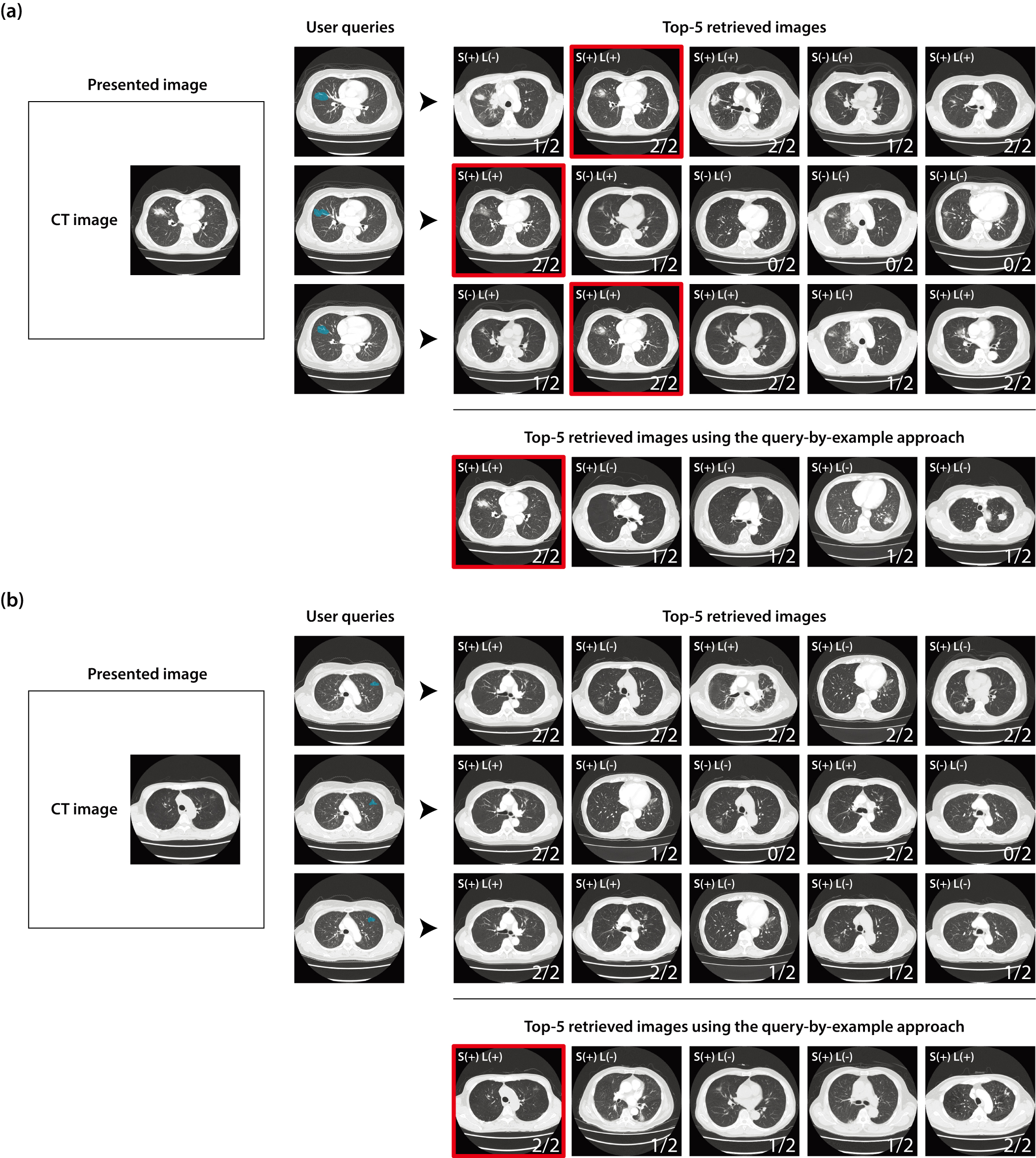}
  \caption{\textbf{Example results of Test-1 for lung cancers.} \textbf{a} The presented image in Q.4 of Test-1 showed the highest recall@5 for lung cancers. Three example user queries and corresponding top-5 retrieved images are shown. The retrieved images are arranged from left to right, starting with the most similar image. The letters S and L in the upper left corner of each retrieved image indicate the judged consistencies of the size (S) and location (L), respectively. The number in the lower right corner indicates the given similarity score, with a maximum score of 2/2. The retrieved images highlighted with red boxes are the images belonging to the same volume as the presented images (i.e., the same-volume images). The corresponding retrieval result by the query-by-example approach is shown at the bottom. \textbf{b} The presented image in Q.1 of Test-1 showed the lowest recall@5 for lung cancers. All three cases shown here failed to retrieve the presented images, which may be caused by a sketch-based limitation, meaning that the single class of tumor-associated labels was insufficient to express the variable internal characteristics of lung cancers. CT, computed tomography.}
  \label{fig:test_1_lung_cancer}
\end{figure*}

Example results of Test-1 for gliomas and those for lung cancers are shown in \textbf{\cref{fig:test_1_glioma}} and \textbf{\cref{fig:test_1_lung_cancer}}, respectively.

\subsection{Example results of Test-2}

\begin{figure*}[ht]
  \centering
  \includegraphics[width=\textwidth]{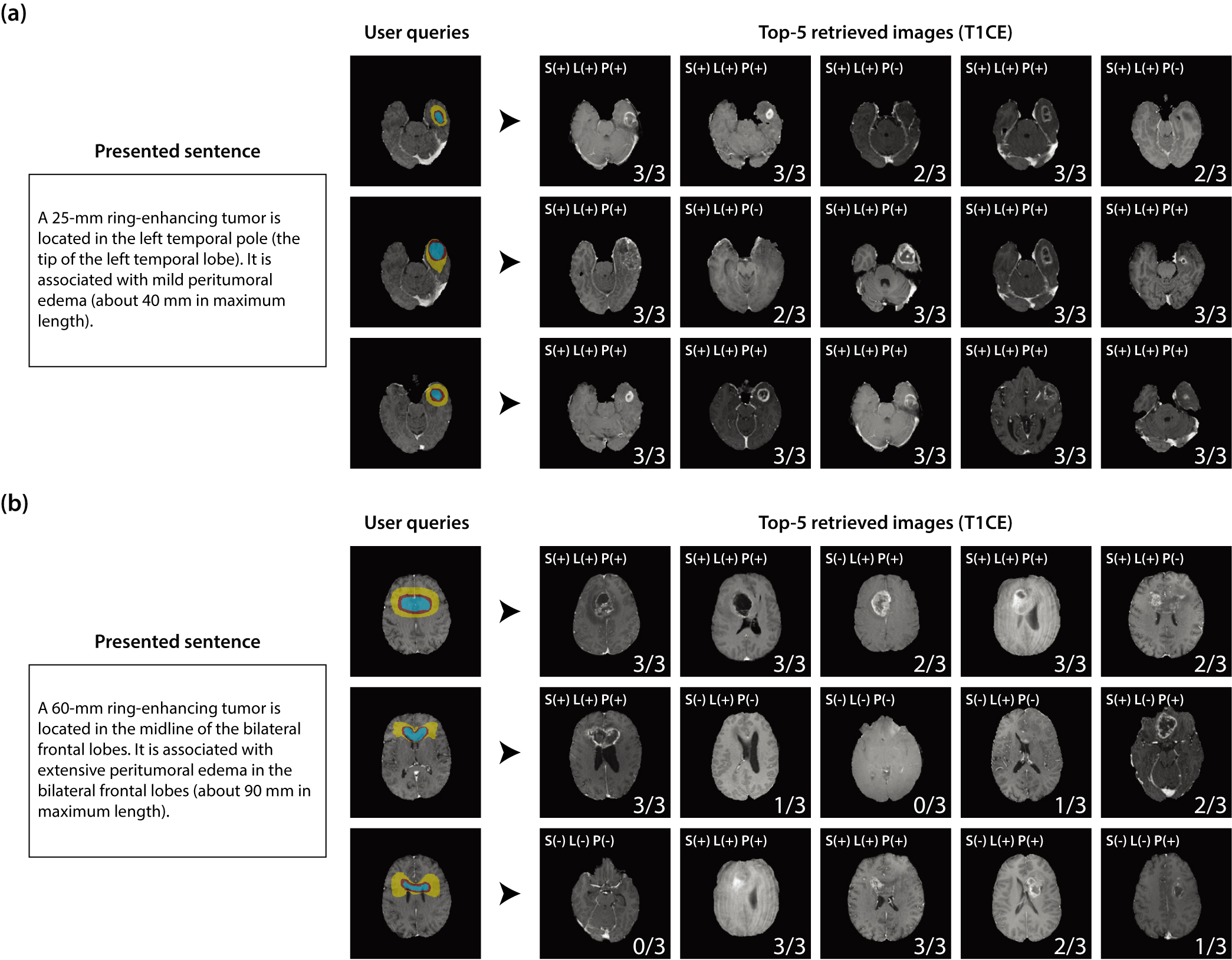}
  \caption{\textbf{Example results of Test-2 for gliomas.} \textbf{a} The presented image in Q.3 of Test-2 showed the highest precision@5 for gliomas. Three example user queries and corresponding top-5 retrieved images are shown. The retrieved images are arranged from left to right, starting with the most similar image. The letters S, L, and P in the upper left corner of each retrieved image indicate the judged consistencies of the size (S), location (L), and pattern of contrast enhancement (P), respectively. The number in the lower right corner indicates the given similarity score, with a maximum score of 3/3. Note that even though the evaluators were presented with the same sentence independently, the user queries resembled each other in terms of their successful retrieval of clinically similar images, as shown by the similarity scores of 3/3. \textbf{b} The presented image in Q.5 of Test-2 showed the lowest precision@5 for gliomas. It can be seen that the semantic sketches differed in terms of how each evaluator expressed the tumor extension beyond the corpus callosum, which could represent a skill-based limitation if this difference had influenced the search results. T1CE, T1-weighted contrast-enhanced sequence.}
  \label{fig:test_2_glioma}
\end{figure*}

\begin{figure*}[ht]
  \centering
  \includegraphics[width=\textwidth]{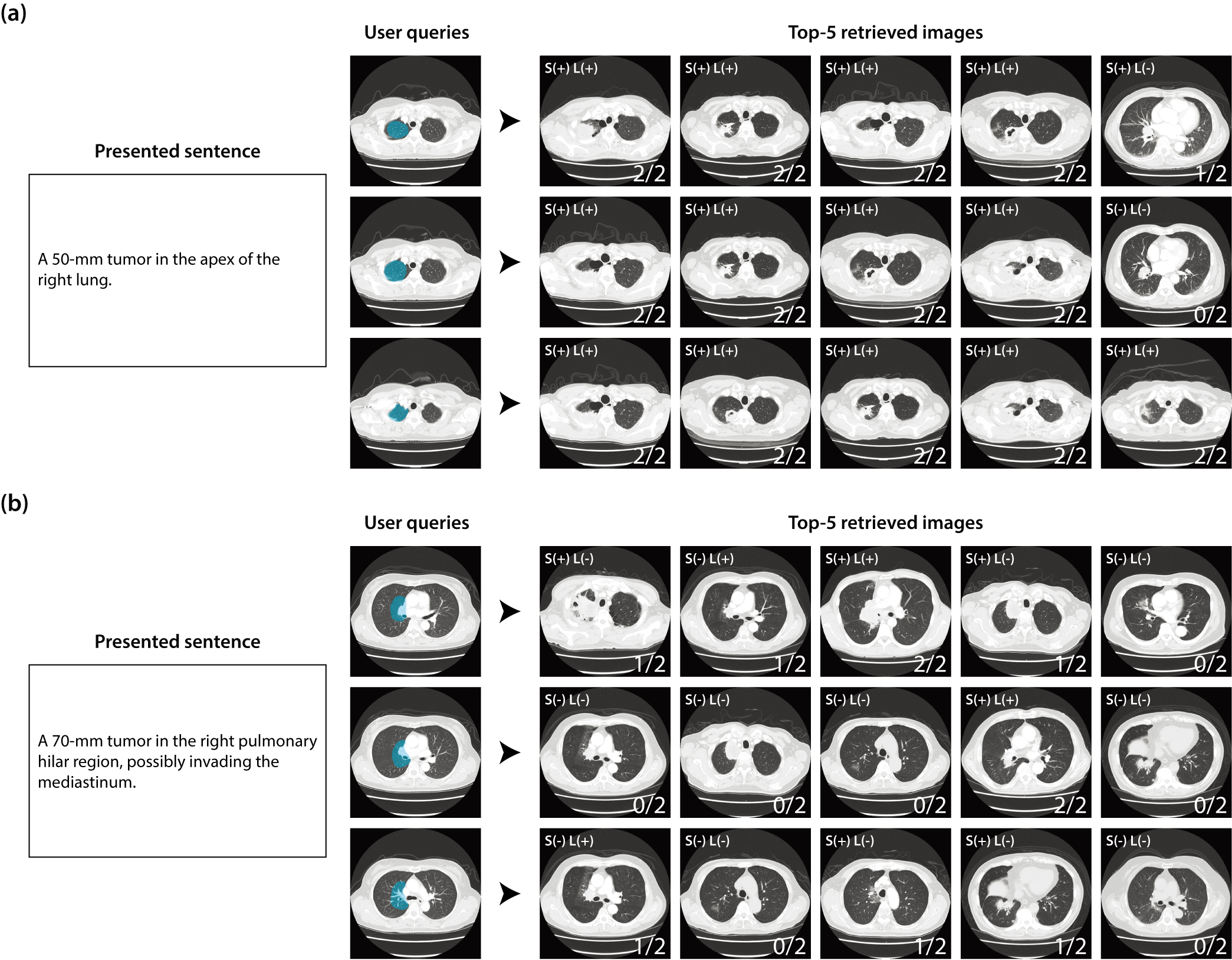}
  \caption{\textbf{Example results of Test-2 for lung cancers.} \textbf{a} The presented image in Q.3 of Test-2 showed the highest precision@5 for lung cancers. Three example user queries and corresponding top-5 retrieved images are shown. The retrieved images are arranged from left to right, starting with the most similar image. The letters S and L in the upper left corner of each retrieved image indicate the judged consistencies of the size (S) and location (L), respectively. The number in the lower right corner indicates the given similarity score, with a maximum score of 2/2. Note that even though the evaluators were presented with the same sentence independently, the user queries resembled each other and achieved successful retrieval of clinically similar images, as shown by the similarity scores of 2/2. \textbf{b} The presented image in Q.4 of Test-2 showed the lowest precision@5 for lung cancers. The low precision@5 may have been caused by a template-image-based limitation, suggesting that such detailed anatomical location, such as the pulmonary hilar region, was not completely learned by the model, which was trained in a self-supervised manner to learn the normal anatomy.}
  \label{fig:test_2_lung_cancer}
\end{figure*}

Example results of Test-2 for gliomas and those for lung cancers are shown in \textbf{\cref{fig:test_2_glioma}} and \textbf{\cref{fig:test_2_lung_cancer}}, respectively.

\subsection{Example results of Test-3}

\begin{figure*}[ht]
  \centering
  \includegraphics[width=\textwidth]{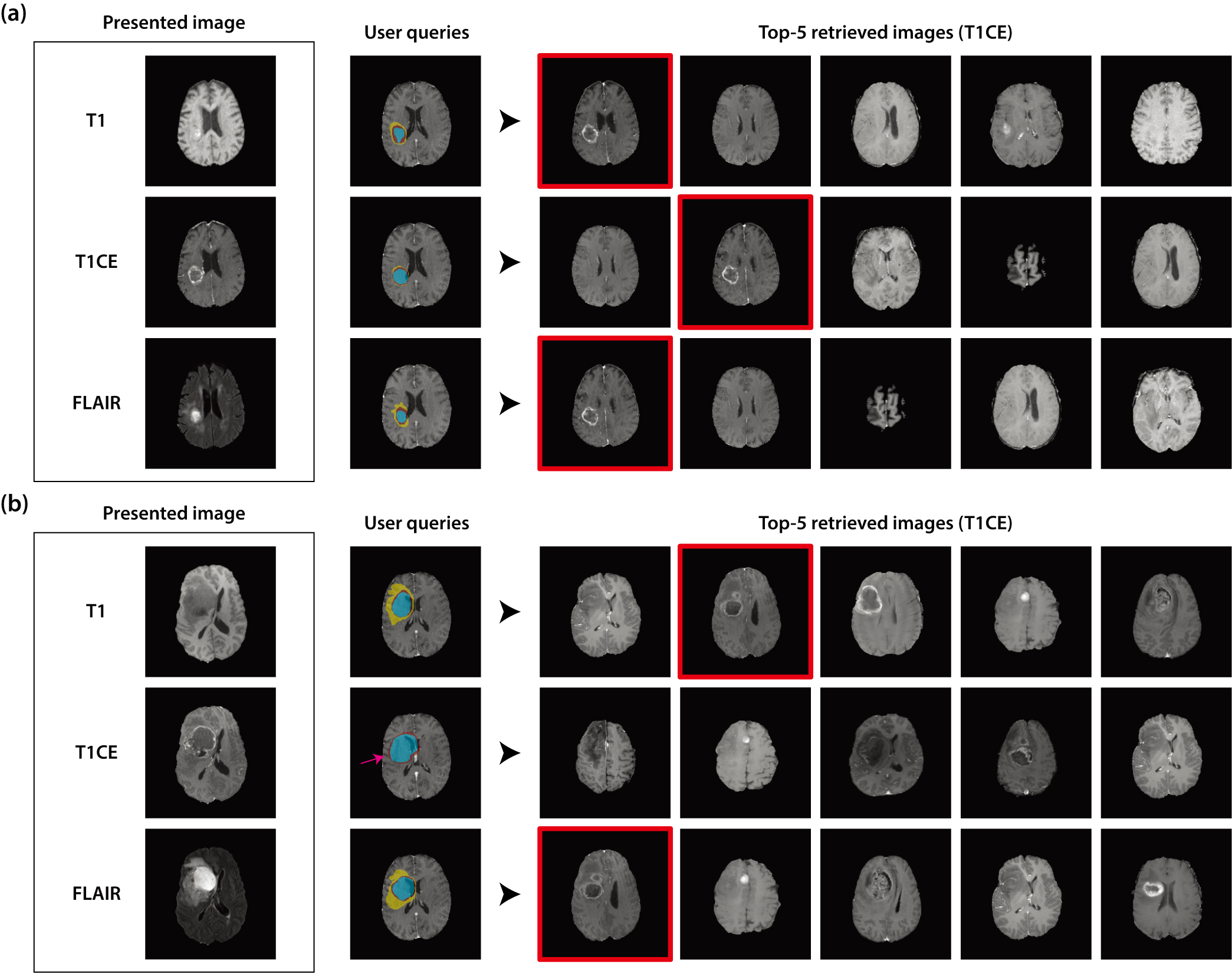}
  \caption{\textbf{Example results of Test-3 for gliomas.} \textbf{a} The presented image, which was an isolated sample, in Q.1 of Test-3 showed the highest recall@5 for gliomas. Three example user queries and corresponding top-5 retrieved images are shown. The retrieved images are arranged from left to right, starting with the most similar image. Retrieved images highlighted with red boxes are the images belonging to the same volume as the presented isolated sample (i.e., the same-volume images). Note that three different user queries successfully retrieved the same-volume images. \textbf{b} The presented image in Q.2 of Test-3 showed the lowest recall@5 for gliomas. The failed case is shown in the middle row, where none of the retrieved images are highlighted in a red box. This failure may be owing to the fact that the peritumoral edema was not sketched in the user query (see the arrow), in contrast with the other user queries, possibly indicating a skill-based limitation. T1, T1-weighted sequence; T1CE, T1-weighted contrast-enhanced sequence; FLAIR, fluid-attenuated inversion recovery sequence.}
  \label{fig:test_3_glioma}
\end{figure*}

\begin{figure*}[ht]
  \centering
  \includegraphics[width=\textwidth]{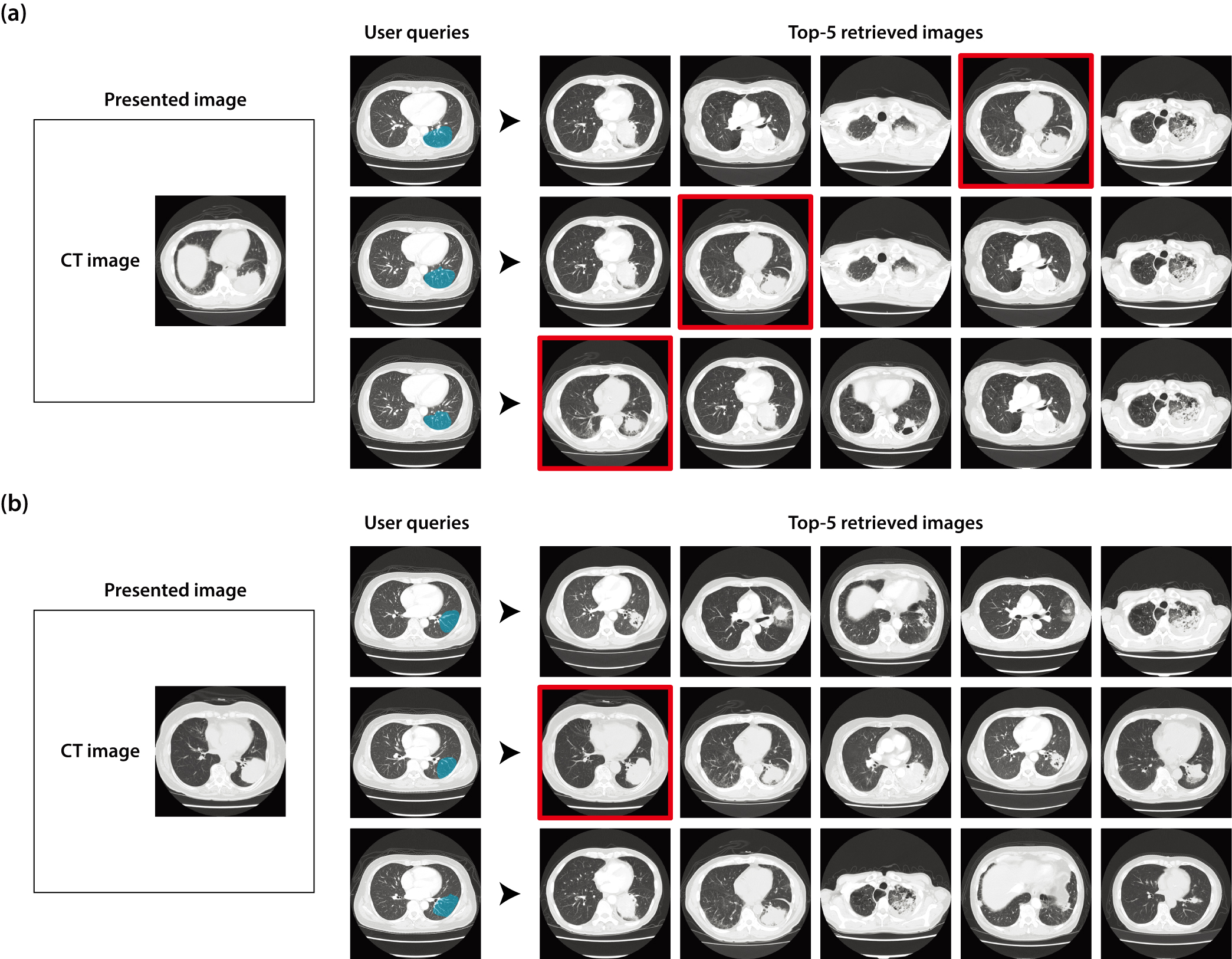}
  \caption{\textbf{Example results of Test-3 for lung cancers.} \textbf{a} The presented image, which was an isolated sample, in Q.4 of Test-3 showed the highest recall@5 for lung cancers. Three example user queries and corresponding top-5 retrieved images are shown. The retrieved images are arranged from left to right, starting with the most similar image. Retrieved images highlighted with red boxes are the images belonging to the same volume as the presented isolated sample (i.e., the same-volume images). Note that three different user queries successfully retrieved the same-volume images. \textbf{b} The presented image in Q.2 of Test-3 showed the lowest recall@5 for lung cancers. Even though the three user queries were relatively similar, only the query in the middle row successfully retrieved the same-volume image. This may be attributed to a template-image-based limitation owing to the large difference in the body size between the presented image and the template image. CT, computed tomography.}
  \label{fig:test_3_lung_cancer}
\end{figure*}

Example results of Test-3 for gliomas and those for lung cancers are shown in \textbf{\cref{fig:test_3_glioma}} and \textbf{\cref{fig:test_3_lung_cancer}}, respectively.

\bibliography{reference}

\clearpage

\end{document}